\documentclass{article}

 \usepackage[preprint]{neurips_2025}

\usepackage[utf8]{inputenc} %
\usepackage[T1]{fontenc}    %
\usepackage{url}            %
\usepackage{booktabs}       %
\usepackage{amsfonts}       %
\usepackage{nicefrac}       %
\usepackage{microtype}      %
\usepackage{xcolor}         %
\usepackage[frozencache=true]{minted}
\usepackage{caption}

\usepackage{amsmath,amsfonts,bm}

\def\eqref#1{equation~\ref{#1}}

\def\1{\bm{1}}

\def\vq{{\bm{q}}}

\def\vx{{\bm{x}}}
\def\vy{{\bm{y}}}

\DeclareMathAlphabet{\mathsfit}{\encodingdefault}{\sfdefault}{m}{sl}
\SetMathAlphabet{\mathsfit}{bold}{\encodingdefault}{\sfdefault}{bx}{n}

\usepackage{algorithm}
\usepackage[noend]{algpseudocode}
\usepackage{xcolor}
\usepackage{booktabs}
\usepackage{threeparttable}
\definecolor{darkGreen}{rgb}{0.2,0.5,0.2}
\definecolor{mydarkblue}{rgb}{0,0.08,0.45}
\usepackage[colorlinks,citecolor=mydarkblue,urlcolor=mydarkblue,linkcolor=mydarkblue]{hyperref}
\usepackage[capitalize,noabbrev]{cleveref}
\usepackage{url}
\usepackage{graphicx}
\usepackage{subcaption}
\usepackage{enumitem}
\usepackage{booktabs,threeparttable}
\usepackage{tikz}           %
\usepackage{multirow}
\usepackage{mathtools}
\usepackage{etoc} 

\usetikzlibrary{patterns,patterns.meta,positioning,backgrounds,calc}

\usepackage{todonotes}
\usepackage[normalem]{ulem}

\makeatletter
\newcommand*\iftodonotes{\if@todonotes@disabled\expandafter\@secondoftwo\else\expandafter\@firstoftwo\fi}  %
\makeatother

\usepackage[most]{tcolorbox}
\usepackage{tikz}
\usetikzlibrary{arrows.meta,positioning}

\definecolor{bfPrompt}{HTML}{6F42C1}
\definecolor{bfGood}{HTML}{1E9E6A}
\definecolor{bfForce}{HTML}{D9534F}

\newtcbox{\cue}{on line,
  colframe=bfForce, colback=bfForce!8,
  arc=1mm, boxrule=0.6pt,
  left=1pt, right=1pt, top=0pt, bottom=0pt}

\newcommand{\WaitCue}{\cue{\textbf{Wait,}}}

\tcbset{
  bf-common/.style={enhanced, breakable, arc=2mm, boxrule=0.9pt,
    left=3mm, right=3mm, top=2mm, bottom=2mm, %
  },
  bf-prompt/.style={bf-common, colframe=bfPrompt, colback=bfPrompt!6, title=Prompt},
  bf-ok/.style={bf-common, colframe=bfGood, colback=bfGood!6, title=Response},
  bf-force/.style={bf-common, colframe=bfForce, colback=bfForce!5},
}

\title{The Markovian Thinker: Architecture-Agnostic Linear Scaling of Reasoning}

\author{
  Milad Aghajohari$^{\text{*} \ 1}$,\,\,
  Kamran Chitsaz$^{\text{*} \ 1,6}$,\,\,
  Amirhossein Kazemnejad$^{\text{*}\ 1}$,\\
  {\normalfont\sffamily\bfseries\fontsize{8}{13}\selectfont Sarath Chandar}$^{1,5,6,7}$\textbf{,}\,\,
  {\normalfont\sffamily\bfseries\fontsize{8}{13}\selectfont Alessandro Sordoni}$^{\dagger \ 1,2}$\textbf{,}\,\,
  {\normalfont\sffamily\bfseries\fontsize{8}{13}\selectfont Aaron Courville}$^{\dagger \ 1,5,8}$\textbf{,}\,\,
  {\normalfont\sffamily\bfseries\fontsize{8}{13}\selectfont Siva Reddy}$^{\dagger \ 1,3,4,5}$\,\,
  \vspace{0.4em} \\
  {\small
  $^1$Mila \,\,$^2$Microsoft Research \,\,$^3$McGill University \,\,$^4$ServiceNow Research \,\,
  \\

  \small
  $^5$Canada CIFAR AI Chair \,\,$^6$Chandar Research Lab \,\,$^7$Polytechnique Montr\'eal \,\,$^8$Universit\'e de Montr\'eal \,\, \vspace{0.1em}
  \vspace{0.5em}\\

  \scriptsize
  $^*$Equal Contribution \ \ $^\dagger$Equal Advising\thanks{Correspondence:  \texttt{\{aghajohm,kamran.chitsaz,amirhossein.kazemnejad\}@mila.quebec}}
  }
}

\begin{document}

\usetikzlibrary{patterns,patterns.meta,positioning,backgrounds,calc}

\definecolor{mila-purple}{RGB}{102,46,125}
\definecolor{mila-khaki}{RGB}{245,231,206}
\definecolor{mila-khaki-dark}{RGB}{185,174,154}
\definecolor{mila-yellow}{HTML}{F9B40D}
\definecolor{mila-yellow-dark}{HTML}{DD950B}
\definecolor{mila-greenblue}{RGB}{130,207,190}
\definecolor{plot-blue}{RGB}{132,182,206}
\definecolor{plot-red}{RGB}{199,140,121}
\definecolor{plot-green}{RGB}{157,207,127}
\definecolor{mila-purple-1}{HTML}{EFE8F1}
\definecolor{mila-purple-2}{HTML}{D8C9DD}
\definecolor{mila-purple-3}{HTML}{C1AACA}
\definecolor{mila-purple-4}{HTML}{AA8BB7}
\definecolor{mila-purple-5}{HTML}{936BA3}
\definecolor{mila-purple-6}{HTML}{7C4D90}
\definecolor{mila-purple-dark}{HTML}{451059}
\definecolor{mila-purple-superdark}{HTML}{340040}

\definecolor{delethink-blue}{HTML}{3ba1ff}

\definecolor{delethink-purple}{HTML}{6441D2}
\definecolor{delethink-dark-purple}{HTML}{3C2880}

\newlength{\sz}         \setlength{\sz}{0.7cm}
\newlength{\sw}         \setlength{\sw}{0.4cm}
\newlength{\sww}        \setlength{\sww}{0.37cm}
\newlength{\rectr}         \setlength{\rectr}{5pt}
\newlength{\rectw}         \setlength{\rectw}{2cm}
\newlength{\rectwlongcot}  \setlength{\rectwlongcot}{13.6cm}
\newlength{\segap}      \setlength{\segap}{1.5cm}
\newlength{\segaparrow} \setlength{\segaparrow}{0.2cm}

\newlength{\BORDER}
\setlength{\BORDER}{1pt}

\tikzset{
  border/.style={line width=\BORDER},
  centerline/.style={dash pattern=on 1pt off 2pt, line cap=round, line width=\BORDER},
  arrowline/.style={->, line width=\BORDER},
  curvedarrow/.style={->, draw=black!30, line width=0.7pt, shorten >=2pt, shorten <=4pt}
}

\tikzset{
  segment marker shape/.is choice,
  segment marker shape/circle/.code={%
    \def\drawmarker##1{\fill[white] (##1) circle[radius=1.2pt];}%
  },
  segment marker shape/square/.code={%
    \def\drawmarker##1{\fill[white]
      ($(##1)+(-1.1pt,-1.1pt)$) rectangle ($(##1)+(1.1pt,1.1pt)$);}%
  },
  segment marker shape/diamond/.code={%
    \def\drawmarker##1{\fill[white]
      ($(##1)+(0,1.4pt)$)--($(##1)+(1.4pt,0)$)--($(##1)+(0,-1.4pt)$)--($(##1)+(-1.4pt,0)$)--cycle;}%
  },
  segment marker shape/triangle/.code={%
    \def\drawmarker##1{\fill[white]
      ($(##1)+(-1.2pt,-1.2pt)$)--($(##1)+(-1.2pt,1.2pt)$)--($(##1)+(1.6pt,0)$)--cycle;}%
  },
  segment marker shape/trangle/.style = {segment marker shape/triangle},
}

\tikzset{
  pics/segment/.style n args={2}{
    code={
      \begin{scope}[node distance=0pt]
        \begin{scope}[local bounding box=sq]
          \path[
            draw=delethink-purple, line width=1.4\BORDER, dashed,
            preaction={fill=delethink-purple!5}, pattern color=black!30
          ]
          (\rectr,0) -- (\sw,0) -- (\sw,\sz) -- (\rectr,\sz)
          arc (90:180:\rectr) -- (0,\rectr) arc (180:270:\rectr) -- cycle;
        \end{scope}
        \node[font=\bfseries\large] at (sq.center) {$\mathrm{q}$};

        \node[
          draw=delethink-purple, line width=1.4\BORDER,
          minimum height=\sz, minimum width=\rectw, inner sep=0pt, right=3pt of sq
        ] (rect) {};

        \draw[dashed, line cap=round,
      shorten >=-3pt, shorten <=-3pt](rect.north) -- (rect.south);

        \begin{scope}[on background layer]
          \fill[#1] (rect.south west) rectangle (rect.north);
          \fill[#2] (rect.south) rectangle (rect.north east);
        \end{scope}

        \begin{scope}
          \clip (rect.south west) rectangle (rect.north east);
        
          \def\dia{1.8pt}
        
          \coordinate (gxW) at ($(rect.west)!0.12!(rect.east)$);
          \coordinate (gxE) at ($(rect.west)!0.42!(rect.east)$);
        
          \coordinate (gy1) at ($(rect.south)!0.75!(rect.north)$);
          \coordinate (p1)  at ($(gxW|-gy1)$);
          \fill[delethink-purple!40]
            ($(p1)+(-\dia,0)$) -- ($(p1)+(0,\dia)$) -- ($(p1)+(\dia,0)$) -- ($(p1)+(0,-\dia)$) -- cycle;
          \draw[delethink-purple!40, line width=\BORDER, line cap=round]
            ($(p1)+(\dia+0.6pt,0)$) -- ([xshift=-2pt]gxE|-gy1);
        
          \coordinate (gy2) at ($(rect.south)!0.50!(rect.north)$);
          \coordinate (p2)  at ($(gxW|-gy2)$);
          \fill[delethink-purple!40]
            ($(p2)+(-\dia,0)$) -- ($(p2)+(0,\dia)$) -- ($(p2)+(\dia,0)$) -- ($(p2)+(0,-\dia)$) -- cycle;
          \draw[delethink-purple!40, line width=\BORDER, line cap=round]
            ($(p2)+(\dia+0.6pt,0)$) -- ([xshift=-5pt]gxE|-gy2);
        
          \coordinate (gy3) at ($(rect.south)!0.25!(rect.north)$);
          \coordinate (p3)  at ($(gxW|-gy3)$);
          \fill[delethink-purple!40]
            ($(p3)+(-\dia,0)$) -- ($(p3)+(0,\dia)$) -- ($(p3)+(\dia,0)$) -- ($(p3)+(0,-\dia)$) -- cycle;
          \draw[delethink-purple!40, line width=\BORDER, line cap=round]
            ($(p3)+(\dia+0.6pt,0)$) -- (gxE|-gy3);
        \end{scope}

        \begin{scope}
          \clip (rect.south west) rectangle (rect.north east);
        
          \def\dia{1.8pt}
        
          \coordinate (gxW) at ($(rect.west)!0.585!(rect.east)$);
          \coordinate (gxE) at ($(rect.west)!0.89!(rect.east)$);
        
          \coordinate (gy1) at ($(rect.south)!0.75!(rect.north)$);
          \coordinate (p1)  at ($(gxW|-gy1)$);
          \fill[delethink-purple!40]
            ($(p1)+(-\dia,0)$) -- ($(p1)+(0,\dia)$) -- ($(p1)+(\dia,0)$) -- ($(p1)+(0,-\dia)$) -- cycle;
          \draw[delethink-purple!40, line width=\BORDER, line cap=round]
            ($(p1)+(\dia+0.6pt,0)$) -- ([xshift=-7pt]gxE|-gy1);
        
          \coordinate (gy2) at ($(rect.south)!0.50!(rect.north)$);
          \coordinate (p2)  at ($(gxW|-gy2)$);
          \fill[delethink-purple!40]
            ($(p2)+(-\dia,0)$) -- ($(p2)+(0,\dia)$) -- ($(p2)+(\dia,0)$) -- ($(p2)+(0,-\dia)$) -- cycle;
          \draw[delethink-purple!40, line width=\BORDER, line cap=round]
            ($(p2)+(\dia+0.6pt,0)$) -- ([xshift=-1pt]gxE|-gy2);
        
          \coordinate (gy3) at ($(rect.south)!0.25!(rect.north)$);
          \coordinate (p3)  at ($(gxW|-gy3)$);
          \fill[delethink-purple!40]
            ($(p3)+(-\dia,0)$) -- ($(p3)+(0,\dia)$) -- ($(p3)+(\dia,0)$) -- ($(p3)+(0,-\dia)$) -- cycle;
          \draw[delethink-purple!40, line width=\BORDER, line cap=round]
            ($(p3)+(\dia+0.6pt,0)$) -- ([xshift=-5pt]gxE|-gy3);
        \end{scope}

        \coordinate (-sqwest)   at (sq.west);
        \coordinate (-rectwest) at (rect.west);
        \coordinate (-recteast) at (rect.east);
        \coordinate (-east)     at (rect.east);
        \coordinate (-rectnorth) at ([xshift=0.25\rectw]rect.north);
        \coordinate (-rectsouth) at (rect.south);
      \end{scope}
    }
  },
  pics/segment/.default={mila-purple-1}{mila-purple-2}
}

\tikzset{
  pics/segmentlongcot/.style n args={2}{
    code={
      \begin{scope}[node distance=0pt]
        \begin{scope}[local bounding box=sq]
          \path[
            draw=#2, line width=1.4\BORDER, dashed,
            preaction={fill=delethink-blue!5}, pattern color=black!30
          ]
          (\rectr,0) -- (\sw,0) -- (\sw,\sz) -- (\rectr,\sz)
          arc (90:180:\rectr) -- (0,\rectr) arc (180:270:\rectr) -- cycle;
        \end{scope}
        \node[font=\bfseries\large] at (sq.center) {$\mathrm{q}$};

        \node[
          draw=#2, line width=1.4\BORDER,
          minimum height=\sz, minimum width=\rectwlongcot, inner sep=0pt, right=3pt of sq
        ] (rect) {};

        \begin{scope}[on background layer]
          \fill[#1] (rect.south west) rectangle (rect.north east);
        \end{scope}

        \begin{scope}
          \clip (rect.south west) rectangle (rect.north east);
        
          \def\dia{1.8pt}
        
          \coordinate (gxW) at ($(rect.west)!0.019!(rect.east)$);
          \coordinate (gxE) at ($(rect.west)!0.065!(rect.east)$);
        
          \coordinate (gy1) at ($(rect.south)!0.75!(rect.north)$);
          \coordinate (p1)  at ($(gxW|-gy1)$);
          \fill[delethink-blue!60]
            ($(p1)+(-\dia,0)$) -- ($(p1)+(0,\dia)$) -- ($(p1)+(\dia,0)$) -- ($(p1)+(0,-\dia)$) -- cycle;
          \draw[delethink-blue!60, line width=\BORDER, line cap=round]
            ($(p1)+(\dia+0.6pt,0)$) -- ([xshift=-2pt]gxE|-gy1);
        
          \coordinate (gy2) at ($(rect.south)!0.50!(rect.north)$);
          \coordinate (p2)  at ($(gxW|-gy2)$);
          \fill[delethink-blue!60]
            ($(p2)+(-\dia,0)$) -- ($(p2)+(0,\dia)$) -- ($(p2)+(\dia,0)$) -- ($(p2)+(0,-\dia)$) -- cycle;
          \draw[delethink-blue!60, line width=\BORDER, line cap=round]
            ($(p2)+(\dia+0.6pt,0)$) -- ([xshift=-5pt]gxE|-gy2);
        
          \coordinate (gy3) at ($(rect.south)!0.25!(rect.north)$);
          \coordinate (p3)  at ($(gxW|-gy3)$);
          \fill[delethink-blue!60]
            ($(p3)+(-\dia,0)$) -- ($(p3)+(0,\dia)$) -- ($(p3)+(\dia,0)$) -- ($(p3)+(0,-\dia)$) -- cycle;
          \draw[delethink-blue!60, line width=\BORDER, line cap=round]
            ($(p3)+(\dia+0.6pt,0)$) -- ([xshift=-0pt]gxE|-gy3);
        \end{scope}

        \begin{scope}
          \clip (rect.south west) rectangle (rect.north east);
        
          \def\dia{1.8pt}
        
          \coordinate (gxW) at ($(rect.west)!0.085!(rect.east)$);
          \coordinate (gxE) at ($(rect.west)!0.130!(rect.east)$);
        
          \coordinate (gy1) at ($(rect.south)!0.75!(rect.north)$);
          \coordinate (p1)  at ($(gxW|-gy1)$);
          \fill[delethink-blue!50]
            ($(p1)+(-\dia,0)$) -- ($(p1)+(0,\dia)$) -- ($(p1)+(\dia,0)$) -- ($(p1)+(0,-\dia)$) -- cycle;
          \draw[delethink-blue!50, line width=\BORDER, line cap=round]
            ($(p1)+(\dia+0.6pt,0)$) -- ([xshift=-6pt]gxE|-gy1);
        
          \coordinate (gy2) at ($(rect.south)!0.50!(rect.north)$);
          \coordinate (p2)  at ($(gxW|-gy2)$);
          \fill[delethink-blue!50]
            ($(p2)+(-\dia,0)$) -- ($(p2)+(0,\dia)$) -- ($(p2)+(\dia,0)$) -- ($(p2)+(0,-\dia)$) -- cycle;
          \draw[delethink-blue!50, line width=\BORDER, line cap=round]
            ($(p2)+(\dia+0.6pt,0)$) -- ([xshift=-0pt]gxE|-gy2);
        
          \coordinate (gy3) at ($(rect.south)!0.25!(rect.north)$);
          \coordinate (p3)  at ($(gxW|-gy3)$);
          \fill[delethink-blue!50]
            ($(p3)+(-\dia,0)$) -- ($(p3)+(0,\dia)$) -- ($(p3)+(\dia,0)$) -- ($(p3)+(0,-\dia)$) -- cycle;
          \draw[delethink-blue!50, line width=\BORDER, line cap=round]
            ($(p3)+(\dia+0.6pt,0)$) -- ([xshift=-4pt]gxE|-gy3);
        \end{scope}

        \begin{scope}
          \clip (rect.south west) rectangle (rect.north east);
        
          \def\dia{1.8pt}
        
          \coordinate (gxW) at ($(rect.west)!0.15!(rect.east)$);
          \coordinate (gxE) at ($(rect.west)!0.19!(rect.east)$);
        
          \coordinate (gy1) at ($(rect.south)!0.75!(rect.north)$);
          \coordinate (p1)  at ($(gxW|-gy1)$);
          \fill[delethink-blue!40]
            ($(p1)+(-\dia,0)$) -- ($(p1)+(0,\dia)$) -- ($(p1)+(\dia,0)$) -- ($(p1)+(0,-\dia)$) -- cycle;
          \draw[delethink-blue!40, line width=\BORDER, line cap=round]
            ($(p1)+(\dia+0.6pt,0)$) -- ([xshift=-0pt]gxE|-gy1);
        
          \coordinate (gy2) at ($(rect.south)!0.50!(rect.north)$);
          \coordinate (p2)  at ($(gxW|-gy2)$);
          \fill[delethink-blue!40]
            ($(p2)+(-\dia,0)$) -- ($(p2)+(0,\dia)$) -- ($(p2)+(\dia,0)$) -- ($(p2)+(0,-\dia)$) -- cycle;
          \draw[delethink-blue!40, line width=\BORDER, line cap=round]
            ($(p2)+(\dia+0.6pt,0)$) -- ([xshift=-9pt]gxE|-gy2);
        
          \coordinate (gy3) at ($(rect.south)!0.25!(rect.north)$);
          \coordinate (p3)  at ($(gxW|-gy3)$);
          \fill[delethink-blue!40]
            ($(p3)+(-\dia,0)$) -- ($(p3)+(0,\dia)$) -- ($(p3)+(\dia,0)$) -- ($(p3)+(0,-\dia)$) -- cycle;
          \draw[delethink-blue!40, line width=\BORDER, line cap=round]
            ($(p3)+(\dia+0.6pt,0)$) -- ([xshift=-6pt]gxE|-gy3);
        \end{scope}

        \begin{scope}
          \clip (rect.south west) rectangle (rect.north east);
        
          \def\dia{1.8pt}
        
          \coordinate (gxW) at ($(rect.west)!0.21!(rect.east)$);
          \coordinate (gxE) at ($(rect.west)!0.26!(rect.east)$);
        
          \coordinate (gy1) at ($(rect.south)!0.75!(rect.north)$);
          \coordinate (p1)  at ($(gxW|-gy1)$);
          \fill[delethink-blue!30]
            ($(p1)+(-\dia,0)$) -- ($(p1)+(0,\dia)$) -- ($(p1)+(\dia,0)$) -- ($(p1)+(0,-\dia)$) -- cycle;
          \draw[delethink-blue!30, line width=\BORDER, line cap=round]
            ($(p1)+(\dia+0.6pt,0)$) -- ([xshift=-5pt]gxE|-gy1);
        
          \coordinate (gy2) at ($(rect.south)!0.50!(rect.north)$);
          \coordinate (p2)  at ($(gxW|-gy2)$);
          \fill[delethink-blue!30]
            ($(p2)+(-\dia,0)$) -- ($(p2)+(0,\dia)$) -- ($(p2)+(\dia,0)$) -- ($(p2)+(0,-\dia)$) -- cycle;
          \draw[delethink-blue!30, line width=\BORDER, line cap=round]
            ($(p2)+(\dia+0.6pt,0)$) -- ([xshift=-6pt]gxE|-gy2);
        
          \coordinate (gy3) at ($(rect.south)!0.25!(rect.north)$);
          \coordinate (p3)  at ($(gxW|-gy3)$);
          \fill[delethink-blue!30]
            ($(p3)+(-\dia,0)$) -- ($(p3)+(0,\dia)$) -- ($(p3)+(\dia,0)$) -- ($(p3)+(0,-\dia)$) -- cycle;
          \draw[delethink-blue!30, line width=\BORDER, line cap=round]
            ($(p3)+(\dia+0.6pt,0)$) -- ([xshift=-0pt]gxE|-gy3);
        \end{scope}

        \coordinate (-sqwest)   at (sq.west);
        \coordinate (-rectwest) at (rect.west);
        \coordinate (-recteast) at (rect.east);
        \coordinate (-east)     at (rect.east);
        \coordinate (-rectnorth) at ([xshift=0.25\rectw]rect.north);
        \coordinate (-rectsouth) at (rect.south);
      \end{scope}
    }
  },
  pics/segmentlongcot/.default={mila-purple-1}{mila-purple}
}

\tikzset{
  pics/segmentintermediate/.style n args={3}{
    code={
      \begin{scope}[node distance=0pt]
        \begin{scope}[local bounding box=sq]
          \path[draw=none, preaction={fill=delethink-purple!5}, pattern color=black!30]
            (\rectr,0) -- (\sww,0) -- (\sww,\sz) -- (\rectr,\sz)
            arc (90:180:\rectr) -- (0,\rectr) arc (180:270:\rectr) -- cycle;
          \draw[delethink-purple, dashed, line width=1.4\BORDER]
            (\sww,\sz) -- (\rectr,\sz) arc (90:180:\rectr) -- (0,\rectr)
            arc (180:270:\rectr) -- (\rectr,0) -- (\sww,0);
          \path[use as bounding box] (0,0) rectangle (\sww,\sz);
        \end{scope}
        \node[font=\bfseries\large] at (sq.center) {$\mathrm{q}$};

        \node[
          draw=none, minimum height=\sz, minimum width=0.42\rectw,
          inner sep=0pt, outer sep=0pt, right=0pt of sq
        ] (rectprmpt) {};
        \begin{scope}[on background layer]
          \fill[#1] (rectprmpt.south west) rectangle (rectprmpt.north east);
        \end{scope}
        \draw[delethink-purple, line width=1.4\BORDER, dashed]
          (rectprmpt.north west) -- (rectprmpt.north east)
          (rectprmpt.north east) -- (rectprmpt.south east)
          (rectprmpt.south east) -- (rectprmpt.south west);

        \node[
          draw=#3, line width=1.4\BORDER,
          minimum height=\sz, minimum width=0.5\rectw,
          inner sep=0pt, outer sep=0pt, right=4pt of rectprmpt
        ] (rect) {};
        \begin{scope}[on background layer]
          \fill[#2] (rect.south west) rectangle (rect.north east);
        \end{scope}

        \begin{scope}
          \clip (rect.south west) rectangle (rect.north east);
        
          \path coordinate (gxW) at ($(rect.west)!0.18!(rect.east)$);
          \path coordinate (gxE) at ($(rect.west)!0.80!(rect.east)$);
        
          \path coordinate (gy1) at ($(rect.south)!0.75!(rect.north)$);
          \fill[white!60] (gxW|-gy1) circle[radius=1.2pt];
          \draw[white!60, line width=\BORDER, line cap=round]
               ($(gxW|-gy1)+(1.6pt,0)$) -- (gxE|-gy1);
        
          \path coordinate (gy2) at ($(rect.south)!0.50!(rect.north)$);
          \fill[white!60] (gxW|-gy2) circle[radius=1.2pt];
          \draw[white!60, line width=\BORDER, line cap=round]
               ($(gxW|-gy2)+(1.6pt,0)$) -- (gxE|-gy2);
        
          \path coordinate (gy3) at ($(rect.south)!0.25!(rect.north)$);
          \fill[white!60] (gxW|-gy3) circle[radius=1.2pt];
          \draw[white!60, line width=\BORDER, line cap=round]
               ($(gxW|-gy3)+(1.6pt,0)$) -- (gxE|-gy3);
        \end{scope}

        \coordinate (-sqwest)   at (sq.west);
        \coordinate (-rectwest) at (rect.west);
        \coordinate (-recteast) at (rect.east);
        \coordinate (-east)     at (rect.east);
        \coordinate (-rectnorth) at (rect.north);
        \coordinate (-rectsouth) at (rectprmpt.south);
      \end{scope}
    }
  },
  pics/segmentintermediate/.default={mila-purple-1}{mila-purple-2}{mila-purple}
}

\tikzset{
  pics/segmentintermediatecircle/.style n args={3}{
    code={
      \begin{scope}[node distance=0pt]
        \begin{scope}[local bounding box=sq]
          \path[draw=none, preaction={fill=delethink-purple!5}, pattern color=black!30]
            (\rectr,0) -- (\sww,0) -- (\sww,\sz) -- (\rectr,\sz)
            arc (90:180:\rectr) -- (0,\rectr) arc (180:270:\rectr) -- cycle;
          \draw[delethink-purple, dashed, line width=1.4\BORDER]
            (\sww,\sz) -- (\rectr,\sz) arc (90:180:\rectr) -- (0,\rectr)
            arc (180:270:\rectr) -- (\rectr,0) -- (\sww,0);
          \path[use as bounding box] (0,0) rectangle (\sww,\sz);
        \end{scope}
        \node[font=\bfseries\large] at (sq.center) {$\mathrm{q}$};

        \node[
          draw=none, minimum height=\sz, minimum width=0.42\rectw,
          inner sep=0pt, outer sep=0pt, right=0pt of sq
        ] (rectprmpt) {};
        \begin{scope}[on background layer]
          \fill[#1] (rectprmpt.south west) rectangle (rectprmpt.north east);
        \end{scope}
        \draw[delethink-purple, line width=1.4\BORDER, dashed]
          (rectprmpt.north west) -- (rectprmpt.north east)
          (rectprmpt.north east) -- (rectprmpt.south east)
          (rectprmpt.south east) -- (rectprmpt.south west);

        \begin{scope}
          \clip (rectprmpt.south west) rectangle (rectprmpt.north east);
        
          \def\dia{1.8pt}
        
          \coordinate (gxW) at ($(rectprmpt.west)!0.19!(rectprmpt.east)$);
          \coordinate (gxE) at ($(rectprmpt.west)!0.82!(rectprmpt.east)$);
        
          \coordinate (gy1) at ($(rectprmpt.south)!0.75!(rectprmpt.north)$);
          \coordinate (p1)  at ($(gxW|-gy1)$);
          \fill[delethink-purple!40]
            ($(p1)+(-\dia,0)$) -- ($(p1)+(0,\dia)$) -- ($(p1)+(\dia,0)$) -- ($(p1)+(0,-\dia)$) -- cycle;
          \draw[delethink-purple!40, line width=\BORDER, line cap=round]
            ($(p1)+(\dia+0.6pt,0)$) -- ([xshift=-6pt]gxE|-gy1);
        
          \coordinate (gy2) at ($(rectprmpt.south)!0.50!(rectprmpt.north)$);
          \coordinate (p2)  at ($(gxW|-gy2)$);
          \fill[delethink-purple!40]
            ($(p2)+(-\dia,0)$) -- ($(p2)+(0,\dia)$) -- ($(p2)+(\dia,0)$) -- ($(p2)+(0,-\dia)$) -- cycle;
          \draw[delethink-purple!40, line width=\BORDER, line cap=round]
            ($(p2)+(\dia+0.6pt,0)$) -- ([xshift=-0pt]gxE|-gy2);
        
          \coordinate (gy3) at ($(rectprmpt.south)!0.25!(rectprmpt.north)$);
          \coordinate (p3)  at ($(gxW|-gy3)$);
          \fill[delethink-purple!40]
            ($(p3)+(-\dia,0)$) -- ($(p3)+(0,\dia)$) -- ($(p3)+(\dia,0)$) -- ($(p3)+(0,-\dia)$) -- cycle;
          \draw[delethink-purple!40, line width=\BORDER, line cap=round]
            ($(p3)+(\dia+0.6pt,0)$) -- ([xshift=-4pt]gxE|-gy3);
        \end{scope}

        \node[
          draw=#3, line width=1.4\BORDER,
          minimum height=\sz, minimum width=0.5\rectw,
          inner sep=0pt, outer sep=0pt, right=4pt of rectprmpt
        ] (rect) {};
        \begin{scope}[on background layer]
          \fill[#2] (rect.south west) rectangle (rect.north east);
        \end{scope}

        \begin{scope}
          \clip (rect.south west) rectangle (rect.north east);
        
          \path coordinate (gxW) at ($(rect.west)!0.18!(rect.east)$);
          \path coordinate (gxE) at ($(rect.west)!0.75!(rect.east)$);
        
          \path coordinate (gy1) at ($(rect.south)!0.75!(rect.north)$);
          \fill[white!60] (gxW|-gy1) circle[radius=1.2pt];
          \draw[white!60, line width=\BORDER, line cap=round]
               ($(gxW|-gy1)+(1.6pt,0)$) -- ([xshift=-0pt]gxE|-gy1);
        
          \path coordinate (gy2) at ($(rect.south)!0.50!(rect.north)$);
          \fill[white!60] (gxW|-gy2) circle[radius=1.2pt];
          \draw[white!60, line width=\BORDER, line cap=round]
               ($(gxW|-gy2)+(1.6pt,0)$) -- ([xshift=-3pt]gxE|-gy2);
        
          \path coordinate (gy3) at ($(rect.south)!0.25!(rect.north)$);
          \fill[white!60] (gxW|-gy3) circle[radius=1.2pt];
          \draw[white!60, line width=\BORDER, line cap=round]
               ($(gxW|-gy3)+(1.6pt,0)$) -- ([xshift=-6pt]gxE|-gy3);
        \end{scope}

        \coordinate (-sqwest)   at (sq.west);
        \coordinate (-rectwest) at (rect.west);
        \coordinate (-recteast) at (rect.east);
        \coordinate (-east)     at (rect.east);
        \coordinate (-rectnorth) at (rect.north);
        \coordinate (-rectsouth) at (rectprmpt.south);
      \end{scope}
    }
  },
  pics/segmentintermediatecircle/.default={mila-purple-1}{mila-purple-2}{mila-purple}
}

\tikzset{
  pics/segmentintermediatesquare/.style n args={3}{
    code={
      \begin{scope}[node distance=0pt]
        \begin{scope}[local bounding box=sq]
          \path[draw=none, preaction={fill=delethink-purple!5}, pattern color=black!30]
            (\rectr,0) -- (\sww,0) -- (\sww,\sz) -- (\rectr,\sz)
            arc (90:180:\rectr) -- (0,\rectr) arc (180:270:\rectr) -- cycle;
          \draw[delethink-purple, dashed, line width=1.4\BORDER]
            (\sww,\sz) -- (\rectr,\sz) arc (90:180:\rectr) -- (0,\rectr)
            arc (180:270:\rectr) -- (\rectr,0) -- (\sww,0);
          \path[use as bounding box] (0,0) rectangle (\sww,\sz);
        \end{scope}
        \node[font=\bfseries\large] at (sq.center) {$\mathrm{q}$};

        \node[
          draw=none, minimum height=\sz, minimum width=0.42\rectw,
          inner sep=0pt, outer sep=0pt, right=0pt of sq
        ] (rectprmpt) {};
        \begin{scope}[on background layer]
          \fill[#1] (rectprmpt.south west) rectangle (rectprmpt.north east);
        \end{scope}
        \draw[delethink-purple, line width=1.4\BORDER, dashed]
          (rectprmpt.north west) -- (rectprmpt.north east)
          (rectprmpt.north east) -- (rectprmpt.south east)
          (rectprmpt.south east) -- (rectprmpt.south west);

        \begin{scope}
          \clip (rectprmpt.south west) rectangle (rectprmpt.north east);
        
          \def\triH{1.2pt}   %
          \def\triL{1.8
          pt}   %
        
          \coordinate (gxW) at ($(rectprmpt.west)!0.19!(rectprmpt.east)$);
          \coordinate (gxE) at ($(rectprmpt.west)!0.82!(rectprmpt.east)$);
        
          \coordinate (gy1) at ($(rectprmpt.south)!0.75!(rectprmpt.north)$);
          \coordinate (p1)  at ($(gxW|-gy1)$);
          \fill[white]
            ($(p1)+(-\triL,-\triH)$) --
            ($(p1)+(-\triL, \triH)$) --
            ($(p1)+(\triL,0)$) -- cycle;
          \draw[white, line width=\BORDER, line cap=round]
            ($(p1)+(\triL+0.6pt,0)$) -- ([xshift=-3pt]gxE|-gy1);
        
          \coordinate (gy2) at ($(rectprmpt.south)!0.50!(rectprmpt.north)$);
          \coordinate (p2)  at ($(gxW|-gy2)$);
          \fill[white]
            ($(p2)+(-\triL,-\triH)$) --
            ($(p2)+(-\triL, \triH)$) --
            ($(p2)+(\triL,0)$) -- cycle;
          \draw[white, line width=\BORDER, line cap=round]
            ($(p2)+(\triL+0.6pt,0)$) -- ([xshift=-5pt]gxE|-gy2);
        
          \coordinate (gy3) at ($(rectprmpt.south)!0.25!(rectprmpt.north)$);
          \coordinate (p3)  at ($(gxW|-gy3)$);
          \fill[white]
            ($(p3)+(-\triL,-\triH)$) --
            ($(p3)+(-\triL, \triH)$) --
            ($(p3)+(\triL,0)$) -- cycle;
          \draw[white, line width=\BORDER, line cap=round]
            ($(p3)+(\triL+0.6pt,0)$) -- (gxE|-gy3);
        \end{scope}

        \node[
          draw=#3, line width=1.4\BORDER,
          minimum height=\sz, minimum width=0.5\rectw,
          inner sep=0pt, outer sep=0pt, right=4pt of rectprmpt
        ] (rect) {};
        \begin{scope}[on background layer]
          \fill[#2] (rect.south west) rectangle (rect.north east);
        \end{scope}

        \begin{scope}
          \clip (rect.south west) rectangle (rect.north east);
        
          \def\sq{2.2pt} %
          \path coordinate (gxW) at ($(rect.west)!0.19!(rect.east)$);
          \path coordinate (gxE) at ($(rect.west)!0.75!(rect.east)$);
        
          \path coordinate (gy1) at ($(rect.south)!0.75!(rect.north)$);
          \fill[white] ($(gxW|-gy1)+(-.5*\sq,-.5*\sq)$) rectangle ++(\sq,\sq);
          \draw[white, line width=\BORDER, line cap=round]
               ($(gxW|-gy1)+(.5*\sq+0.6pt,0)$) -- ([xshift=-8pt]gxE|-gy1);
        
          \path coordinate (gy2) at ($(rect.south)!0.50!(rect.north)$);
          \fill[white] ($(gxW|-gy2)+(-.5*\sq,-.5*\sq)$) rectangle ++(\sq,\sq);
          \draw[white, line width=\BORDER, line cap=round]
               ($(gxW|-gy2)+(.5*\sq+0.6pt,0)$) -- ([xshift=-5pt]gxE|-gy2);
        
          \path coordinate (gy3) at ($(rect.south)!0.25!(rect.north)$);
          \fill[white] ($(gxW|-gy3)+(-.5*\sq,-.5*\sq)$) rectangle ++(\sq,\sq);
          \draw[white, line width=\BORDER, line cap=round]
               ($(gxW|-gy3)+(.5*\sq+0.6pt,0)$) -- (gxE|-gy3);
        \end{scope}

        \coordinate (-sqwest)   at (sq.west);
        \coordinate (-rectwest) at (rect.west);
        \coordinate (-recteast) at (rect.east);
        \coordinate (-east)     at (rect.east);
        \coordinate (-rectnorth) at (rect.north);
        \coordinate (-rectsouth) at (rectprmpt.south);
      \end{scope}
    }
  },
  pics/segmentintermediatesquare/.default={mila-purple-1}{mila-purple-2}{mila-purple}
}

\tikzset{
  pics/segmentintermediatetriangle/.style n args={3}{
    code={
      \begin{scope}[node distance=0pt]
        \begin{scope}[local bounding box=sq]
          \path[draw=none, preaction={fill=delethink-purple!5}, pattern color=black!30]
            (\rectr,0) -- (\sww,0) -- (\sww,\sz) -- (\rectr,\sz)
            arc (90:180:\rectr) -- (0,\rectr) arc (180:270:\rectr) -- cycle;
          \draw[delethink-purple, dashed, line width=1.4\BORDER]
            (\sww,\sz) -- (\rectr,\sz) arc (90:180:\rectr) -- (0,\rectr)
            arc (180:270:\rectr) -- (\rectr,0) -- (\sww,0);
          \path[use as bounding box] (0,0) rectangle (\sww,\sz);
        \end{scope}
        \node[font=\bfseries\large] at (sq.center) {$\mathrm{q}$};

        \node[
          draw=none, minimum height=\sz, minimum width=0.42\rectw,
          inner sep=0pt, outer sep=0pt, right=0pt of sq
        ] (rectprmpt) {};
        \begin{scope}[on background layer]
          \fill[#1] (rectprmpt.south west) rectangle (rectprmpt.north east);
        \end{scope}
        \draw[delethink-purple, line width=1.4\BORDER, dashed]
          (rectprmpt.north west) -- (rectprmpt.north east)
          (rectprmpt.north east) -- (rectprmpt.south east)
          (rectprmpt.south east) -- (rectprmpt.south west);

        \begin{scope}
          \clip (rectprmpt.south west) rectangle (rectprmpt.north east);
        
          \path coordinate (gxW) at ($(rectprmpt.west)!0.19!(rectprmpt.east)$);
          \path coordinate (gxE) at ($(rectprmpt.west)!0.81!(rectprmpt.east)$);
        
          \path coordinate (gy1) at ($(rectprmpt.south)!0.75!(rectprmpt.north)$);
          \fill[white!60] (gxW|-gy1) circle[radius=1.2pt];
          \draw[white!60, line width=\BORDER, line cap=round]
               ($(gxW|-gy1)+(1.6pt,0)$) -- ([xshift=-0pt]gxE|-gy1);
        
          \path coordinate (gy2) at ($(rectprmpt.south)!0.50!(rectprmpt.north)$);
          \fill[white!60] (gxW|-gy2) circle[radius=1.2pt];
          \draw[white!60, line width=\BORDER, line cap=round]
               ($(gxW|-gy2)+(1.6pt,0)$) -- ([xshift=-3pt]gxE|-gy2);
        
          \path coordinate (gy3) at ($(rectprmpt.south)!0.25!(rectprmpt.north)$);
          \fill[white!60] (gxW|-gy3) circle[radius=1.2pt];
          \draw[white!60, line width=\BORDER, line cap=round]
               ($(gxW|-gy3)+(1.6pt,0)$) -- ([xshift=-6pt]gxE|-gy3);
        \end{scope}

        \node[
          draw=#3, line width=1.4\BORDER,
          minimum height=\sz, minimum width=0.5\rectw,
          inner sep=0pt, outer sep=0pt, right=4pt of rectprmpt
        ] (rect) {};
        \begin{scope}[on background layer]
          \fill[#2] (rect.south west) rectangle (rect.north east);
        \end{scope}

        \begin{scope}
          \clip (rect.south west) rectangle (rect.north east);
        
          \def\triH{1.2pt}   %
          \def\triL{1.8
          pt}   %
        
          \coordinate (gxW) at ($(rect.west)!0.18!(rect.east)$);
          \coordinate (gxE) at ($(rect.west)!0.73!(rect.east)$);
        
          \coordinate (gy1) at ($(rect.south)!0.75!(rect.north)$);
          \coordinate (p1)  at ($(gxW|-gy1)$);
          \fill[white]
            ($(p1)+(-\triL,-\triH)$) --
            ($(p1)+(-\triL, \triH)$) --
            ($(p1)+(\triL,0)$) -- cycle;
          \draw[white, line width=\BORDER, line cap=round]
            ($(p1)+(\triL+0.6pt,0)$) -- ([xshift=-3pt]gxE|-gy1);
        
          \coordinate (gy2) at ($(rect.south)!0.50!(rect.north)$);
          \coordinate (p2)  at ($(gxW|-gy2)$);
          \fill[white]
            ($(p2)+(-\triL,-\triH)$) --
            ($(p2)+(-\triL, \triH)$) --
            ($(p2)+(\triL,0)$) -- cycle;
          \draw[white, line width=\BORDER, line cap=round]
            ($(p2)+(\triL+0.6pt,0)$) -- ([xshift=-5pt]gxE|-gy2);
        
          \coordinate (gy3) at ($(rect.south)!0.25!(rect.north)$);
          \coordinate (p3)  at ($(gxW|-gy3)$);
          \fill[white]
            ($(p3)+(-\triL,-\triH)$) --
            ($(p3)+(-\triL, \triH)$) --
            ($(p3)+(\triL,0)$) -- cycle;
          \draw[white, line width=\BORDER, line cap=round]
            ($(p3)+(\triL+0.6pt,0)$) -- (gxE|-gy3);
        \end{scope}

        \coordinate (-sqwest)   at (sq.west);
        \coordinate (-rectwest) at (rect.west);
        \coordinate (-recteast) at (rect.east);
        \coordinate (-east)     at (rect.east);
        \coordinate (-rectnorth) at (rect.north);
        \coordinate (-rectsouth) at (rectprmpt.south);
      \end{scope}
    }
  },
  pics/segmentintermediatetriangle/.default={mila-purple-1}{mila-purple-2}{mila-purple}
}

\tikzset{
  pics/segmentlegend/.style n args={3}{
    code={
      \begin{scope}[node distance=0pt]
        \begin{scope}[local bounding box=sq]
          \path[draw=none, pattern color=black!30]
            (\rectr,0) -- (\sww,0) -- (\sww,0.5\sz) -- (\rectr,0.5\sz)
            arc (90:180:\rectr) -- (0,\rectr) arc (180:270:\rectr) -- cycle;
          \draw[black!70, dashed, line width=0.8\BORDER]
            (\sww,0.5\sz) -- (\rectr,0.5\sz) arc (90:180:\rectr) -- (0,\rectr)
            arc (180:270:\rectr) -- (\rectr,0) -- (\sww,0);
          \path[use as bounding box] (0,0) rectangle (\sww,0.5\sz);
        \end{scope}

        \node[
          draw=none, minimum height=0.5\sz, minimum width=0.04\rectw,
          inner sep=0pt, outer sep=0pt, right=0pt of sq
        ] (rectprmpt) {};
        \draw[black!70, line width=0.8\BORDER, dashed]
          (rectprmpt.north west) -- (rectprmpt.north east)
          (rectprmpt.north east) -- (rectprmpt.south east)
          (rectprmpt.south east) -- (rectprmpt.south west);
        \node[right=1pt of rectprmpt] (txtprompt)
          {\footnotesize \sffamily \textcolor{black!50}{\textit{prompt}}};

        \node[
          draw=black!50, line width=1.3\BORDER,
          minimum height=0.5\sz, minimum width=0.25\rectw,
          inner sep=0pt, outer sep=0pt, right=5pt of txtprompt
        ] (rect) {};
        \node[right=1pt of rect]
          {\footnotesize \sffamily \textcolor{black!50}{\textit{response}}};
      \end{scope}
    }
  },
  pics/segmentlegend/.default={mila-purple-1}{mila-purple-2}{mila-purple}
}

\newcommand{\subfiglongCoT}{
\begin{tikzpicture}
  \path (0,0) pic (seg1) {segmentlongcot={delethink-blue!20}{delethink-blue}};
  \path[fill=black, fill opacity=0, draw opacity=0]
    ([xshift=-0.8cm, yshift=-0.5\sz]seg1-sqwest)
    rectangle ([yshift=0.5\sz]seg1-sqwest);
  \node[font=\bfseries, right=0pt of seg1-east] {$\texttt{</s>}$};

  \coordinate (mid23) at ($(seg1-recteast)!0.5!(seg1-sqwest)$);
  \node[anchor=center, fill=none, draw=none, inner xsep=8pt, minimum height=3cm] at (mid23) {};

  \draw[|-|, line width=0.5, draw=black!100, line cap=round]
    ([xshift=1pt, yshift=-25pt]seg1-sqwest) --
    node[midway, below=7pt, align=center, text width=10cm]
      {\footnotesize \sffamily \textcolor{black!90}{LongCoT reasoning}
       \textcolor{black!90}{\scriptsize (context grows linearly with more thinking tokens)}}
    ([xshift=1pt, yshift=-25pt]seg1-recteast);

  \node[anchor=south west]
    at ([xshift=5pt, yshift=-23pt]current bounding box.north west)
    {\sffamily \large \textbf{LongCoT Environment}
     \textcolor{black!60}{}};

  \coordinate (NW) at (current bounding box.north west);
  \draw[opacity=0, line width=0pt] (NW) -- ([xshift=16.5cm]NW);
\end{tikzpicture}
}

\newcommand{\subfigdelethink}{
\begin{tikzpicture}
  \path (0,0) pic (seg1) {segment={delethink-purple!10}{delethink-purple!10}};
  \path ([yshift=-0.5\sz, xshift=\segap]seg1-east)
    pic (seg2) {segmentintermediatecircle={delethink-purple!10}{delethink-purple!25}{delethink-purple}};
  \path ([yshift=-0.5\sz, xshift=1\segap]seg2-east)
    pic (seg3) {segmentintermediatetriangle={delethink-purple!40}{delethink-purple!55}{delethink-purple}};
  \path ([yshift=-0.5\sz, xshift=\segap]seg3-east)
    pic (seg4) {segmentintermediatesquare={delethink-purple!65}{delethink-purple!90}{delethink-dark-purple}};
  \node[font=\bfseries, right=0pt of seg4-east] {$\texttt{</s>}$};

  \draw[curvedarrow]
    ([yshift=1pt]seg1-rectnorth) to[out=40, in=-130, looseness=1.4] ([yshift=-1pt]seg2-rectsouth);
  \draw[curvedarrow]
    ([yshift=1pt]seg2-rectnorth) to[out=40, in=-130, looseness=1.4] ([yshift=-1pt]seg3-rectsouth);
  \draw[curvedarrow]
    ([yshift=1pt]seg3-rectnorth) to[out=40, in=-130, looseness=1.4] ([yshift=-1pt]seg4-rectsouth);

  \draw[arrowline, shorten >=2pt, shorten <=2pt]
    ([xshift=\segaparrow]seg1-recteast) -- ([xshift=-0.2\segaparrow]seg2-sqwest);

  \draw[arrowline, shorten >=2pt, shorten <=2pt]
    ([xshift=\segaparrow]seg2-recteast) -- node[midway, font=\large, fill=white] {\strut$\cdots$}
    ([xshift=-0.2\segaparrow]seg3-sqwest);

  \draw[arrowline, shorten >=2pt, shorten <=2pt]
    ([xshift=\segaparrow]seg3-recteast) -- ([xshift=-0.2\segaparrow]seg4-sqwest);

  \draw[|-|, line width=0.5, draw=black!100, line cap=round]
    ([xshift=1pt, yshift=-25pt]seg1-sqwest) --
    node[midway, below=7pt, align=center, text width=4cm]
      {\footnotesize \sffamily \textcolor{black!90}{Chunk $1$\\}}
    ([xshift=1pt, yshift=-25pt]seg1-recteast);

   \draw[|-|, line width=0.5, draw=black!100, line cap=round]
    ([xshift=1pt, yshift=-25pt]seg2-sqwest) --
    node[midway, below=7pt, align=center, text width=4cm]
      {\footnotesize \sffamily \textcolor{black!90}{Chunk $2$\\}}
    ([xshift=1pt, yshift=-25pt]seg2-recteast);

  \draw[|-|, line width=0.5, draw=black!100, line cap=round]
    ([xshift=1pt, yshift=-25pt]seg3-sqwest) --
    node[midway, below=7pt, align=center, text width=4cm]
      {\footnotesize \sffamily \textcolor{black!90}{Chunk $L-1$}}
    ([xshift=1pt, yshift=-25pt]seg3-recteast);
  
  \draw[|-|, line width=0.5, draw=black!100, line cap=round]
    ([xshift=1pt, yshift=-25pt]seg4-sqwest) --
    node[midway, below=7pt, align=center, text width=4cm]
      {\footnotesize \sffamily \textcolor{black!90}{Chunk $L$}}
    ([xshift=1pt, yshift=-25pt]seg4-recteast);

  \node[anchor=south]
    at ([yshift=-17pt]current bounding box.south)
    {\footnotesize \sffamily \textcolor{black!90}{\scriptsize (context size remains constant throughout)}};

  \path ([xshift=-6.5cm, yshift=-4.3\sz]seg4-sqwest)
    pic (seg5) {segmentlegend={mila-purple!75}{mila-purple!90}{mila-purple!100}};

  \node[anchor=south west]
    at ([xshift=5pt, yshift=-20pt]current bounding box.north west)
    {\sffamily \large \textbf{Delethink Environment} \textcolor{black!60}{\textit{(Markovian Thinking)}}};

  \coordinate (NW) at (current bounding box.north west);
  \draw[opacity=0, line width=0pt] (NW) -- ([xshift=16.5cm]NW);
\end{tikzpicture}
}

\newcommand{\figmethod}{
\begin{figure*}[t]
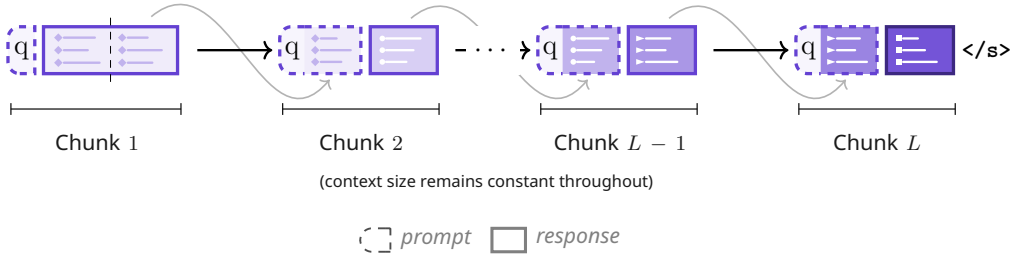

    \centering
    \begin{minipage}{0.9\textwidth}
      \centering
      \resizebox{\linewidth}{!}{\subfiglongCoT}\par
      \vspace{2mm}
      \vspace{0.5ex}
      \resizebox{\linewidth}{!}{\subfigdelethink}
    \end{minipage}
    \caption{
    \textbf{Delethink} redefines the thinking RL environment as a chunked, markovian process: generation proceeds in fixed-size chunks, and at each boundary the environment resets the context to a fresh prompt containing the query plus a short carryover from the previous chunk. This forces the policy to learn to progress across chunks by maintaining a textual state, creating a \emph{Markovian Thinker}. In contrast, the LongCoT environment concatenates tokens indefinitely, so the state (and model context) grows with the trace.
    }
    \label{fig:delethink_inference}
\end{figure*}

}

\maketitle

\begin{abstract}
Reinforcement learning (RL) has recently become a strong recipe for training reasoning LLMs that produce long chains of thought (LongCoT). Yet the standard RL “thinking environment’’, where the state is the prompt plus all prior reasoning tokens, makes the state unbounded and forces attention-based policies to pay quadratic compute as thoughts lengthen. 
We revisit the environment itself. We propose \emph{Markovian Thinking}, a paradigm in which the policy advances reasoning while conditioning on a constant-size state, decoupling thinking length from context size. As an immediate consequence this yields linear compute with constant memory.
We instantiate this idea with \emph{Delethink}, an RL environment that structures reasoning into fixed-size chunks. Within each chunk, the model thinks as usual; at the boundary, the environment resets the context and reinitializes the prompt with a short carryover.
Through RL, the policy learns to write a textual state near the end of each chunk sufficient for seamless continuation of reasoning after reset.
Trained in this environment, an R1-Distill 1.5B model reasons in 8K-token chunks yet thinks up to 24K tokens, matching or surpassing LongCoT-RL trained with a 24K budget. With test-time scaling, Delethink continues to improve where LongCoT plateaus. 
The effect of linear compute is substantial: we empirically estimate at 96K average thinking length LongCoT-RL costs 27 H100-months vs.\ 7 for Delethink.
Analysis at RL initialization shows off-the-shelf reasoning models (1.5B–120B) often sample Markovian traces zero-shot across diverse benchmarks, providing positive samples that make RL effective at scale.
Our results show that redesigning the thinking environment is a powerful lever: it enables very long reasoning without quadratic overhead and opens a path toward efficient, scalable reasoning LLMs.

\vspace{3.3mm}
{\small \sffamily \textbf{Model Weights:} \,\, \href{https://huggingface.co/collections/McGill-NLP/the-markovian-thinker-68debd2919c4ae47f50706cd}{\textbf{\texttt{huggingface.co/McGill-NLP/the-markovian-thinker}}}}

\vspace{-0.4em}
{\small\sffamily \textbf{Code Repository:} \href{https://github.com/McGill-NLP/the-markovian-thinker}{\textbf{\texttt{github.com/McGill-NLP/the-markovian-thinker}}}}

\end{abstract}

\section{Introduction}
Reinforcement learning (RL) has emerged as an effective recipe for training reasoning LLMs that produce a long chain-of-thought (LongCoT) before answering \citep{jaech2024o1systemcard, guo2025r1zeropaper, openai_gpt5_system_card_2025}; with scaling the number of “thinking tokens’’ improving capability \citep{Agarwal2025:GPT-OSS}. 
As in any RL problem, there is an environment that determines how trajectories are generated.
For reasoning LLMs this environment is rather trivial that it's often omitted: the \emph{state} is the prompt concatenated with generated reasoning tokens so far, and the \emph{action} is the next token sampled from the policy (reasoning LLM). 
This seemingly innocuous choice leaves the state size unbounded, growing with longer thoughts. 
For attention-based policies this entails prohibitive, \emph{quadratic growth} of compute throughout.

Most recent work tames this compute growth by constraining how much thinking occurs during RL:
multi-stage training limits how often long reasoning traces are used \citep{deepscaler2025}; length-regularized objectives implicitly prefer shorter correct solutions \citep{aggarwal2025L1, shen2025dast, li2025adaptive, hou2025thinkprune}; and pruning or early-exit methods terminate generation while preserving accuracy \citep{luo2025o1pruner, dai202earlyexitsgrpo, zhao2025letllmbreakfree}. 
In contrast, we ask a different question: \emph{what if the environment never creates quadratic growth to begin with?}

We propose a paradigm in which the policy advances its reasoning while conditioning on a \emph{constant-size} state. We call such a policy a \emph{Markovian Thinker}. 
The key idea is to restructure the RL formulation so the effective state that the policy reads is bounded, regardless of total thinking length. 
The immediate consequence is profound: longer thinking requires linear compute and constant memory with respect to thinking length, decoupling “how long the model thinks’’ from “how much context it must process.’’

\figmethod

We instantiate this paradigm with Delethink, an RL environment that induces Markovian behavior by organizing reasoning into a sequence of fixed-size chunks (\Cref{fig:delethink_inference}). Within each chunk, the model reasons as usual. 
At chunk boundaries, the environment \emph{resets the context} and reinitializes the prompt using only a short carryover from the previous chunk (e.g., the last few tokens). 
The policy is forced to learn, through RL, to write a textual \emph{state} near the end of each chunk so that, after reset, it can continue the thread of reasoning while conditioning only on this bounded state.

Delethink is highly effective.
Despite reasoning in 8K chunks, R1-Distill 1.5B \citep{guo2025r1zeropaper} trained with Delethink can think up to 24K tokens, matching and surpassing LongCoT-RL on math benchmarks with the same 24K thinking budget (\Cref{fig:final_teaser_figure,fig:task_performance}). 
With test-time scaling, Delethink continues to improve where LongCoT-RL plateaus, yielding additional gains (\Cref{fig:final_teaser_figure,fig:extended_thinking}). 
Pushing further, we train R1-Distill 1.5B with Delethink to think \emph{up to} 96K tokens; with only few additional training steps, it reaches 49\% on AIME’24 with solutions averaging 36K tokens. 
The effect of linear-compute is substantial: 
We empirically estimate that training for an average of 94K thinking requires 27 H100-months under LongCoT-RL versus 7 H100-months with Delethink (\Cref{fig:final_teaser_figure}).

To probe \emph{why} Delethink trains effectively, we analyze models at RL initialization. 
We observe that the R1-Distill family (1.5B-14B), without additional training or prompting, already samples Markovian traces \emph{zero-shot}, even recovering most of standard LongCoT performance (\Cref{fig:offtheshelf_r1_bench}). 
This strong initialization (many positive, in-distribution samples of the desired behavior) provides a favorable starting point for RL. 
We further study reasoning models up to 120B parameters in the Delethink environment. For example, GPT-OSS 120B \citep{Agarwal2025:GPT-OSS} exhibits robust Markovian Thinking across PhD-level questions, coding tasks, math competitions, and crossword puzzles (\Cref{sec:why_works}).
Together, these results suggest that Delethink is compatible and scales with state-of-the-art models.

The success of Markovian Thinking demonstrates that decoupling thinking length from context size can, in principle, let next-generation reasoning models think for millions of tokens. 
It highlights the RL environment, often treated as fixed, as a powerful lever for progress.
It also suggests that non-quadratic sequence architectures \citep{katharopoulos2020linear, yang2023gated-linear-attention, gu2023mamba} may particularly benefit reasoning models, since thinking can be made effectively Markovian. We summarize our contributions as follows:

\begin{figure}[!t]
    \centering
    \includegraphics[width=\textwidth]{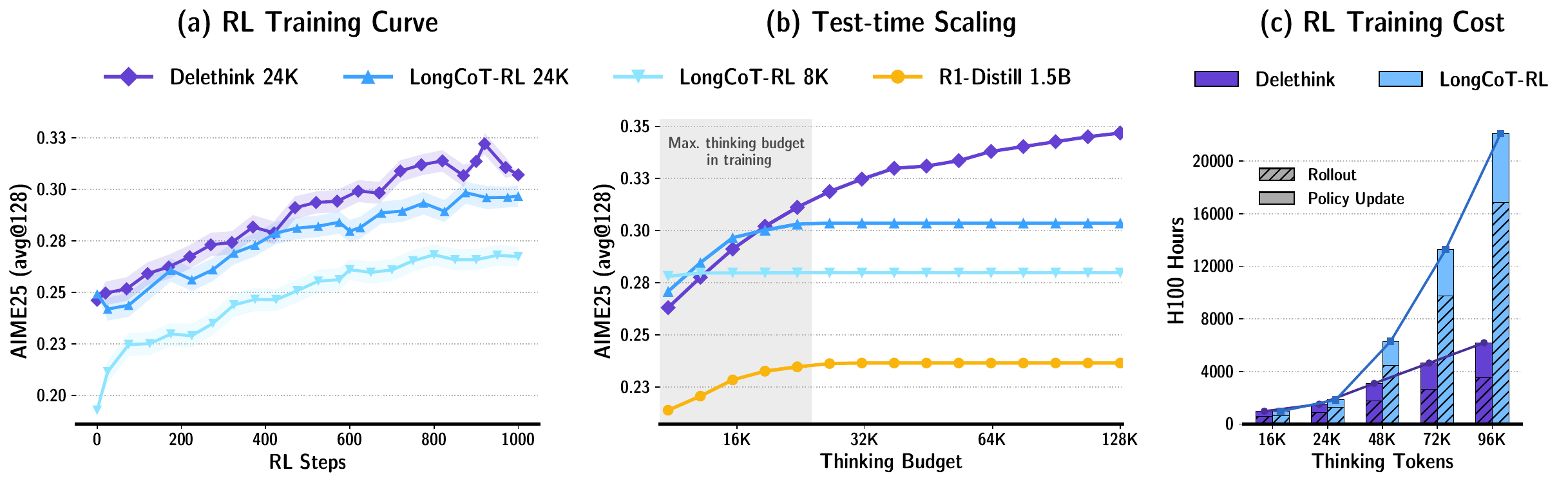} %
    \caption{
    \textbf{(a)} Delethink 24K (a Markovian Thinker) matches and surpasses LongCoT-RL 24K in accuracy during RL training while using less compute; both methods improve as the thinking budget scales from 8K to 24K.
    \textbf{(b)} Beyond the trained thinking budget, Delethink significantly outperforms and keeps improving while others plateau; within the budget, Delethink and LongCoT-RL 24K scale similarly with test-time compute (reported using sequential sampling).
    \textbf{(c)} Training cost of R1-Distill 1.5B vs. average thinking length with an optimized stack of verl \citep{sheng2024verl} + SGLang \citep{zheng2023sglang} on H100s: quadratic for LongCoT and linear for Delethink, as predicted.
    }
    \label{fig:final_teaser_figure}
\end{figure} 

\begin{itemize}[leftmargin=*]
\item We introduce the Markovian Thinking paradigm that bounds the policy’s input state and yields linear compute and constant memory with thinking length (\Cref{sec:compute_scaling}). 
\item We present a simple instantiation, Delethink (\Cref{sec:delethink}), that trains off-the-shelf reasoning LLMs to be native Markovian Thinkers, empirically matching and surpassing LongCoT-RL training (\Cref{sec:results,sec:empirical_compute}).
\item We show that Delethink test-time scale well beyond train-time limits whereas LongCoT plateaus (\Cref{sec:delethink_test_time_scales}). 
\item We provide empirical evidence that large SOTA reasoning LLMs has strong support for Markovian Thinking in-distribution, signaling that Delethink scales.(\Cref{sec:why_works}).
\end{itemize}

\section{Related Works}
\paragraph{Efficient RL training of LongCoT}
Prior works try to reduce the cost of RL training under LongCoT. \citet{liu2025drgrpo} stabilizes GRPO to avoid wasted length. Others train models to prune or exit early while maintaining accuracy \citep{luo2025o1pruner, dai202earlyexitsgrpo, zhao2025letllmbreakfree}, or implicitly reward shorter thoughts among correct solutions \citep{aggarwal2025L1, shen2025dast, li2025adaptive, hou2025thinkprune}. Additionally, \citet{deepscaler2025} limits the fraction of training with long thoughts, an approach known as multi-stage RL training although \citet{zeng2025glm-multi-stage-fails} shows it hurts final performance. While lighter, these approaches operate under LongCoT and therefore are quadratic. 

\paragraph{Inference Time Efficient Thinking}
Prior works seek efficiency by shortening reasoning traces. Some distill traces with skipped steps or tokens \citep{liu2024canllmlearntoskip, xia2025tokenskip}. Others control length by early exits \citep{ding2024break}, adjusting the budget to problem complexity \citep{han2024tokenbudgetaware}, or steering activations toward brevity \citep{zhao2025exploring}. Structured prompting and collaboration help reduce tokens: CoThinking outlines then reasons \citep{fan2025cothink}. \citet{cheng2024compressed} thinks in shorter trace of contemplative tokens. 
Some other works approximate the attention at test-time by changing its computation and masking tokens. \citet{zhang2023h2o,yang2024doublesparse} select important tokens based on their estimated contribution to attention, while \citet{xiao2023streamingllm} preserves a small set of attention-sink tokens to stabilize quality under sliding windows. Compression approaches shrink each retained token’s footprint via quantization to sub-4-bit while maintaining accuracy \cite{hooper2024kvquant,liu2024kivi}. Recent approaches use exisiting parameters to learn to evict tokens in inference using distillation on the original model predictions plus a sparsity regularizer~\citep{ancucki2025inferencetimehyperscalingkvcache}. These methods are used to ensure faster inference time after RL training. 

\label{sec:delethink}
\begin{algorithm}[t]
\caption{Delethink Step}
\label{alg:Delethink_training}
\begin{algorithmic}
\State \textbf{Inputs:} query $\vq$; reasoning LLM $\pi_\theta$; thinking context size $\mathcal{C}$; markovian state size $m$; Delethink iterations cap $\mathcal{I}$; group size $G$;  reward function $\mathcal{R}$; learning rate $\eta$.
\State $T,R \gets [\,],\,[\,]$ \Comment{Delethink traces, rewards}
\State \textbf{Generate Delethink Traces:}
\For{$i \gets 1$ to $G$}
  \State $\vx \gets \vq$ \Comment{prompt}
  \State $\vy \gets \pi_\theta(\vx; \mathcal{C})$ \Comment{generate up to $\mathcal{C}$ tokens}
  \State $\vq \gets \vq \oplus \vy_{1:100}$ \Comment{concatenate first few tokens of $\vy$}
  \State $\tau \gets [(\vx, \vy)]$ \Comment{trace for group $i$}

  \For{$t \gets 1$ to $\mathcal{I} - 1$}
    \If{$\operatorname{last}(\vy)=\texttt{[EOS]}$} \textbf{break} \EndIf
    \State $\vx \gets \vq \oplus \vy_{-m:}$ \Comment{keep last $m$ thinking tokens}
    \State $\vy \gets \pi_\theta(\vx, \mathcal{C}-m)$ \Comment{generate up to $\mathcal{C} - m$ tokens}
    \State $\mathrm{append}(\tau, (\vx, \vy))$ \Comment{appending chunk to trace} 
  \EndFor
  
  \State $\mathrm{append}(R,\, \mathcal{R}(\tau))$ \Comment{trace reward}
  \State $\mathrm{append}(T,\, \tau)$
\EndFor

\State \textbf{Estimate Advantages:} 
\State $\{\hat A[i]\}_{i=1}^G \gets \mathrm{ComputeAdvantage}\!\left(\{R[i]\}_{i=1}^G\right )$ \Comment{off-the-shelf advantage estimator}

\State \textbf{Updating Parameters:} 
\State $J \gets \frac{1}{G} \sum_{\tau_g}
  \frac{1}{\ell(\tau_g)} \sum_{l=1}^{|\tau_g|}\mathcal{U}(\vx_l, \vy_l; \theta)$ \Comment{Compute the loss according to \Cref{eq:delethink_loss_1}}

\State $\theta \gets \theta + \eta\,\nabla_{\theta}\,J$  
\end{algorithmic}
\end{algorithm}

\paragraph{Making Room for Extra Thinking}
Prior work has tried to shorten reasoning traces at inference time. \citet{lin2025trimrverifierbasedtrainingfreethinking} drops irrelevant tokens and redundant steps using an extra judge LLM. \citet{xiao2025limopro} prunes low-value segments by perplexity. \citet{Yan2025InftyThinkBT} iteratively summarizes current reasoning so later steps continue depending only on the summary. InftyThink’s style is fixed by design, distilled from a hand-crafted dataset that locks the model into one pattern. In contrast, Delethink is an RL training approach that learns native Markovian Thinking through RL.

\paragraph{Linear Architectures}
Sliding or sparse attention \citep{beltagy2020longformer,zaheer2020bigbird} and kernel-based linear attention and low-rank linear attention \citep{katharopoulos2020linear,choromanski2021performer,wang2020linformer} avoid all-pairs interactions, the quadratic component of attention, by approximations. Mamba architecture replaces self-attention with state-space models \citep{gu2021s4,gu2023mamba,dao2024transformersssm}, achieving constant-memory, linear-time generation via recurrent state updates. Despite progress in this field, all state-of-the-art models \citep{Agarwal2025:GPT-OSS, gemini2024report, qwen3, guo2025r1zeropaper} include self-attention even the  hybrid systems only interleave self-attention with alternatives. All scale asymptotically quadratic \citep{lieber2024jamba,jamba2024v15}. Delethink does not modify the architecture and is not quadratic in context. It decouples reasoning length and context length, avoiding the quadratic cost. However, Delethink shows continuing reasoning effectively with constant number of tokens is possible, providing evidence that these methods might be more successful in reasoning than other areas.

\section{Background}
We adopt the RLVR formulation, following \citet{lambert2024:tulu3,kazemnejad2025:vineppo}. 
In this setup, the policy $\pi_\theta$ is a reasoning language model that, given a prompt $\vx = [x_1,\ldots,x_N]$ (which may include the system prompt, chat history, and a user query), generates a response $\vy = [y_1,\ldots,y_T]$ autoregressively: $\vy \sim \pi_\theta(\cdot \mid \vx)$. 
The goal of RL training is to maximize expected return while regularizing the policy via a KL penalty against a reference model.
Formally, the objective is
\begin{equation}
\label{eq:rlvr}
\mathcal{J}(\theta) = \mathbb{E}_{\vx \sim \mathcal{D}, \vy \sim \pi_\theta(\cdot|\vx)}\left[ \mathcal{R}(\vx, \vy) \right] - \beta \, \mathrm{KL}[\pi_\theta \| \pi_\text{ref}],
\end{equation}
where $\mathcal{D}$ is a dataset of queries, $\pi_\text{ref}$ is the reference policy, $\beta$ is the KL coefficient, and $\mathcal{R}(\vx,\vy)$ is a trajectory-level reward function (verifiable rewards in our case). 
The policy $\pi_\theta$ is typically initialized from $\pi_\text{ref}$.
Then, $\mathcal{J}(\theta)$ is optimized by sampling trajectories and applying policy-gradient updates \citep{shao2024deepseekmath}.

\begin{figure}[ht]
\centering
\begin{minipage}[t]{0.85\textwidth}
\centering
\vspace*{0cm}     %

\resizebox{\linewidth}{!}{%
\begin{tabular}{l@{\hspace{7em}}ccccc}
\toprule
\textbf{Method} & \textbf{Think Tokens} & \textbf{FLOP} & \textbf{Memory} & \textbf{Backward Time} & \textbf{Generation Time} \\
\midrule
Base      & $n$   & $O(n^{2})$       & $O(n)$   & $T_B$       & $T_G$ \\
LongCoT   & $nS$  & $O(n^{2} S^{2})$ & $O(nS)$  & $O(T_B S^2)$& $O(T_G S^2)$ \\
Markovian Thinking (Delethink) & $nS$  & $O(n^{2} S)$     & $O(n)$   & $O(T_B S)$  & $O(T_G S)$ \\
\bottomrule
\end{tabular}
}

\end{minipage}

\begin{minipage}[t]{0.85\textwidth}
\vspace*{0.3cm} 
\centering
\includegraphics[width=\linewidth]{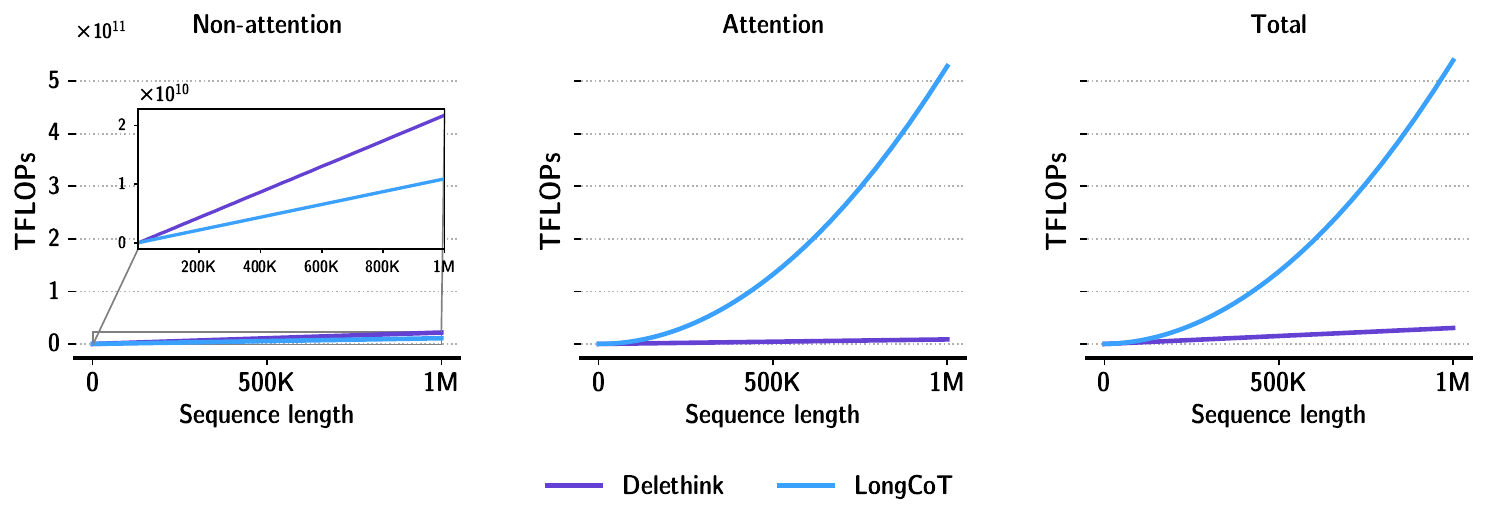}
\label{fig:flop_theory_comparison}
\end{minipage}
\caption{Computational profiles of LongCoT-RL and Delethink scaling from $n$ to $nS$ tokens.}
\label{tab:compute_memory}
\end{figure}

\paragraph{Thinking RL environment (MDP)}
\Cref{eq:rlvr} is defined over trajectories induced by an underlying language-generation Markovian Decision Process (MDP). This MDP is straightforward and typically left implicit, yet its design choice bears fundamental consequences.
Concretely, token-by-token generation MDP (as used in standard LongCoT RLVR) specifies the following states $\mathcal{S}$, actions $\mathcal{A}$, and transition dynamics.
At time $t$, the state $s_t \in \mathcal{S}$ is the concatenation of the prompt and the tokens generated so far:
$
s_t = \vx \oplus \vy_{<t} = [x_1,\ldots,x_N,\, y_1,\ldots,y_{t-1}],
$
where $\oplus$ denotes concatenation.
The action $a_t \in \mathcal{A}$ corresponds to sampling the next token $y_t$ from the policy given the generation context.
The process starts at $s_0 = \vx$ (where $\vx$ is the prompt), and after each action, the environment deterministically transitions to the next state, $s_{t+1}$, by appending the action $a_t$ to the current state $s_t$:
\begin{equation}
\label{eq:mdp_transition}
P(s_{t+1}=s' \mid s_t=s,\, a_t=a) \;=\;
\begin{cases}
1, & \text{if } s' = s \oplus [a],\\
0, & \text{otherwise.}
\end{cases}
\end{equation}
Policies trained in this environment learn to think so as to maximize reward (e.g., solve math problems correctly).
Although such thinking environment satisfies the Markov property (i.e. future only depends on the current state), the state size is unbounded: $|s_t|=O(t)$ as actions are appended. Consequently, the final token of a response $\vy$ is generated with context size $|s_T|=|\vx|+|\vy|-1$. 
For attention-based policies (e.g., transformer LLMs), this occurs a quadratic computation cost with respect to thinking tokens throughout the training and inference.

\section{Delethink}

Delethink reformulates the standard RLVR by changing the underlying RL environment (MDP) that induces trajectories.
The idea is to train policies to be native \emph{Markovian Thinkers}: instead of accumulating one ever-growing chain of thought, the policy is forced to reason in \emph{a sequence of fixed-size chunks}.

Within chunks, the model thinks as usual, But, at the end of each chunk, the environment resets the context to a fresh prompt seeded with a short carryover from the previous chunk (e.g., the last $m$ tokens).
Progress therefore depends on what the policy chooses to write into this carryover---its \emph{textual Markovian state}---so it learns to advance reasoning while conditioning only on a bounded state.

These dynamics impose a hard cap on state size: at every step, the environment enforces $|s_t| = O(\mathcal{C})$ for a fixed per-chunk budget $\mathcal{C}$ (e.g., $8\textnormal{K}$ tokens).
The policy's effective context is therefore independent of the total reasoning length.
For attention-based policies, this turns quadratic compute in the (growing) context length into compute that is quadratic only in the constant chunk size $\mathcal{C}$; consequently, the overall training cost scales linearly with the number of thinking tokens because the full trace never materializes in context at once. 
We formalize this compute scaling in \Cref{sec:compute_scaling} and compare against standard LongCoT RL. We next detail trajectory generation under this chunked dynamics and derive the corresponding policy-gradient estimator, completing the RL training loop.

\begin{figure}[!t]
    \centering
    \includegraphics[width=\textwidth]{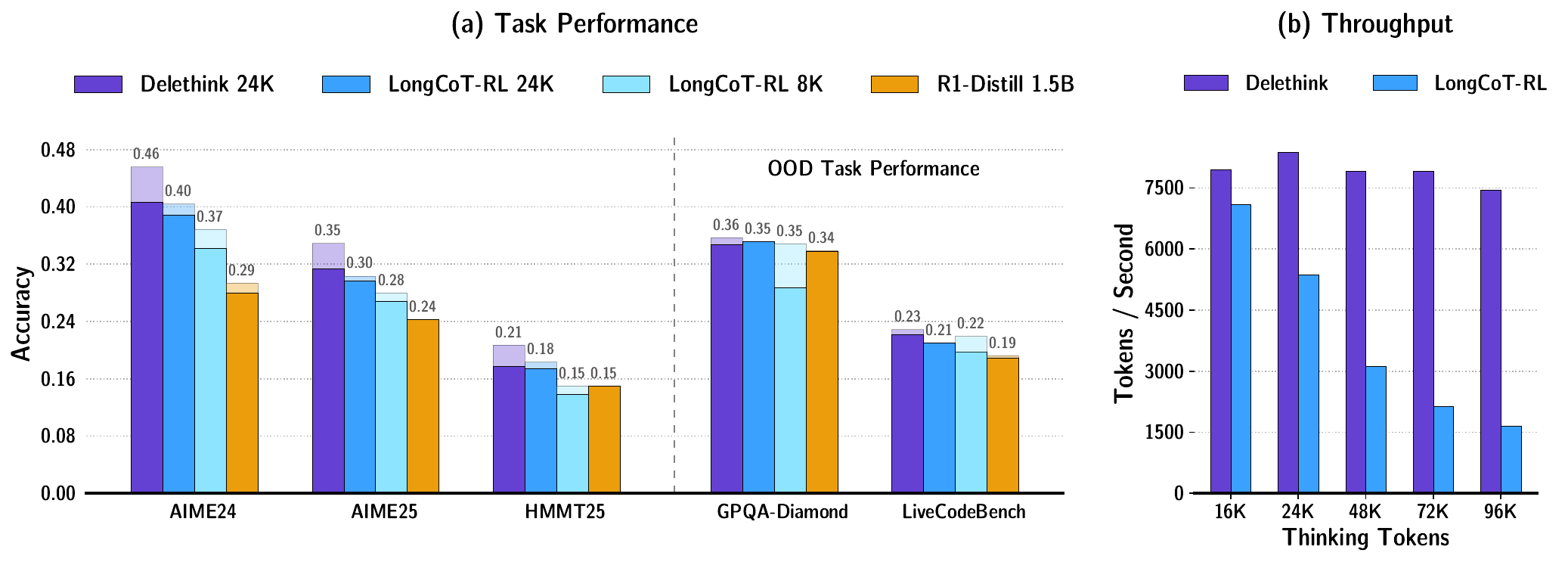} %
    \caption{
    \textbf{(Left)} On IID math tasks (AIME’24/’25, HMMT’25), Delethink outperforms LongCoT-RL 24K. Shaded regions show gains from test-time scaling (through sequential sampling), where Delethink improves the performance even more; on OOD tasks (GPQA-Diamond, LiveCodeBench) gains are modest, yet Delethink still matches or slightly beats LongCoT-RL 24K.
    \textbf{(Right)}
    Per-GPU rollout throughput during RL training (R1-Distill 1.5B, H100 cluster). Delethink’s RL environment design keeps peak memory constant, sustaining throughput as thinking scales; LongCoT’s memory grows linearly, driving throughput down at longer budgets.
    }
    \label{fig:1000rlstep}
    \label{fig:task_performance}
\end{figure}

\subsection{Delethink Tracing}
\label{sec:DI}
A policy trained under Delethink reasons in a sequence of chunks.
We start by explaining trajectory generation process and then discuss how it effect  transition dynamics, followed by practical implications for LLM inference.

Let $\vq$ denote the query, typically containing a system prompt, chat history, and the user inquiry.
In the first chunk, the model generates a response up to $\mathcal{C}$ tokens as usual with the input query as prompt: $\vy_1 \sim \pi(\cdot | \vx_1 = \vq)$, where $\vx_1$ and $\vy_1$ represent the first chunk's prompt and response, respectively. 
If the response $\vy_1$ ends with \texttt{\small [EOS]}, the trace is complete.
Otherwise, Delethink advances by constructing the next prompt from the original query and a \emph{markovian state} consisting of the last $m < \mathcal{C}$ tokens of the previous chunk output\footnote{In practice, we fold the first hundred tokens of the initial chunk into $\vq$ as it may contain planning tokens.}:
\[
\vx_l \;=\; \vq\, \oplus\,\vy_{(l-1)}[-m:]\,, \quad l \ge 2,
\]
where $\oplus$ denotes concatenation and $\vx_l$ and $\vy_l$ is the prompt and response for chunk $l$, respectively. 
In effect, preceding reasoning tokens are deleted when forming the next prompt (hence the name Delethink).
Given $\vx_l$, the model generates up to $\mathcal{C}-m$ new thinking tokens for chunk $l$: $\vy_l \sim \pi(\cdot|\vx_l)$.
This procedure repeats until \texttt{\small [EOS]} is produced or the iteration cap $\mathcal{I}$ is reached.
We refer to the resulting sequence as a \emph{Delethink trace}
$\tau \;=\; \left[\,(\vx_1, \vy_1), \ldots, (\vx_L, \vy_L)\,\right],$ where $L$ is number of chunks in the Delethink trace $\tau$.
We illustrate this process in \Cref{fig:delethink_inference}.\footnote{We provide a sample implementation based on HuggingFace model in \Cref{fig:delethink-hf}}

This process for sampling trajectories can be seen as modifying the transition dynamics of the environment MDP at the boundary of chunks.
Let $B \subseteq \mathcal{S} \times \mathcal{A}$ be a set of boundary state-action pairs, where $\mathcal{S}$ is the set of states and $\mathcal{A}$ is a set of all actions.
Delethink redefines the transition dynamics as follows:
\[
P\!\left(s_{t+1}=s' \mid s_t=s,\, a_t=a\right)
=
\begin{cases}
1, & (s,a)\in B \ \text{and}\ s' = s_{:|\vq|}\oplus s_{-m:} \oplus a,\\[3pt]
1, & (s,a)\notin B \ \text{and}\ s' = s \oplus a,\\[3pt]
0, & \text{otherwise.}
\end{cases}
\]

In total, the LLM may think up to $\mathcal{C} + (\mathcal{I}-1)\,(\mathcal{C}-m)$ tokens under Delethink environment, while each chunk's context remains bounded by $|\vq|+\mathcal{C}$. Since query remains constant through the Delethink process and $|\vq| \ll \mathcal{C}$ in practice, the maximum per-chunk model context is $O(\mathcal{C}) = O(1)$ with respect to the total number of thinking tokens.
In the paper, we fix $m=\mathcal{C}/2$; \Cref{app:markovian_state_size_ablation} ablates the markovian state size.

\paragraph{KV cache under Delethink.}
Within each chunk, decoding uses the KV cache exactly as in standard transformer inference: generating $\vy_l$ reuses keys/values from the prefix of the same chunk. 
At the chunk boundary, the environment resets the context and starts a new prompt $\vx_{l+1}$,
which includes $\vy_l[-m:]$.
Intuitively, these tokens are no longer conditioned on $\vx_l$; they are recontextualized under $\vq$. 
Therefore, the KV cache must be cleared at this point and the carryover tokens will be re-encoded as part of the new prefix. 
This adds an $O(m^2)$ encoding cost at each boundary, which 
does not change compute complexity of Delethink.
Moreover, the prefill operation is quite fast compared to decoding in modern inference engine \citep{zheng2023sglang}.

\begin{figure}[!t]
    \centering
    \includegraphics[width=\textwidth]{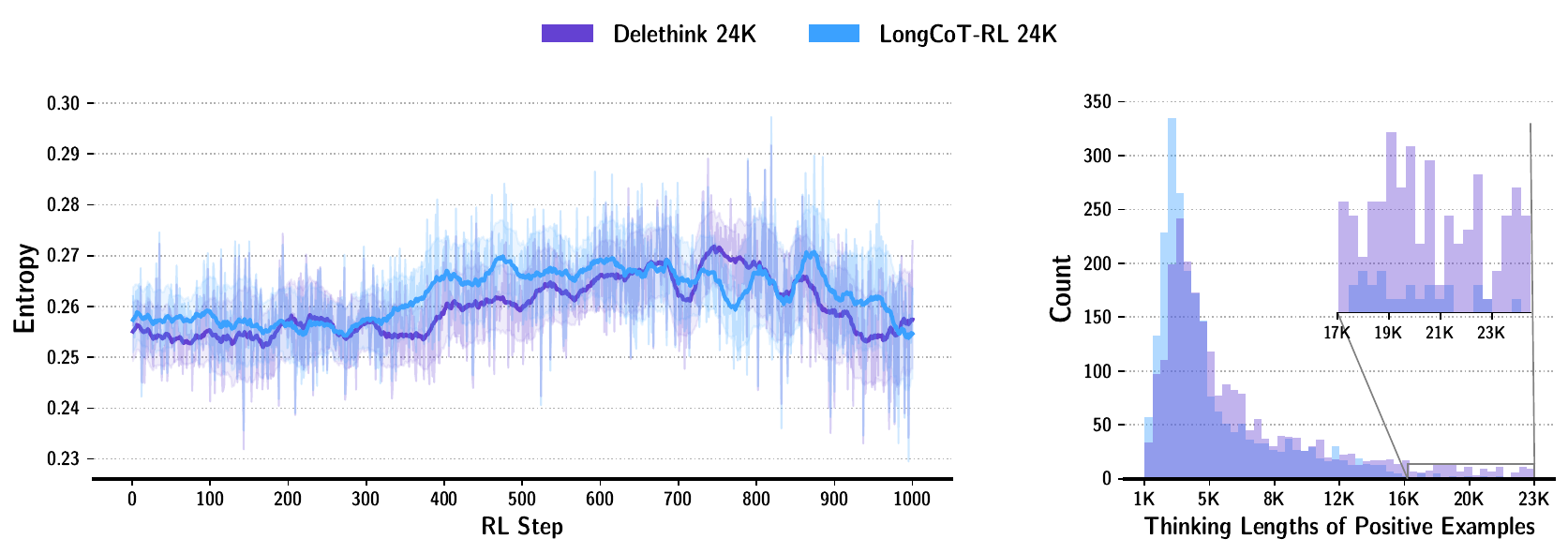} %
    \caption{
    \textbf{(Left)} Smoothed entropy over RL steps for Delethink and LongCoT-RL. Both remain roughly flat and non-collapsing \citep{cui2025entropy}, indicating stable learning. Note that rising entropy typically precedes divergence.
    (\textbf{Right})
    Delethink and LongCoT use their thinking budgets well. At longer lengths, Delethink produces more correct answers, showing it spends its budget effectively.
    }
    \label{fig:1000rlstep_entropy}
\end{figure}

\subsection{Policy Gradient}
We start with the RL objective, maximizing the expected reward:
$$
\mathcal{J}(\theta) = \mathbb{E}_{\vq \sim \mathcal{D}, \tau \sim \pi_\theta(\cdot|\vq)}\left[ \mathcal{R}(\tau) \right] - \beta \, \mathrm{KL}[\pi_\theta \| \pi_\text{ref}],
$$
where $\tau$ is a generated trace using under Delethink RL environment (\Cref{sec:DI}).
Optimizing this objective requires re-derivation of policy gradient as standard PPO-based estimators are derived under LongCoT MDP.
Therefore, we derive the policy gradient estimator under Delethink dynamics in \Cref{app:deriving_Delethink_loss}. 
The resulting objective closely mirrors the policy gradient form of standard RL training for LLMs \citep{lambert2024:tulu3}, allowing an easy adaptation of existing infrastructure for Delethink.
Intuitively, the policy gradient objective under Delethink environment sums over all the chunks in the trace, as each contributed to the final reward. 

Given a query $\vq$, we sample a group $G$ of Delethink traces $\tau_1, \ldots, \tau_G$ from the current policy $\pi_{\theta_\text{old}}$, where each trace is a sequence of chunks $\tau = \{(\vx_l, \vy_l)\}_{l=1}^{L}$ with $L$ denoting the number of chunks in the trace. 
Optimizing the expected return proceeds by taking gradients with respect to the following objective function:
\begin{equation}
\label{eq:delethink_loss_1}
J(\theta) = \mathbb{E}_{\tau_1, \ldots, \tau_G \sim \pi_{\theta_{\text{old}}}(.|\vq)} \Bigg[
  \frac{1}{G} \sum_{g=1}^{G}
  \underbrace{\frac{1}{\ell(\tau_g)} \sum_{l=1}^{L}\mathcal{U}(\vx_l, \vy_l; \theta)}_{\text{per-Delethink trace loss}}
\Bigg],
\end{equation}
 where $\ell(\tau) = \sum_l |\vy_l|$ is the total number of response tokens in a Delethink trace $\tau$. This is similar to GRPO, where per-trace term is normalized by the total number of thinking tokens. 
$\mathcal{U}(\vx, \vy; \theta)$ represents the per-chunk $(\vx, \vy)$ objective which closely follow that of PPO in LLMs. Specifically,
\begin{equation*}
  \label{eq:delethink_loss}
  \mathcal{U}(\vx, \vy, \theta) = \sum_{t=1}^{|\vy|} \min \left[ \frac{\pi_\theta(y_t)}{\pi_{\theta_{\text{old}}}(y_t)} \hat{A_t}, \text{clip}\left( \frac{\pi_\theta(y_t)}{\pi_{\theta_{\text{old}}}(y_t)}, 1-\epsilon, 1+\epsilon\right)  \hat{A_t} \right] - \beta \mathrm{KL}[\pi||\pi_\text{ref}],
\end{equation*}
where $\pi(y_t)$ is the probability of predicting token $y_t$ of the chunk's response $\vy$\footnote{We omit conditioning on the context $\vx, \vy_{<t}$ for brevity.} and $\beta$ controls the Kullback–Leibler term \citep{shao2024deepseekmath}.
The advantage $\hat{A_t}$ can be estimated with any off-the-shelf estimator \citep{kazemnejad2025:vineppo, yu2025dapo}. For simplicity we use the GRPO formulation:
$A_{l,t} \equiv A_{\tau_g} \coloneqq (\mathcal{R}(\tau_g)-\mu) / \sigma$, where $\mathcal{R}(\tau_g)$ is the reward for $g$-th trace (e.g. whether model reaches the correct answer at the end of last chunk). $\mu$ and $\sigma$ are the mean and standard deviation of rewards across the trace group $\{\tau_g\}$.
Pseudo-code is shown in \Cref{alg:Delethink_training}, illustrating training on a single query (we optimize over batches in practice).

\begin{figure}[!t]
    \centering
    \includegraphics[width=1\textwidth]{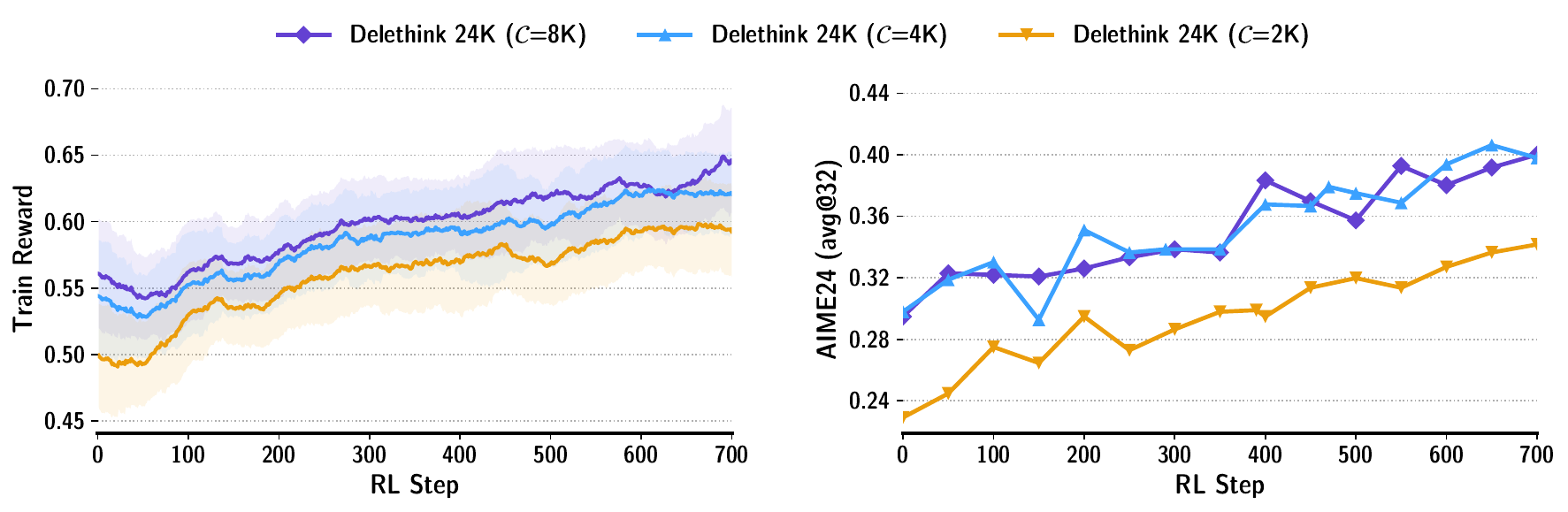} %
    \caption{
    We vary $\mathcal{C}$ while holding the total thinking budget at $\approx\text{24K}$ tokens, so smaller $\mathcal{C}$ implies a smaller Markovian state. \textbf{(Left)} smoothed training reward vs. RL step. \textbf{(Right)} validation accuracy (AIME’24) vs. RL step. Delethink 24K with $\mathcal{C}=8\text{K}$ and $4\text{K}$ performs similarly on both training reward and validation score, whereas 2K trails but improves steadily over the base model during Delethink RL.
    }
    \label{fig:rl_context_ablation}
\end{figure}

\subsection{Computational Cost of Scaling Thinking}

\label{sec:compute_scaling}
An RL step has two parts: generation and backward for updating the policy. We study how both scale when an LLM’s thinking length grows from $n$ tokens to $nS$ tokens under LongCoT and Delethink. As Table~\ref{tab:compute_memory} shows, both stages scale linearly in Delethink but quadratically in LongCoT.

\paragraph{Total FLOPs and Backward Time}  
Suppose training an LLM to think for $n$ tokens costs $O(n^2)$ FLOPs. With LongCoT, scaling by $S$ will cost $O(n^2S^2)$ because self-attention grows quadratically with length. Delethink instead runs in $2S$ chunks. Each chunk carries forward $\tfrac{n}{2}$ tokens and generates $\tfrac{n}{2}$ new ones (i.e., $m=\mathcal{C}/2$). The result is $O(2n^2S)=O(n^2S)$ FLOPs, linear in $S$. A factor of two arises because each token is both generated once and then reprocessed as the next chunk’s prompt. These theoretical insights mirror the exact computed FLOPs for our experimental setup shown in \Cref{tab:compute_memory} based on R1Distill 1.5B architecture; in particular, since the compute of non-attention layers is linear in tokens, the two factor is evident.

\paragraph{Peak memory}  
KV Cache entries have fixed size per token. In LongCoT, the KV cache grows linearly with thinking length, limiting parallel requests on the GPU. For example, the KV cache of a 1M-token trace on R1-Distill 1.5B, a small model, alone fills an entire H100. Going beyond requires sharding the sequence across GPUs with sequence-parallelism, adding heavy communication. In contrast, Delethink keeps only the current chunk’s KV cache, so usage stays constant. 

\paragraph{Generation Time}  
Assume an infinite stream of requests saturating the GPU, each with maximum length $L$. The optimal throughput satisfies $T \propto \tfrac{1}{L}$ under simplifying assumptions (\citealp{ao2025fluid-guided}; see \ref{app:fluid_guided}). This can be intuited from the fact that generation must access the KV cache, which grows linearly with sequence length, limiting parallel requests. LongCoT uses a growing KV cache; Delethink keeps it fixed per chunk. Both must generate $nS$ tokens, but LongCoT’s throughput falls by a factor of $S$, scaling time by $S^2$. Delethink’s throughput is constant, so its time scales only as $S$.  

\paragraph{Effect of $\mathcal{C}$ and $m$}  
The analysis above takes $\mathcal{C}=n$ and $m=\tfrac{n}{2}$ as the baseline (our training setup). However, $\mathcal{C}$ and $m$ are hyperparameters of Delethink. Relative to this baseline: if Delethink thinks for $nS$ tokens using $S$ chunks of length $2n$ (i.e., $\mathcal{C}=2n$), the total FLOPs become $O(4n^2S)$, still linear in $S$ but with a higher constant. Alternatively, if it thinks for $nS$ tokens in $\tfrac{4}{3}S$ chunks with a context size of $\mathcal{C}=n$ and a smaller state $m=\tfrac{n}{4}$, the total FLOPs are $O(\tfrac{4}{3}n^2S)$, again linear in $S$ but with a lower constant.

\begin{figure}[!t]
  \centering
  \includegraphics[width=1\textwidth]{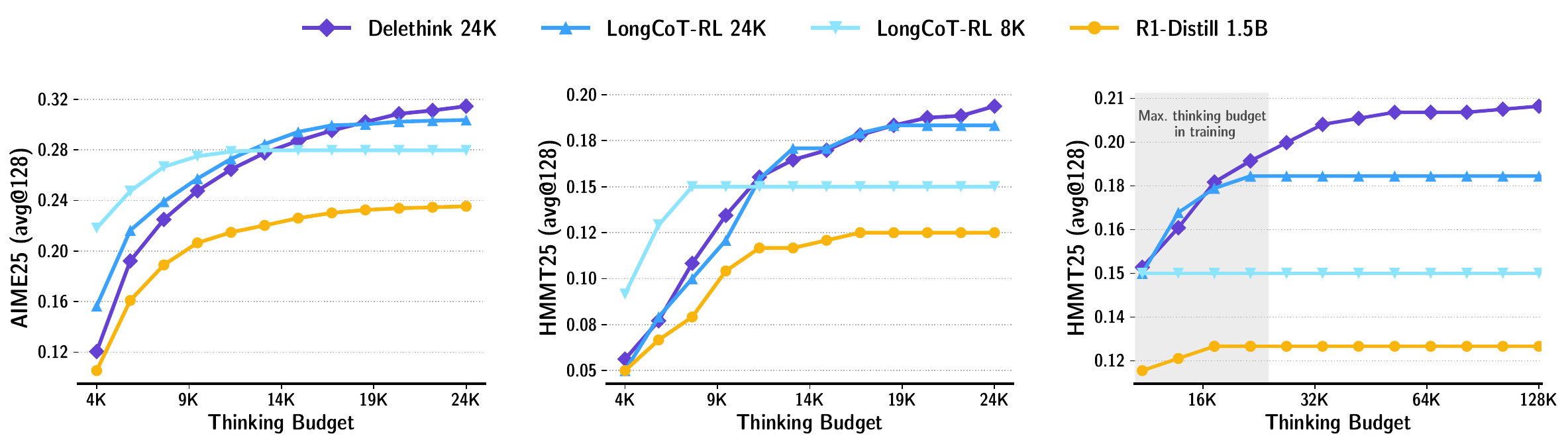}
  \caption{
    \textbf{Scaling the Thinking Budget at Inference.} Within the train-time budget, Delethink 24K and LongCoT-RL both gain accuracy as thinking tokens increase. Beyond that budget (shaded region), only Delethink continues to improve scaling from 24K up to 128K, while LongCoT-RL 24K and 8K plateau at their respective training limits.
  }
  \label{fig:extended_thinking}
\end{figure}

\section{Experimental Setup}
\label{sec:experiment_setup}

\paragraph{Model and Datasets}
We follow the RL training setup of \citet{deepscaler2025}. Specifically, we train on the DeepScaleR dataset, which contains approximately 40K curated, competition-level math problem–answer pairs. For all RL runs we initialize from \emph{R1-Distill 1.5B} \citep{guo2025r1zeropaper}, a strong reasoning LLM. All models use \emph{full-parameter fine-tuning}.
For benchmarks, we evaluate on AIME'24, AIME'25 \citep{maa_aime_2024}, and HMMT'25 \citep{hmmt25}. As out-of-distribution (OOD) tests, we additionally evaluate on the GPQA Diamond split \citep{rein2024gpqa}. We also report results on LiveCodeBench \citep{jain2025livecodebench}, restricting questions to those dated 2024-08-01–2025-02-01, following \citet{guo2025r1zeropaper}. These two are OOD relative to our math-only training data because GPQA Diamond consists of Ph.D.-level, open-domain questions and LiveCodeBench evaluates code generation.

\paragraph{Baselines}
We compare Delethink-RL to LongCoT-RL under identical maximum thinking budgets and similar training environments. Our primary baseline is LongCoT-RL with a 24K token budget. We also include LongCoT-RL with an 8K budget to quantify the benefit of scaling thinking tokens. All accuracy numbers are \emph{Pass@1} estimated from \(K\) samples, i.e. \emph{Accuracy (avg@\(K\))} denotes Pass@1 computed by averaging the correctness of \(K\) independently sampled responses for the same prompt. We select the best checkpoint using a hold-out validation set (AIME'24). To reduce evaluation variance, we sample 256 responses per prompt for every method and checkpoint, and report ``Accuracy (avg@128)'' via bootstrapping, yielding a standard deviation below 0.004 throughout.

\paragraph{Training Setup}
We follow the DeepScaleR recipe \citep{deepscaler2025} for our RL runs and rigorously implement established best practices. To ensure LongCoT-RL represents a strong performance, we conduct an hyperparameter search for LongCoT-RL. For Delethink, we adopt the same hyperparameters to ensure a fair comparison. In particular, we sweep the PPO clip-high ratio to maintain non-collapsing entropy during training for all runs (\Cref{fig:1000rlstep_entropy}). 
Each RL step samples 8 traces per prompt with a batch of 128 prompts (1024 episodes in total per step) and uses two optimizer updates per step.
We continue the RL training for 1000 steps.
Following recent practice, we disable the KL penalty (\(\beta=0\)) for best performance \citep{yu2025dapo}. We also apply truncated importance sampling to mitigate the distributional mismatch between the inference and training engines, which stabilizes learning \citep{yao2025:TIS}. In all runs, we employ a temperature 0.6 during training.
The reward function give a score 1 if the models correctly answers question within the budget, and zero otherwise.
For Delethink, we set the thinking context size to \(\mathcal{C}=8\text{K}\) based on its superior performance under a 24K budget on the base model (\Cref{fig:offtheshelf_r1_bench}). We use a cap of \(\mathcal{I}=5\) iterations and a markovian state size \(m=\mathcal{C}/2\). This configuration enables up to \(8\text{K} + (5-1)\times 4\text{K} = 24\text{K}\) thinking tokens (\Cref{sec:DI}), matching the LongCoT budget.
We employ an optimized implementation of both Delethink and LongCoT-RL under the verl framework \citep{sheng2024verl} using SGLang \citep{zheng2023sglang} and FSDP \citep{zhao2023:fsdp} as backend with sequence-packing and dynamic micro-batch sizing enabled for most efficient performance. 
Training is performed on 8×H100 GPUs without sequence parallelism to reduce communication overhead. See \Cref{sec:app:experiment_setup} for full details. 

\section{Results}
\label{sec:results}

\begin{figure}[!t]
  \centering
  \includegraphics[width=1\textwidth]{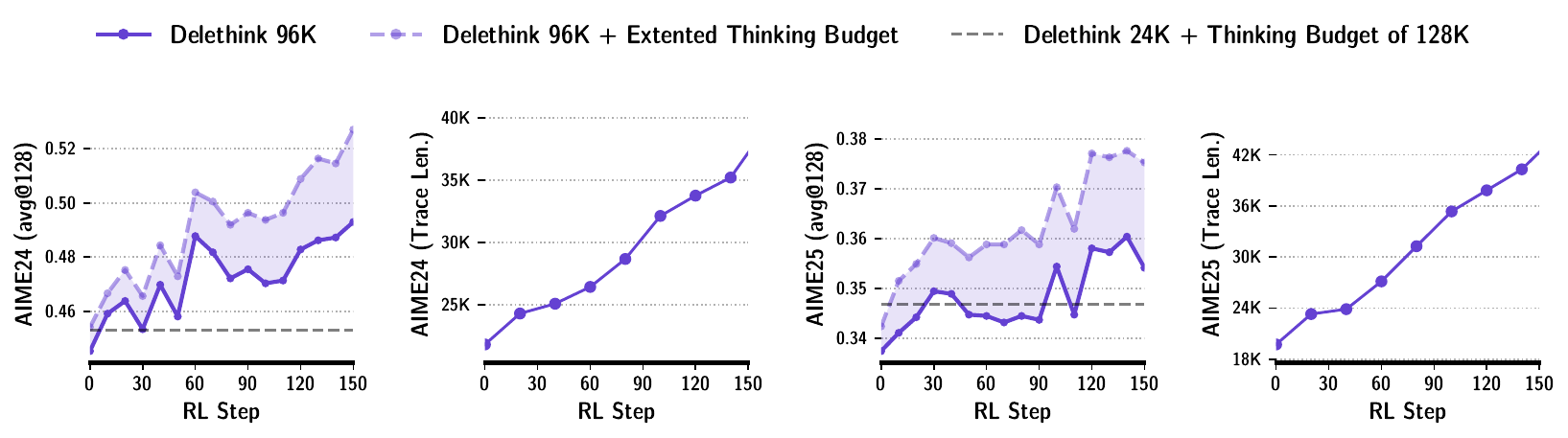}
  \caption{
    \textbf{Scaling Delethink to 96K.} AIME’24/’25 accuracy and average trace length vs. RL step for Delethink 96K; dashed curves extend the thinking budget to 256K and 128K tokens for Delethink 96K and Delethink 24K, respectively. Despite only 150 RL steps, 96K surpasses both the baseline and its extended thinking variant, with mean trace lengths reaching up to 42K.
  }
  \label{fig:delethink_96k}
\end{figure}

Delethink and LongCoT-RL 24K both have the same 24K-token thinking budget and start at the same pre-training performance (\Cref{fig:1000rlstep}). 
This equal initial performance shows that Delethink successfully samples markovian traces, with R1-Distill 1.5B, zero-shot which is favorable for RL training (We extensively investigate this zero-shot performance and its implications for scaling Delethink in \Cref{sec:why_works}). 
Despite Delethink runs in chunks of 8K, it can effectively think up to 24K budget matching and, at equivalent RL steps, outperforming LongCoT-RL 24K during training (\Cref{fig:1000rlstep}), with each Delethink RL step being less costly.
Thus, Delethink learns as effectively as LongCoT-RL while using fewer compute resources. 
The extended budget size is critical. 
With only 8K tokens, LongCoT-RL 8K underperforms Delethink by 5.5\% (\Cref{fig:1000rlstep}), confirming that the additional 16K thinking tokens are necessary for the stronger results and that Delethink learns to leverage them.
\Cref{fig:extended_thinking} supports this as it shows that both LongCoT-RL 24K and Delethink increase their accuracy while spending their thinking budget. 
 
\subsection{Task Performance}
We select the best checkpoint of each run based on validation performance on AIME'24 and then evaluate on math benchmarks.
As shown in \Cref{fig:1000rlstep}, with a 24K training-time thinking-token budget, Delethink attains higher scores than LongCoT-RL 24K on AIME'24, AIME'25, and HMMT'25. LongCoT-RL 8K consistently underperforms, underscoring the necessity of extended reasoning. 
On out-of-distribution tasks, GPQA-Diamond and LiveCodeBench, the absolute gains are modest; nevertheless, Delethink matches or slightly surpasses LongCoT-RL 24K. Overall, these results indicate that Delethink uses its thinking tokens as effectively as LongCoT-RL with reduced compute.

\subsection{Test-Time Scaling}
\label{sec:delethink_test_time_scales}
The benefits of a Markovian Thinker are not limited to train-time limits.
We investigate the test-time scaling behavior of Delethink and LongCoT-RL by reasoning beyond the thinking length they are originally trained on.
Note that this is sequential test-time scaling, Pass@1, and not parallel test-time scaling methods like majority@K or pass@K.\footnote{We omit budget-forcing test-time scaling methods such as S1 \citep{guo2025r1zeropaper} because they hurt performance on reasoning LLMs; see \Cref{app:budget_force_ablation}.} 
As shown in \Cref{fig:final_teaser_figure}, both LongCoT-RL 24K and LongCoT-RL 8K quickly plateau once they reach their training-time limits, indicating that LongCoT-RL’s test-time scaling is largely constrained by its training budget. In contrast, Delethink continues to improve even when reasoning with 100K more tokens than it encountered during training. For example, some AIME'25 problems are only solved after reasoning with up to 140K thinking tokens (See \Cref{app:extended_thinking_aime_detailed}), despite the model being trained with 24K. This pattern holds across all benchmarks, indicating that Delethink enables genuine test-time scaling far beyond training-time limits.

\subsection{Empirical Compute}
\label{sec:empirical_compute}
We evaluate the compute efficiency of Delethink against LongCoT-RL 24K. Delethink completes each RL step in 215s, faster than LongCoT-RL 24K at 248.5s.
Its token generation is also more efficient, reaching 8{,}500 tokens per second per H100 versus 6{,}000 for LongCoT-RL 24K.
As a result, Delethink finishes batch response generation in 130s compared to LongCoT-RL’s 170s, in line with the predictions of \Cref{sec:compute_scaling}.
The backward pass takes 80s for Delethink while 70s for LongCoT-RL 24K. At first glance, this appears to contradict theory.
However, the difference arises from constant and lower-order terms that dominate at shorter sequence lengths.
\Cref{fig:crossover} shows the crossover for R1-Distill 1.5B.
Below 32K tokens, Delethink can be slower since non-attention blocks dominate, but beyond that point, the quadratic scaling of attention makes Delethink increasingly faster.
At one million tokens, Delethink achieves $17\times$ reduction in FLOPs. In \Cref{fig:final_teaser_figure}, we empirically measure the time of a single RL step, assuming a target average thinking length.
As the theory predicts, the time grows quadratically for LongCoT but only linearly for Delethink.
To better understand this, we also measure the throughput of inference engines under both methods (\Cref{fig:1000rlstep}).
Consistent with the theory, Delethink maintains the same throughput regardless of thinking length, while LongCoT shows a steady decline.
\begin{figure}[!t]
    \centering
    \includegraphics[width=\textwidth]{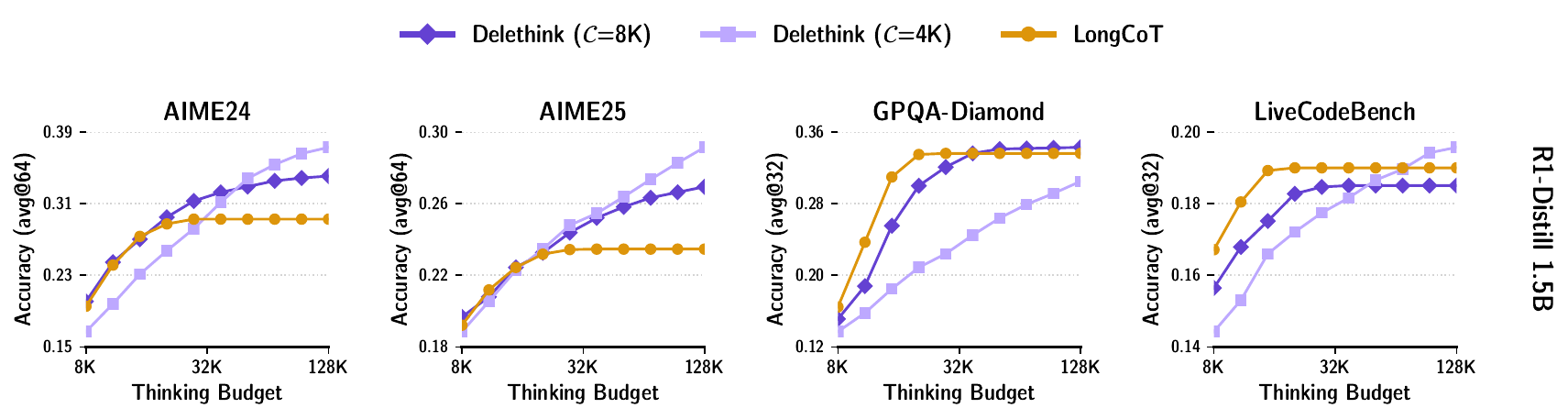}
    \caption{
    \textbf{Delethink Tracing at Initialization.} Even at initialization of RL, applying Delethink Tracing, without extra prompts or training,recovers most of LongCoT performance across thinking budgets. This indicates early signs of \emph{Markovian Thinking} in the base model. 
    }    
    \label{fig:offtheshelf_r1_bench}
\end{figure}

\subsection{Context Size Ablation}
Our main experiment runs Delethink with a context size, $\mathcal{C}$, of 8K.
We ablate the context size to 4K and 2K, set the state size $m$ to half the context size, and adjust iterations so the maximum thinking length remains 24K tokens.
We refer to these as 4K and 2K, respectively, while calling the main run 8K.
As shown in \Cref{fig:rl_context_ablation}, both 4K and 2K learn to exploit the very small state size to perform Markovian Thinking.
The 4K variant slightly underperforms 8K.
The 2K variant starts from a much lower initialization point.
Although it achieves lower accuracy than both 8K and 4K throughout, it still improves beyond the base model performance, indicating that Delethink works even under minimal context sizes.\footnote{Minimal context-size Delethink models might especially be of interest for low-memory settings like mobile devices. Instead of running a small reasoning model with LongCoT, they can run a much larger reasoning model with Delethink with a compact context size.}

\subsection{Scaling Delethink to 96K}
We study scaling the thinking budget from 24K to 96K tokens, made feasible by the linear compute cost of Delethink RL training. Concretely, we keep the thinking context to $\mathcal{C}=8\text{K}$ and increase the iteration cap from $\mathcal{I}=5$ to $\mathcal{I}=23$, yielding a 96K total budget. 
For this run, we train on OpenMath \citep{moshkov2025aimo}, which contains more difficult math competition problems than DeepScaleR and typically requires longer reasoning.
We initialize from the Delethink 24K checkpoint and continue training for 150 steps, using the same hyperparameters as earlier runs. \Cref{fig:delethink_96k} reports AIME'24 and AIME'25 results.
Despite the short training schedule, Delethink 96K not only surpasses the base performance of Delethink 24K checkpoint, but also matches or exceeds a test-time–scaled Delethink 24K evaluated with a 128K token budget. 
Moreover, the model’s average thinking lengths reach 36K (AIME'24) and 42K (AIME'25) tokens, indicating effective use of the larger budget. 
These preliminary results provide clear evidence that Delethink can scale to very long reasoning traces.

\begin{figure}[!t]
    \centering
    \includegraphics[width=\textwidth]{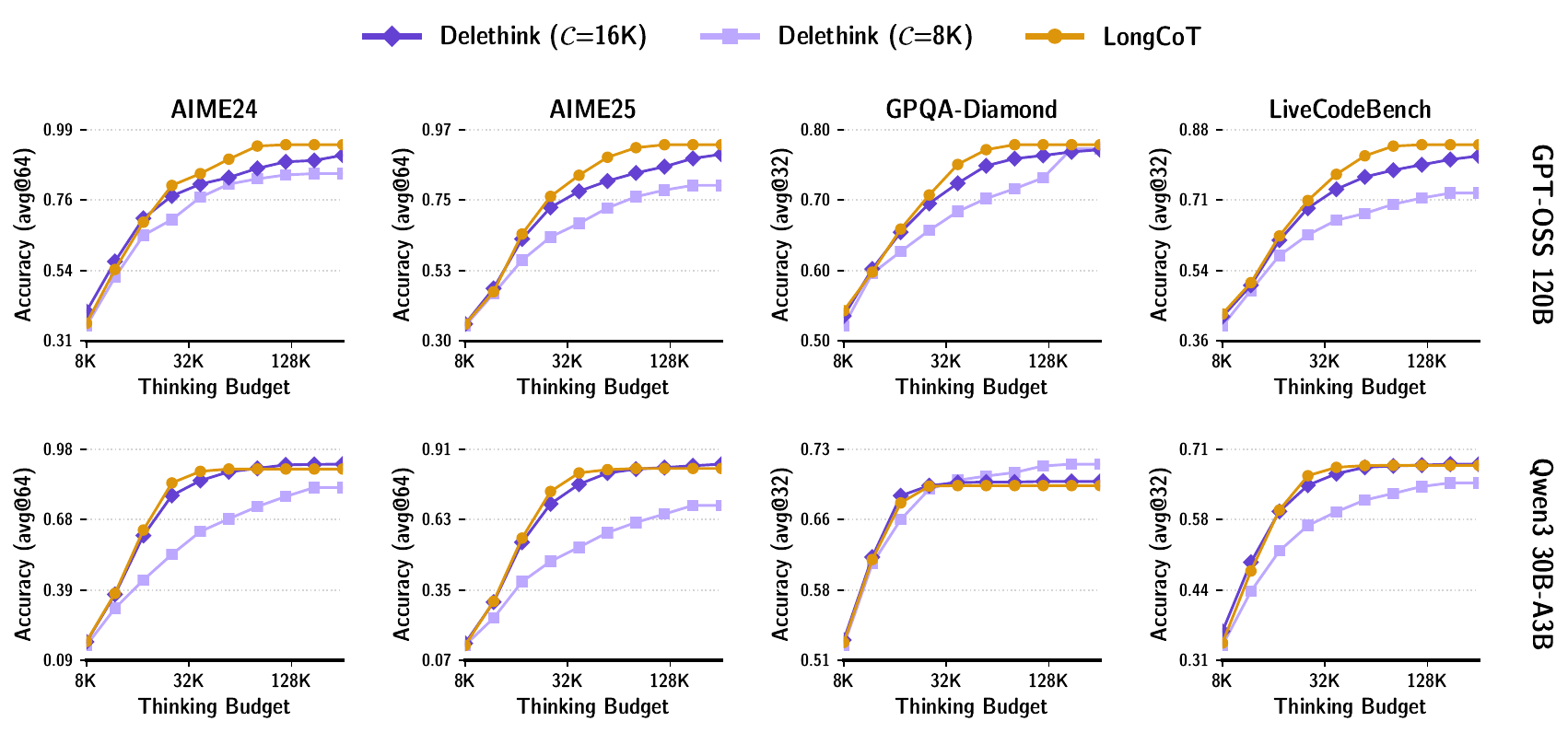}
    \caption{State-of-the-art reasoning LLMs, GPT-OSS-120B and Qwen3-30B-A3B, are capable of Markovian Thinking zero-shot, providing a strong initialization for training, signaling scalability. Delethink closely tracks LongCoT and recovers most of its final accuracy (the curves nearly coincide on Qwen3)}

    \label{fig:offtheshelf_qwen_bench}
\end{figure}

\section{Why Delethink Works?}
\label{sec:why_works}
RL works best when the policy has a decent prior.
That is, there is non-trivial probability of sampling trajectories with desired behavior.
We investigate Delethink Tracing at initialization of RL training: when Delethink Tracing applies broadly to off-the-shelf reasoning LLMs without any training or prompting, zero-shot.
We observe that these models show strong evidence of Markovian Thinking: they either recover or exceed their standard LongCoT performance.
Therefore, Delethink starts RL training from a strong and reliable initialization point. This explains, at least partially, why delethink training succeeds.
First, we showcase this phenomenon on R1-Distill 1.5B which is used in our training. Second, we show the strong presence of this phenomenon in SOTA LLMs, providing early signal that Delethink scales.

\subsection{R1-Distill}
We evaluate the R1-Distill 1.5B up to 128K thinking tokens. For Delethink, we vary the per-chunk context $\mathcal{C}$ from 4K to 8K to measure sensitivity to chunk size. As shown in \Cref{fig:offtheshelf_r1_bench}, Delethink Tracing with 8K matches LongCoT across tasks (\Cref{fig:offtheshelf_r1_bench}), despite deviating from its training-time regime. This provides a strong initialization for Delethink, exactly where our RL experiments started. Delethink Tracing with 4K only falls behind in the GPQA-Diamond, but exceeds on rest.
Furthermore, Delethink Tracing exhibits superior test-time scaling: on both AIME'24 and AIME'25 it surpasses its LongCoT performance by significant margin. In contrast, LongCoT quickly plateaus at 32K tokens, which is its recommended max thinking length \citep{guo2025r1zeropaper}. Interestingly, Delethink Tracing with 4K test-time scales further. We hypothesize one reason is because the model is unaware of the tracing procedure, it appears to perceive more headroom, emits \texttt{[EOS]} less, and thus thinks more. 
Finally, see \Cref{sec:app:test_time_scale_longcot24k} for evidence that Delethink can resume from LongCoT-RL checkpoints, where LongCoT-RL 24K shows solid performance under Delethink Tracing.

\begin{figure}[!t]
    \centering
    \includegraphics[width=\textwidth]{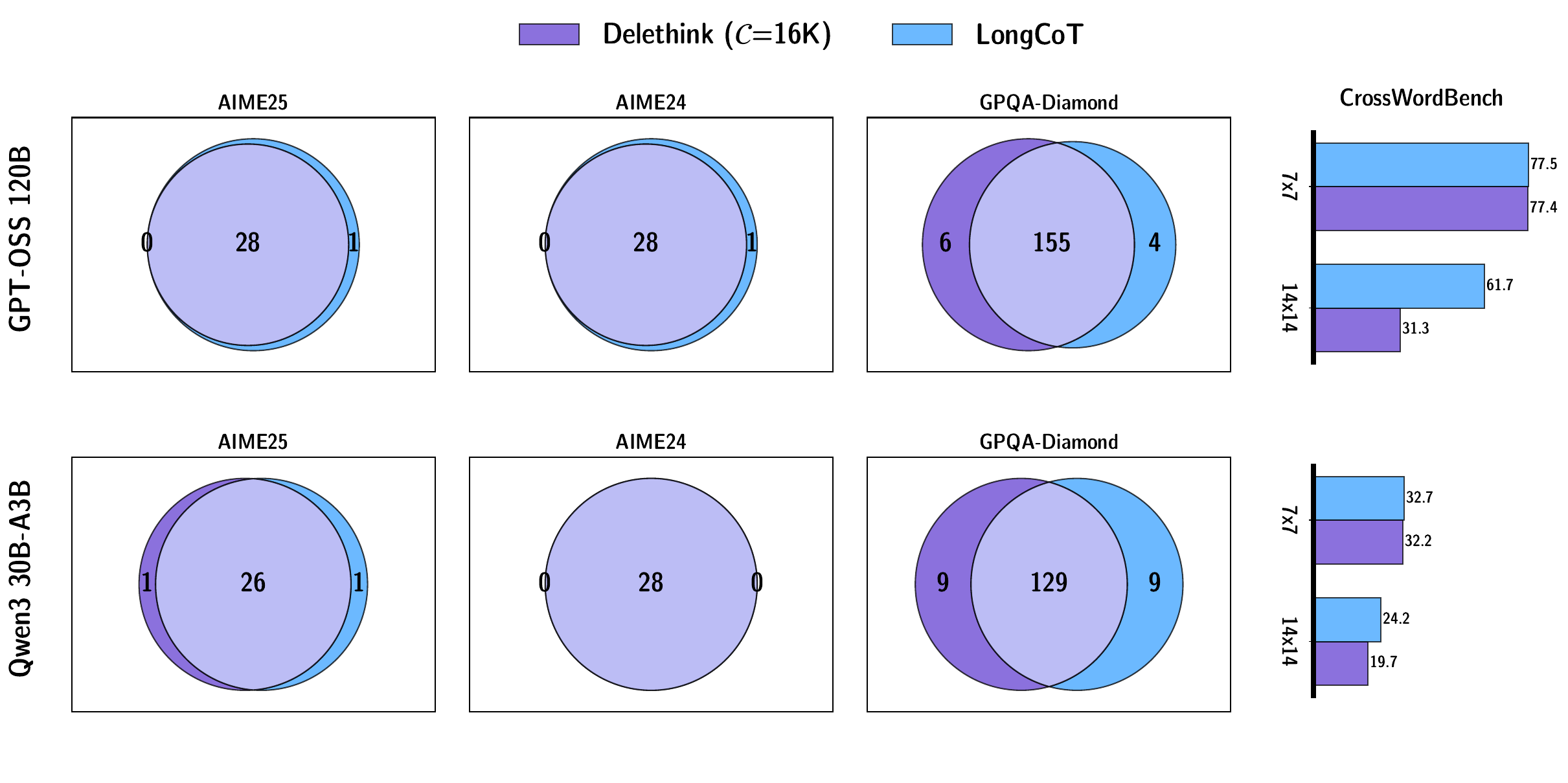} %
    \caption{
    (\textbf{Left}) Problem-solving overlap of Delethink vs. LongCoT: on AIME '24 and '25 they solve nearly the same set of questions. On GPQA, each solves an equal number that the other misses.
    (\textbf{Right}) CrossWordBench stress-tests Delethink: deleting previous tokens removes access to already found words. Delethink remains competitive, but its zero-shot limits are evident.
    }
    \label{fig:offtheshelf_qwen_ven}
\end{figure} 

\subsection{Will Delethink Scale?}
Training larger models is not possible within our compute budget. However, we can assess Delethink at step zero, when Delethink Tracing is applied zero-shot, to test whether the policy already supports Markovian traces. Strong performance at initialization, without training or prompting, is a promising signal that subsequent RL training will succeed, since it begins with many positive samples and the desired behavior already in-distribution.  

To test this, we evaluate GPT-OSS 120B \citep{Agarwal2025:GPT-OSS} and Qwen3 30B-A3B Thinking\footnote{\url{https://huggingface.co/Qwen/Qwen3-30B-A3B-Thinking-2507}} \citep{qwen3}, two state-of-the-art open-source reasoning LLMs, under Delethink. We find that a 16K context is a solid default as it recovers most of the original performance, while 8K achieves a reasonable score and is suitable when compute is tight. We also study problem coverage at initialization to check whether the model loses the ability to solve some questions compared to standard LongCoT, and find that coverage remains equally strong. Finally, we benchmark the models on a task that requires live memory to deliberately break Delethink, and find that even there they admit Markovian solutions. Overall, these results suggest that Delethink has a strong chance of scaling effectively.

\paragraph{Results and Test-time scaling}
As shown in \Cref{fig:offtheshelf_qwen_bench}, both models achieve higher scores by more reasoning, although less pronounced for Qwen3, under both LongCoT and Delethink Tracing. Delethink Tracing, with $\mathcal{C}=16\text{K}$, matches or exceeds LongCoT on Qwen3 and almost matches on GPT-OSS. Tracing with $\mathcal{C}=8\text{K}$ has a lower test-time scaling slope and lags behind LongCoT.
These results indicate that current SOTA reasoning LLMs exhibit latent markovian behavior even without explicit training for it and Delethink with 8K context size provides a strong initialization for starting Delethink.

\paragraph{Problem Coverage}
We compare problem coverage on AIME'24, AIME'25, and GPQA-Diamond (\Cref{fig:offtheshelf_qwen_ven}), counting a problem as solved if the majority vote is correct.\footnote{Majority voting does not apply to code, so LiveCodeBench is excluded.} On both AIMEs, both methods solve essentially the same number of problems. In detail, LongCoT solves one more problem on both AIMEs with GPT-OSS. For Qwen3, AIME’24 is identical; on AIME’25 each method uniquely solves one problem (equal totals). On GPQA-Diamond, GPT-OSS solves 161 problems with Delethink vs 159 with LongCoT; Qwen3 solves 138 with both.

\paragraph{Stress testing Delethink Tracing}
We search for a task where Delethink does not start from a strong initialization. As Delethink loses access to old thoughts, tasks that require writing to a memory like crossword might be challenging as solved cases may be lost if not carried by the model. We evaluate on CrossWordBench \citep{leng2025:CrossWordBench}, which contains crossword puzzles at varying difficulty levels. Reasoning in this task requires maintaining a live grid plus filled entries—state that can exceed the capacity of $m$. As shown in \Cref{fig:offtheshelf_qwen_ven}, Delethink Tracing achieves performance on $7\times7$ puzzles that is broadly comparable to LongCoT across both GPT-OSS and Qwen3. On the more challenging $14\times14$ puzzles, performance declines relative to LongCoT, though both models continue to produce a nontrivial number of valid Markovian traces. Overall, these results suggest that even in settings designed to stress-test Delethink, it preserves meaningful probabilistic coverage and readily discovers markovian solutions, giving headroom for training.

\paragraph{Effect of per-chunk context size $\mathcal{C}$}
Delethink Tracing with 16K context size recovers almost full-context performance. Smaller $\mathcal{C}$, 8K, however underperforms larger $\mathcal{C}$ and the LongCoT baseline. As the markovian state $m$ carried between chunks shrinks too much, the model may lose the thread of reasoning, suggesting the zero-shot state size should be higher. These results show 16K context works decently from scratch, but 8K requires extra RL training to make the state more markovian.

\section{Discussion}
Scaling thinking length has advanced reasoning LLMs significantly, but further scaling in the current LongCoT paradigm incurs a prohibitive quadratic compute cost. In this work, we present a new paradigm that decouples the thinking length and the context size. We propose the Markovian Thinking Paradigm, where each step retains only the minimal state needed to continue, extending thinking under a bounded state size; yielding linear compute and constant memory for both training and subsequent test-time generation. We show Delethink is practical. Delethink trains native markovian thinkers by starting from off-the-shelf reasoning LLMs.  Experiments match LongCoT performance and show superior test-time scaling. Furthermore, we empirically verify the practicality of Delethink in scaling the thinking budget to nearly one hundred thousand tokens. Finally, we both theoretically and empirically demonstrate the practical benefits of scaling thinking with Delethink compared to LongCoT.

Additionally, here we share some hypotheses that might help explain some surprising results further; these are speculative and meant to inspire rigorous follow-up. In our experiments, Delethink surpassed LongCoT-RL 24K. Pretraining might have played a role. As most LLM pretraining uses short contexts (e.g., 8K tokens), models may think sharper in shorter contexts. This aligns with retrieval research showing effective context lengths shorter than nominal limits \citep{hsieh2024ruler,modarressi2025nolima}. We also hypothesize that Delethink samples positive responses zero-shot partly because LLMs are pre-trained on human reasoning traces and human reasoning may be roughly Markovian, carrying necessary results forward in spelled-out thoughts. Note that 8K tokens is near 20 textbook pages. Finally, RL on Delethink might encourage more abstract reasoning. With limited context, an LLM may learn tokens that encapsulate broader ideas to save space. Observing this may require large-scale RL.

Thanks to Delethink, Markovian Thinking is no longer a theoretical concept but a practical, near-term reality. 
The possibility and effectiveness of markovian thinking, that Delethink shows, could change how we design reasoning models.
Delethink thinks and learns just as well without its full history as LongCoT does with it.
This could never work in areas like long-context retrieval, where the model depends on information the whole context.
This shows that thinking is different and can, in fact, be done in a Markovian way.
Delethink provides evidence that reasoning LLMs can be trained differently and may exploit more efficient architectures such as non-quadratic attention, sliding-window attention, sparse attention, or state-space models, since reasoning can be done Markovian. We hope this paper serves as a stepping stone toward models that think across millions of tokens.

\section*{Acknowledgements}
SR is supported by a CIFAR AI Chair and a Mila–Samsung grant. AC is supported by Canada CIFAR AI Chair, the Canada Research Chair, the NSERC Discovery Grant, and funding from Microsoft Research. SC is supported by the Canada CIFAR AI Chair, the Canada Research Chair in Lifelong Machine Learning, and the NSERC Discovery Grant. We thank the Mila IDT team and the Digital Research Alliance of Canada for providing the compute resources used in our experiments.

\bibliography{main}
\bibliographystyle{main}

\newpage
\appendix
\renewcommand\thefigure{\thesection.\arabic{figure}}
\renewcommand\thetable{\thesection.\arabic{table}}
\setcounter{figure}{0}
\setcounter{table}{0}

\part*{Appendix}
\addcontentsline{toc}{part}{Appendix} %
\etocsetnexttocdepth{2}               %
\etocsettocstyle{\subsection*{Table of Content}}{}  %
\localtableofcontents   
\newpage

\section{Deriving Delethink Loss}
\label{app:deriving_Delethink_loss}
In this section, we show that the Delethink Training loss is basically the policy gradient and derive it from scratch. This proof will be trivial to readers familiar with policy gradients (\cite{sutton1999pg}). We include it for completeness.

The RL objective is to maximize the expected return which is the following:
\begin{equation}
\mathcal{J}(\theta) = \mathbb{E}_{\vq \sim \mathcal{D}, \tau \sim \pi_\theta(\cdot|\vq)}\left[ \mathcal{R}(\tau) \right]
\label{eq:rl_objective_1}
\end{equation}

Policy gradient is the gradient of the expected return with respect to the policy parameters. That is:
\begin{equation}
\nabla_\theta \mathcal{J}(\theta) = \nabla_\theta \mathbb{E}_{\vq \sim \mathcal{D}, \tau \sim \pi_\theta(\cdot|\vq)}\left[ \mathcal{R}(\tau) \right]
\label{eq:rl_objective_2}
\end{equation}
First, note that the gradient cannot pass through the expectations as we have distributional dependency. That is, the distribution of $\tau$ depends on $\theta$. However, the gradient can pass through the expectation over the queries as the distribution of dataset does not depend on $\theta$.
\begin{equation}
\nabla_\theta \mathcal{J}(\theta) = \nabla_\theta  \mathbb{E}_{\vq \sim \mathcal{D}} \int_{\tau \in \pi_\theta(\cdot|\vq)} \pi_\theta(\tau)\mathcal{R}(\tau)
\end{equation}

As noted, the distribution of $q$ does not depend on $\theta$, the gradient passes through this expectation:
\begin{equation}
\nabla_\theta \mathcal{J}(\theta) = \mathbb{E}_{\vq \sim \mathcal{D}}  \nabla_\theta \int_{\tau \in \pi_\theta(\cdot|\vq)} \pi_\theta(\tau)\mathcal{R}(\tau)
\end{equation}
Next, the gradient passes through the integration as the bounds of the integration does not depend on $\theta$: 
\begin{equation}
\nabla_\theta \mathcal{J}(\theta) = \mathbb{E}_{\vq \sim \mathcal{D}}   \int_{\tau \in \pi_\theta(\cdot|\vq)} \nabla_\theta \pi_\theta(\tau)\mathcal{R}(\tau)
\end{equation}
However, the term insides the integral is intractable as it is an integration over all possible traces. In order to write this as an expectation so we can estimate this via samples we divide and multiply $\pi_\theta(\tau)$:
\begin{equation}
\begin{aligned}
\nabla_\theta \mathcal{J}(\theta) 
& = \mathbb{E}_{\vq \sim \mathcal{D}}   \int_{\tau \in \pi_\theta(\cdot|\vq)} \nabla_\theta \pi_\theta(\tau)\mathcal{R}(\tau)
\\&= \mathbb{E}_{\vq \sim \mathcal{D}}   \int_{\tau \in \pi_\theta(\cdot|\vq)} \pi_\theta(\tau) \frac{\nabla_\theta \pi_\theta(\tau)}{\pi_\theta(\tau)}\mathcal{R}(\tau)
\\&
\end{aligned}
\end{equation}
However, the term inside the integral is the derivative of the log probability with respect to the parameters, famously known as the log-trick. That is,  $\frac{\nabla_{\theta} \pi_\theta(\tau)}{\pi_\theta(\tau)} = \nabla_{\theta} \log(\pi_\theta(\tau))$. Therefore, we have:
\begin{equation}
\begin{aligned}
\nabla_\theta \mathcal{J}(\theta) 
 &= \mathbb{E}_{\vq \sim \mathcal{D}}   \int_{\tau \in \pi_\theta(\cdot|\vq)} \pi_\theta(\tau) \nabla_{\theta} \log(\pi_\theta(\tau)) \mathcal{R}(\tau)
 \\&= \mathbb{E}_{\vq \sim \mathcal{D}, \tau \sim \pi_\theta(\cdot|\vq)}    \nabla_{\theta} \log \pi_\theta(\tau) \mathcal{R}(\tau)
\label{eq:reinforce}
\end{aligned}
\end{equation}
Note that, this is the famous REINFORCE (\cite{williams1992reinforce}). Next, we expand $\log(\pi_\theta(\tau))$. We know the Delethink trace $\tau$ is a sequence of chunks. That is, $\tau \;=\; \left[\,(\vx_1, \vy_1), \ldots, (\vx_L, \vy_L)\,\right]$. Therefore, we can write the log probability of generating $\tau$ as the sum of log probabilities of generating each chunk. That is,
\begin{equation}
\log(\pi_\theta(\tau)) = \sum_{l=1}^{L} \log \pi_\theta(\vy_l)
\end{equation}
And if we expand the log probability of generating each chunk into sum of log probability of generating the response tokens:
\begin{equation}
\log(\pi_\theta(\tau)) = \sum_{l=1}^{L} \sum_{t=1}^{|\vy_l|}\log\pi_\theta(\vy_{l,t}\mid x_l,\vy_{:t-1}) 
\end{equation}
By substituting this to \ref{eq:reinforce}, we get the following:
\begin{equation}
\nabla_\theta \mathcal{J}(\theta)  =  \mathbb{E}_{\vq \sim \mathcal{D}, \tau \sim \pi_\theta(\cdot|\vq)}      \sum_{l=1}^{L} \sum_{t=1}^{|\vy_l|} \nabla_{\theta} \log \pi_\theta(\vy_{l,t})  \mathcal{R}(\tau)
\end{equation}
Next, according to \cite{sutton1999pg}, instead of the reward, we can multiply the log probability of each token with its corresponding advantage:
\begin{equation}
\nabla_\theta \mathcal{J}(\theta)  =  \mathbb{E}_{\vq \sim \mathcal{D}, \tau \sim \pi_\theta(\cdot|\vq)}      \sum_{l=1}^{L} \sum_{t=1}^{|\vy_l|}\nabla_{\theta}\log \pi_\theta(\vy_{l,t})\hat{A}_{t}.
\end{equation}
This concludes the derivation of the policy gradient form of the objective. However, we will proceed to add PPO clipping, KL loss, and the normalization. They bias the gradient, so the term is not the exact policy gradient anymore. First, we add the PPO clippings:
\begin{equation}
\mathbb{E}_{\vq \sim \mathcal{D}, \tau \sim \pi_\theta(\cdot|\vq)} 
\sum_{l=1}^{L} \sum_{t=1}^{|\vy_l|}
\min \left[
\frac{\pi_\theta(\vy_{l,t})}{\pi_{\theta_{\text{old}}}(\vy_{l,t})} \hat{A}_{t},
\;
\text{clip}\!\left(
\frac{\pi_\theta(\vy_{l,t})}{\pi_{\theta_{\text{old}}}(\vy_{l,t})}, 1-\epsilon, 1+\epsilon
\right) \hat{A}_{t}
\right] 
\end{equation}
Additionally, we add the normalization by the total number of response tokens in the trace $\ell(\tau_g)$:
\begin{equation}
\mathbb{E}_{\vq \sim \mathcal{D}, \tau \sim \pi_\theta(\cdot|\vq)} 
\frac{1}{\ell(\tau_g)}\sum_{l=1}^{L} \sum_{t=1}^{|\vy_l|}
\min \left[
\frac{\pi_\theta(\vy_{l,t})}{\pi_{\theta_{\text{old}}}(\vy_{l,t})} \hat{A}_{t},
\;
\text{clip}\!\left(
\frac{\pi_\theta(\vy_{l,t})}{\pi_{\theta_{\text{old}}}(\vy_{l,t})}, 1-\epsilon, 1+\epsilon
\right) \hat{A}_{t}
\right] 
\end{equation}
Finally, we add the KL term that limits deviation from the reference policy. 
\begin{equation}
\mathbb{E}_{\vq \sim \mathcal{D}, \tau \sim \pi_\theta(\cdot|\vq)} 
\frac{1}{\ell(\tau_g)}\sum_{l=1}^{L} \sum_{t=1}^{|\vy_l|}
\min \left[
\frac{\pi_\theta(\vy_{l,t})}{\pi_{\theta_{\text{old}}}(\vy_{l,t})} \hat{A}_{t},
\;
\text{clip}\!\left(
\frac{\pi_\theta(\vy_{l,t})}{\pi_{\theta_{\text{old}}}(\vy_{l,t})}, 1-\epsilon, 1+\epsilon
\right) \hat{A}_{t}
\right] 
- \beta \, \mathrm{KL}\!\left[ \pi_\theta \,\|\, \pi_\text{ref} \right]
\end{equation}
Note that this is basically our objective. 
\section{Deriving Throughput Relation}
\label{app:fluid_guided}
We follow \citet{ao2025fluid-guided}. Consider a GPU with maximum memory \(M\), serving an infinite stream of incoming requests and an equally unbounded stream of completed outputs. Let \(n^{*}\) denote the equilibrium number of concurrent requests. Each request has \(l\) prefill tokens and \(l'\) decode tokens. In steady state, the KV-cache memory required is
\begin{equation}
M^{*} \;=\; n^{*}\!\left( l + \tfrac{l'}{2} \right).
\end{equation}

According to \citet{ao2025fluid-guided}, the equilibrium throughput of an attention-based LLM under a memory constraint is
\begin{equation}
\mathcal{T}^{*}
\;=\;
\frac{n^{*}}{\,d_{0} \;+\; d_{1}\, n^{*}\!\left( l + \tfrac{l'}{2} \right)},
\end{equation}
where \(d_{0}\) is the fixed per-batch overhead and \(d_{1}\) is the time cost per unit of memory. Thus, throughput exhibits an inverse dependence on the total effective context length. In regimes where
\(
d_{1}\, n^{*}\!\left( l + \tfrac{l'}{2} \right) \gg d_{0},
\)
we obtain the approximation
\begin{equation}
\text{Throughput}^{*}
\;\approx\;
\frac{n^{*}}{\,d_{1}\, n^{*}\!\left( l + \tfrac{l'}{2} \right)}
\;=\;
\frac{1}{\,d_{1}\!\left(l+\tfrac{l'}{2}\right)}.
\end{equation}

Consequently, for the same GPU, the ratio of throughputs when decoding lengths are \(l'_{1}\) and \(l'_{2}\) satisfies
\[
\frac{\mathcal{T}_{l'_1}}{\mathcal{T}_{l'_2}}
\;\approx\;
\frac{\frac{1}{d_1\!\left(l+\tfrac{l'_1}{2}\right)}}{\frac{1}{d_1\!\left(l+\tfrac{l'_2}{2}\right)}}
\;=\;
\frac{\,l+\tfrac{l'_2}{2}\,}{\,l+\tfrac{l'_1}{2}\,}.
\]
In the long-thinking regime where \(l' \gg l\), this simplifies to
\[
\frac{\mathcal{T}_{l'_1}}{\mathcal{T}_{l'_2}}
\;\approx\;
\frac{l'_2}{l'_1},
\]
establishing that throughput is (approximately) inversely proportional to the decoding length in this regime. \hfill\(\square\)

\paragraph{Intuition.}
Decoding repeatedly accesses the KV cache, whose size, and therefore memory-time cost, grows linearly with the number of decoded tokens \(l'\). This linear growth drives the inverse relationship between throughput and \(l'\) in the long-thinking regime.

\section{Delethink FLOPs Crossover vs. LongCoT 24K}
\begin{figure}[!t]
    \centering
    \includegraphics[width=0.37\textwidth]{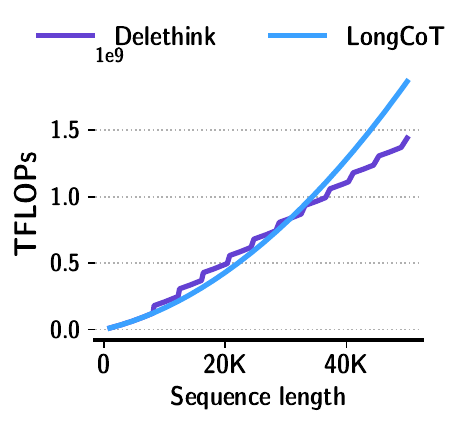}
    \caption{When training R1DistillQwen1.5B with Delethink with $8\text{k}$ thinking context, for thinking lengths $\sim 30\text{k}$ the number of FLOPs for both is equal. This is because the non-attention components like dense layers dominate here. However, after $30\text{k}$ quadratic cost of attention dominates.}
    \label{fig:crossover}
\end{figure}

We can compute, in closed form, the FLOPs required for reasoning of a certain length under Delethink and LongCoT. We calculated this for our main experiment setup, where Delethink runs with $\mathcal{C}=8\text{K},m=4\text{K}$, single step with 1000 episodes, and results are shown in \Cref{tab:compute_memory}. We highlight the crossover in \Cref{fig:crossover} at short sequences. Up to roughly $30\text{K}$ tokens, Delethink spends slightly more FLOPs than LongCoT. That seems counterintuitive: Delethink scales linearly, LongCoT quadratically. This is because in Delethink each token is generated once and then reprocessed as input to the next chunk. While the quadratic compute of attention quickly overtakes this double-counting cost in longer sequences, it matters at short lengths. This is clear when considering non-attention layers. FLOPs in non-attention layers scale linearly, and in Delethink each token is generated once and then reprocessed in the next chunk as a prompt. Therefore, the cost of non-attention layers doubles. The crossover occurs around $30\text{K}$ tokens in our setting. Beyond this point, Delethink is linearly more efficient in terms of FLOPs. Note that while total backward time tracks total FLOPs, generation time at these response lengths is still slower for LongCoT. This is because generation throughput is limited by memory-access and memory-footprint bottlenecks, not just FLOPs. As explained in \Cref{sec:empirical_compute}, our training still runs faster even in this regime due to faster generation. Our next step, Delethink 96K, is significantly faster than its counterpart LongCoT-RL 96K in both backward pass and generation, to the point that we could not train the latter under our compute resources.

\begin{figure}[t]
\centering
\begin{minipage}{0.98\linewidth}
\begin{minted}[style=xcode,fontsize=\footnotesize,breaklines,linenos,frame=single]{python}
import torch
from transformers import AutoTokenizer, AutoModelForCausalLM

# Load model and tokenizer
model_path = "deepseek-ai/DeepSeek-R1-Distill-Qwen-1.5B"
tokenizer = AutoTokenizer.from_pretrained(model_path)
model = AutoModelForCausalLM.from_pretrained(model_path)

C, m, I, T = 8192, 4096, 5, 0.6  # chunk context, markovian tail, iteration cap, temperature

query_ids = tokenizer.apply_chat_template(
    [{"role": "user", "content": "Solve the problem step by step; ..."}],
    tokenize=True,
    add_generation_prompt=True,
)

trace_ids = []  # list of generated token IDs across chunks
prompt_ids = list(query_ids)  # current-chunk prompt (reset each iteration)
it = 0

while it < I:
    max_new = C if it == 0 else (C - m)

    inp = torch.tensor([prompt_ids], device=model.device)
    out = model.generate(
        input_ids=inp,
        max_new_tokens=max_new,
        do_sample=True, temperature=T,
        eos_token_id=tokenizer.eos_token_id, pad_token_id=tokenizer.pad_token_id,
        use_cache=True,
    )

    full_seq = out.sequences[0].tolist()
    response_ids = full_seq[len(prompt_ids) :]
    trace_ids.extend(response_ids)

    if it == 0:
        # Fold the first hundered tokens of the first chunk into query
        query_ids = query_ids + response_ids[:100] 
    
    if response_ids[-1] == tokenizer.eos_token_id:
        break

    prompt_ids = query_ids + response_ids[-m:]
    it += 1

response_text = tokenizer.decode(trace_ids, skip_special_tokens=False)
response_text
\end{minted}
\end{minipage}
\caption{Example of \textbf{Delethink-tracing} chunked rollout implemented with HuggingFace \texttt{model.generate} (batch size $=1$). In practice, we use high-performance inference engines and a large batch size.}
\label{fig:delethink-hf}
\end{figure}

\section{Detailed RL Training Setup}
In this section we describe our training setup and hyperparameters meticulously. 
\label{app:training_setup}
\label{sec:app:experiment_setup}

\paragraph{Model}
All reinforcement learning runs begin from R1-Distill 1.5B \citep{guo2025r1zeropaper}, a strong reasoning LLM. We employ full-parameter fine-tuning throughout training.

\paragraph{Datasets}
Our experiments follow the RL training framework introduced in \citet{deepscaler2025}. We train on the DeepScaleR dataset, which includes roughly 40K carefully curated, competition-style math problem–solution pairs.
Evaluation is conducted a hold-out math competition benchmark: AIME’24, AIME’25 \citep{maa_aime_2024}, and HMMT’25 \citep{hmmt25}.

\paragraph{Baselines}
We benchmark Delethink-RL against LongCoT-RL, ensuring identical maximum thinking budgets and closely matched training conditions. The main baseline is LongCoT-RL with a 24K-token budget, alongside an additional 8K-token configuration to highlight the effects of scaling reasoning tokens. Accuracy is reported as Pass@1, estimated from K independent samples. Specifically, “Accuracy (avg@K)” is the average Pass@1 computed over K sampled responses for the same prompt. To quantify uncertainty, we use a nonparametric bootstrap over continuations: for B{=}5000 replicates, for each prompt we resample K continuations \emph{with replacement}, average within-prompt, then average across prompts. Model checkpoints are selected based on performance on the held-out AIME’24 validation set. To reduce noise, we generate 256 responses per prompt for each method and checkpoint, and report Accuracy (avg@128) using this bootstrap aggregation, yielding standard deviations consistently below 0.004.
Evaluation is done with temperature $0.6$, top$-p = 1.0$ and no top-$k$ (disabled). For test-time scaling, we right-trimmed each trace so its thinking token count matched the specified test-time budget.

\paragraph{Training Setup}
We adopt the DeepScaleR RL recipe \citep{deepscaler2025}, adhering to widely recognized best practices. To establish LongCoT-RL as a strong baseline, we perform a hyperparameter sweep, and then apply the same settings to Delethink-RL for fairness. In particular, we tune the PPO clip-high ratio (from 0.2 to 0.28) to maintain entropy stability across all runs (\Cref{fig:1000rlstep_entropy}). Each RL step consists of sampling 8 traces for 128 prompts (1024 episodes total) and applying two optimizer updates. Training is run for 1000 steps. Following recent findings, the KL penalty is disabled ($\beta=0$) for optimal results \citep{yu2025dapo}. Truncated importance sampling is also applied to address distributional mismatches between inference and training engines, improving stability \citep{yao2025:TIS}. Training is conducted with temperature 0.6.

\paragraph{Delethink Parameters}
For Delethink-RL, we set the reasoning context size to $\mathcal{C}=8\text{K}$, chosen due to its effectiveness under a 24K-token budget with the base model (\Cref{fig:offtheshelf_r1_bench}). We cap iterations at $\mathcal{I}=5$, with a Markovian state size of $m=\mathcal{C}/2$. This setup allows up to $8\text{K} + (5-1)\times 4\text{K} = 24\text{K}$ reasoning tokens (\Cref{sec:DI}), aligning with the LongCoT-RL budget.

\paragraph{Implementation Framework}
Both Delethink-RL and LongCoT-RL are implemented efficiently within the verl framework \citep{sheng2024verl}, leveraging SGLang \citep{zheng2023sglang} and PyTorch FSDP \citep{zhao2023:fsdp}, with sequence-packing and dynamic micro-batching for throughput optimization. Training is executed on 8×H100 GPUs, avoiding sequence parallelism to minimize communication overhead. 
We use HF Math-verify \citep{math_verify} as our reward function.

\subsection{Hyperparameters} \label{app:hyperparameter}
\Cref{tab:hps} represents the key hyperparamters used in our RL training experiments.
\begin{table}[!t]
\centering
\small
\setlength{\tabcolsep}{6pt}
\caption{Key hyperparameters for \textbf{LongCoT RL} vs. \textbf{Delethink}.}
\resizebox{\textwidth}{!}{%
\begin{tabular}{@{}l@{\hspace{3cm}}cc@{}}
\toprule
\textbf{Hyperparameter} & \textbf{Delethink} & \textbf{LongCoT RL} \\
\midrule
\multicolumn{3}{@{}l}{\textbf{Rollout (inference)}}\\
Sampling (T / top-p / top-k / $n$) & 0.6 / 1.0 / $-1$ / 8 & 0.6 / 1.0 / $-1$ / 8 \\
Prompt / response length (tokens) & 2{,}048 / 8{,}192 & 2{,}048 / 24{,}576 \\
\midrule
\multicolumn{3}{@{}l}{\textbf{Algorithm}}\\
Advantage estimator & GRPO & GRPO \\
KL penalty & N/A & N/A \\
\midrule
\multicolumn{3}{@{}l}{\textbf{Policy optimization}}\\
PPO epochs / number of grad updates & 1 / 2 & 1 / 2 \\
Clip ratio (low / high) & 0.20 / 0.26 & 0.20 / 0.26 \\
Grad clip (global-norm) & 1.0 & 1.0 \\
Loss aggregation & Trace Length-Batch Mean & Seq-mean-Batch Mean \\
\midrule
\multicolumn{3}{@{}l}{\textbf{Actor optimizer}}\\
Learning rate & $1\times 10^{-6}$ & $1\times 10^{-6}$ \\
Weight decay & 0 & 0 \\
Warmup & constant; steps $=-1$ (disabled) & constant; steps $=-1$ (disabled) \\
Total training steps (optim) & 1{,}000 & 1{,}000 \\
\midrule
\multicolumn{3}{@{}l}{\textbf{Data}}\\
Train / val batch size & 128 / 16 (per step, logical) & 128 / 16 (per step, logical) \\
\midrule
\multicolumn{3}{@{}l}{\textbf{Rewarding}}\\
Reward function & HF-Math-Verify & HF-Math-Verify \\
\bottomrule
\end{tabular}
}
\label{tab:hps}
\end{table}

\section{Markovian State Size Ablation}
\label{app:markovian_state_size_ablation}

\begin{figure}[!t]
    \centering
    \includegraphics[width=\textwidth]{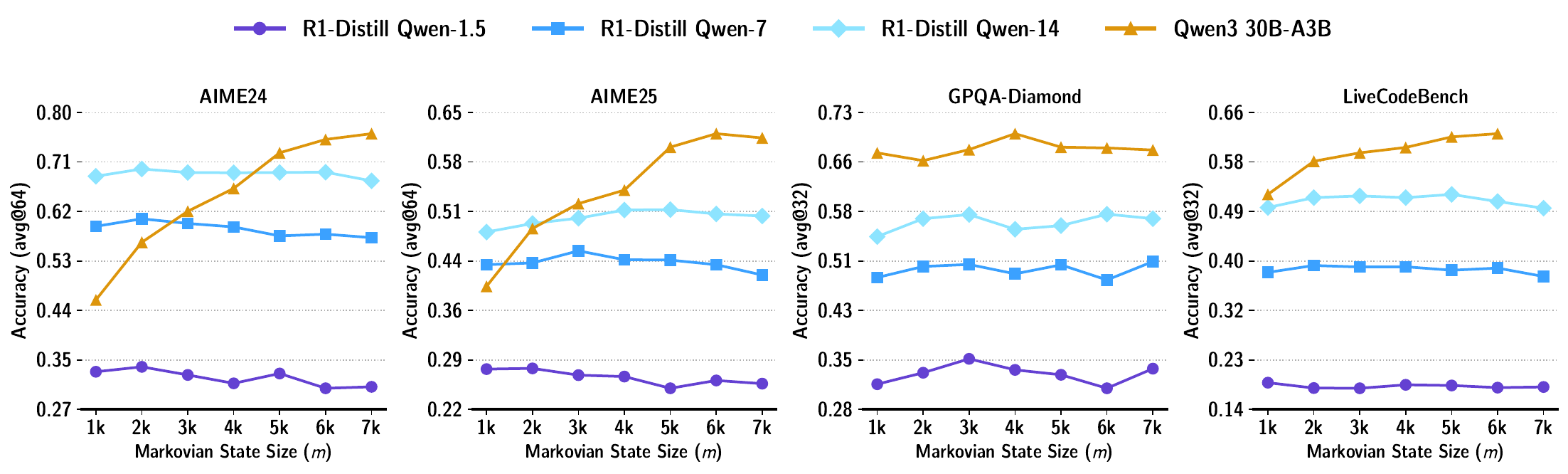}
    \caption{Ablation of the Markovian state size \(m\) at fixed per-chunk context \(\mathcal{C}=8\text{k}\). R1-Distill models (1.5B to 14B) achieve stable accuracy even with small carried states (\(m\ll \mathcal{C}\)), indicating that only modest memory is required for Markovian behavior. In contrast, Qwen3 (native 256k context) shows clear gains from larger state sizes, reflecting its long-context prior and the longer Delethink traces on AIME and LiveCodeBench tasks.}
    \label{fig:prefill_ablation}
\end{figure}

In this ablation we vary the size of the Markovian state \(m\) while holding the per-chunk thinking context fixed at \(\mathcal{C}=8\text{k}\) and keeping the overall token budget constant. We sweep \(m \in \{1\text{k},2\text{k},3\text{k},4\text{k},5\text{k},6\text{k},7\text{k}\}\). To ensure a matched budget across settings, we adjust the Delethink iteration cap \(\mathcal{I}\) using the identity $T_{\max}=\mathcal{C}+(\mathcal{I}-1)(\mathcal{C}-m)$. Concretely, we set \(\mathcal{I}=\{6,7,8,10,13,19,37\}\) for \(m=\{1\text{k},2\text{k},3\text{k},4\text{k},5\text{k},6\text{k},7\text{k}\}\), respectively. We evaluate R1--Distill models (1.5B, 7B, 14B) and Qwen3 30B-A3B on AIME’24, AIME’25, GPQA--Diamond, and LiveCodeBench under identical decoding and verification. Results are summarized in Figure~\ref{fig:prefill_ablation}.

\begin{figure}[!t]
    \centering
    \includegraphics[width=\textwidth]{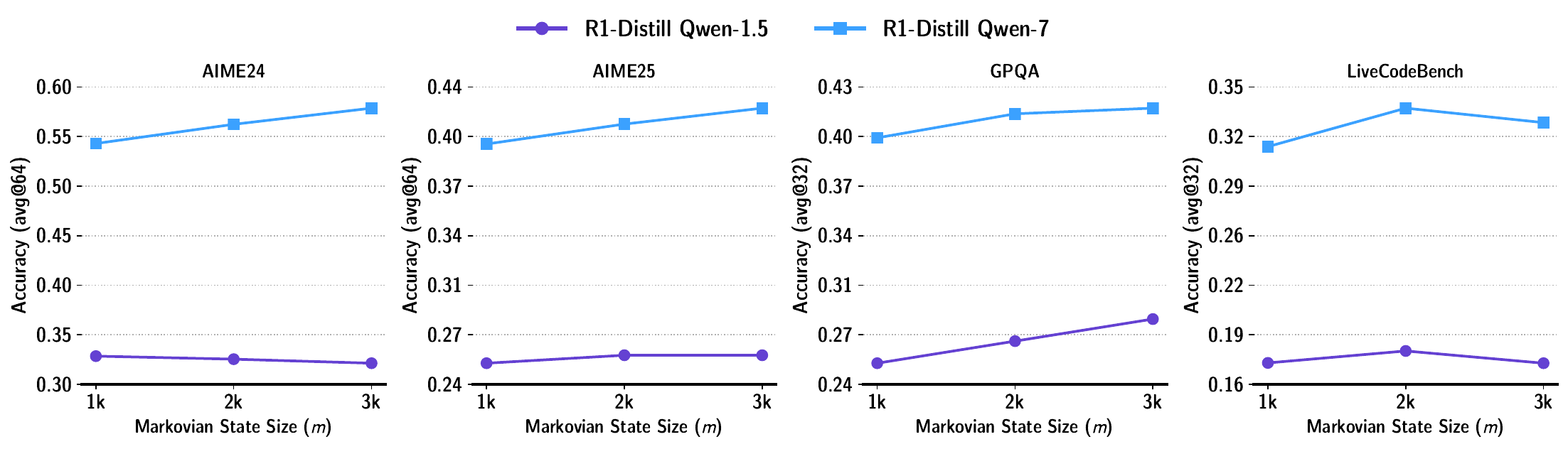}
    \caption{Ablation of the Markovian state size \(m\) at fixed per-chunk context \(\mathcal{C}=4\text{k}\). 
    In contrast to $\mathcal{C}=8\text{k}$, R1-Distill 7B shows gains from larger state sizes across most tasks,
    whereas R1-Dstill 1.5B only benefits from larger state sizes in AIME'25 and GPQA.
    }
    \label{fig:prefill_ablation_4k}
\end{figure}

Surprisingly, for the R1--Distill models the final accuracy is essentially flat across the entire range of \(m\), Given \(\mathcal{C}=8\text{k}\), shrinking the carried state from \(7\text{k}\) down to \(1\text{k}\) has negligible effect on all tasks. In contrast, Qwen3 30B-A3B exhibits a marked dependence on \(m\) for AIME’24, AIME’25, and LiveCodeBench, with larger states yielding higher accuracy. This difference aligns with two observations. First, Qwen3 is trained with a substantially larger native context window \(256\text{k}\) compared to \(32\text{k}\) for the R1 models. Second, the average Delethink trace lengths on AIME and LiveCode are considerably longer than on GPQA--Diamond, which explains why the state size matters more on these tasks. Overall, these results indicate that \(\mathcal{C}=8\text{k}\) suffices for robust Markovian behavior in R1 models with only a modest state carryover, while Qwen3 benefits from a larger state that better matches its long--context prior on long--reasoning workloads.
However, with $\mathcal{C}=4\text{k}$, even R1-Distill family show similar pattern as Qwen3 (\Cref{fig:prefill_ablation_4k}).
This suggest that there is an interplay between the context size and the average response length.

\subsection{Thinking Context Size Ablation}
\label{app:context_size_ablation}

We next ablate the effect of the per-chunk thinking context size \(\mathcal{C}\) on test-time scaling across all R1--Distill models (1.5B, 7B, 14B) and four benchmarks (AIME’24, AIME’25, GPQA--Diamond, LiveCodeBench).  
In these experiments, the $x$--axis reports the total thinking budget (number of generated tokens).  
For smaller context sizes, Delethink necessarily performs more iterations to reach the same total budget: for example, Delethink with \(\mathcal{C}=2\text{k}\) executes roughly four times as many iterations as \(\mathcal{C}=8\text{k}\) to reach 128k tokens.  We set the carried state to \(m=\mathcal{C}/2\) in every configuration so that only \(\mathcal{C}\) is varied.
This setup enables a direct comparison of how context size alone impacts performance under a fixed token budget.

\begin{figure}[!t]
    \centering
    \includegraphics[width=\textwidth]{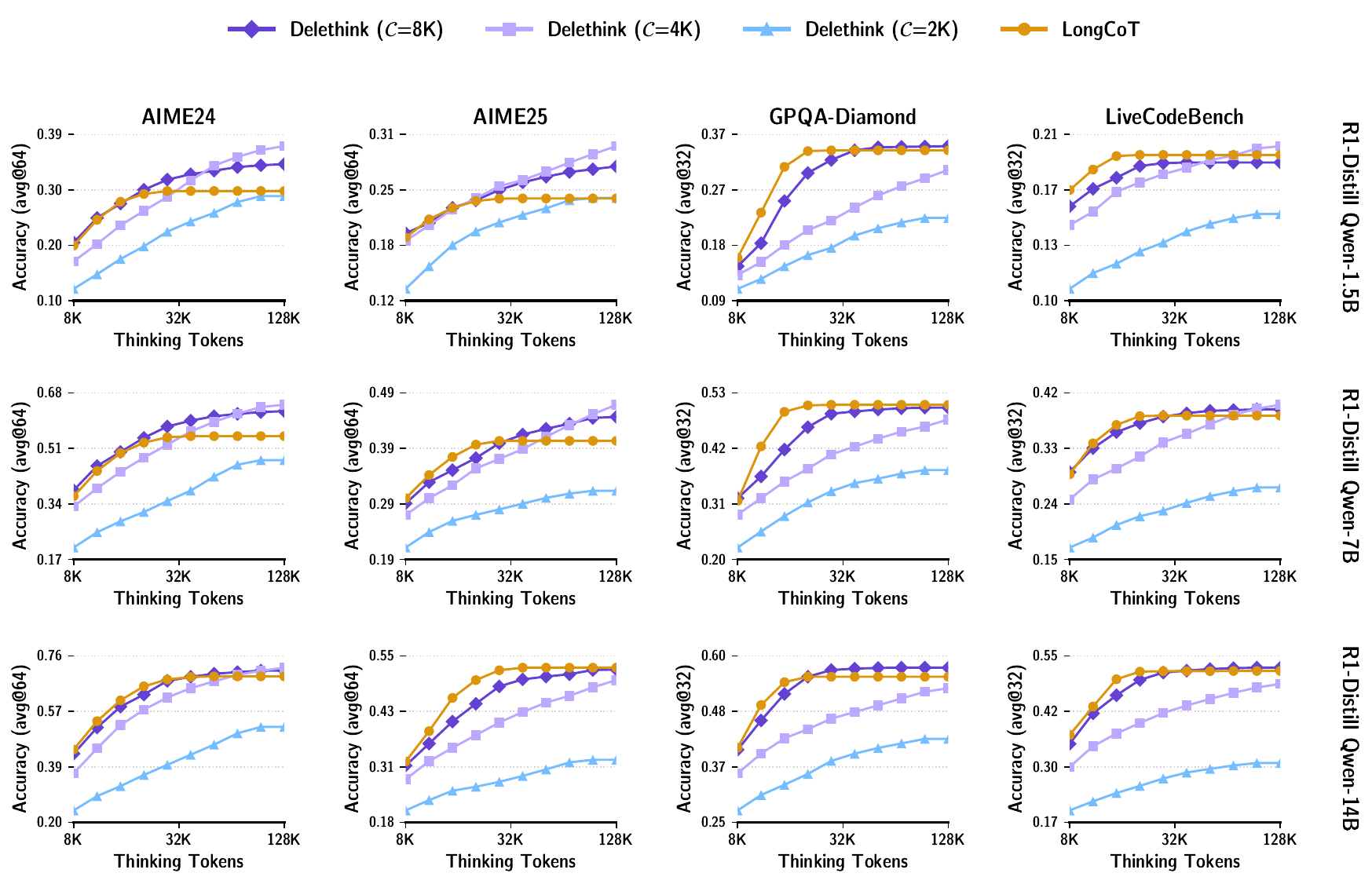}
    \caption{
    Thinking Context Size Ablation. Scaling behavior of R1-Distill models across AIME’24, AIME’25, GPQA-Diamond, and LiveCodeBench under varying per-hunk contexts \(\mathcal{C}\in\{2\text{k},4\text{k},8\text{k}\).  
    Accuracy is plotted against total thinking tokens, with smaller contexts requiring proportionally more iterations. 
    }
    \label{fig:rl_context_ablation_appendix}
\end{figure}

The results, shown in Figure~\ref{fig:rl_context_ablation_appendix}, emphasize the strong zero--shot Markovian behavior of R1 models. Despite not being trained for Markovian tracing, all R1 models steadily improve with increased budget and in many cases surpass the LongCoT baseline.  
Most notably, for the 1.5B model the \(\mathcal{C}=4\text{k}\) configuration is highly effective: its scaling curve is nearly log--linear, and it even outperforms the \(\mathcal{C}=8\text{k}\) setting on AIME’24, AIME’25, and LiveCodeBench.  
This surprising robustness highlights that smaller chunks can sometimes be more efficient for test--time scaling in smaller models.  
As the model size increases to 7B and 14B, the advantage of \(\mathcal{C}=4\text{k}\) diminishes and \(\mathcal{C}=8\text{k}\) is typically best or tied.  
By contrast, \(\mathcal{C}=2\text{k}\) proves extremely limited: it rarely approaches the LongCoT baseline and only manages to catch up on the AIME tasks for the 1.5B model at very high budgets.  
This consistent underperformance suggests that such a short context severely constrains Delethink’s ability to maintain stable traces.

\begin{figure}[]
    \centering
    \includegraphics[width=\textwidth]{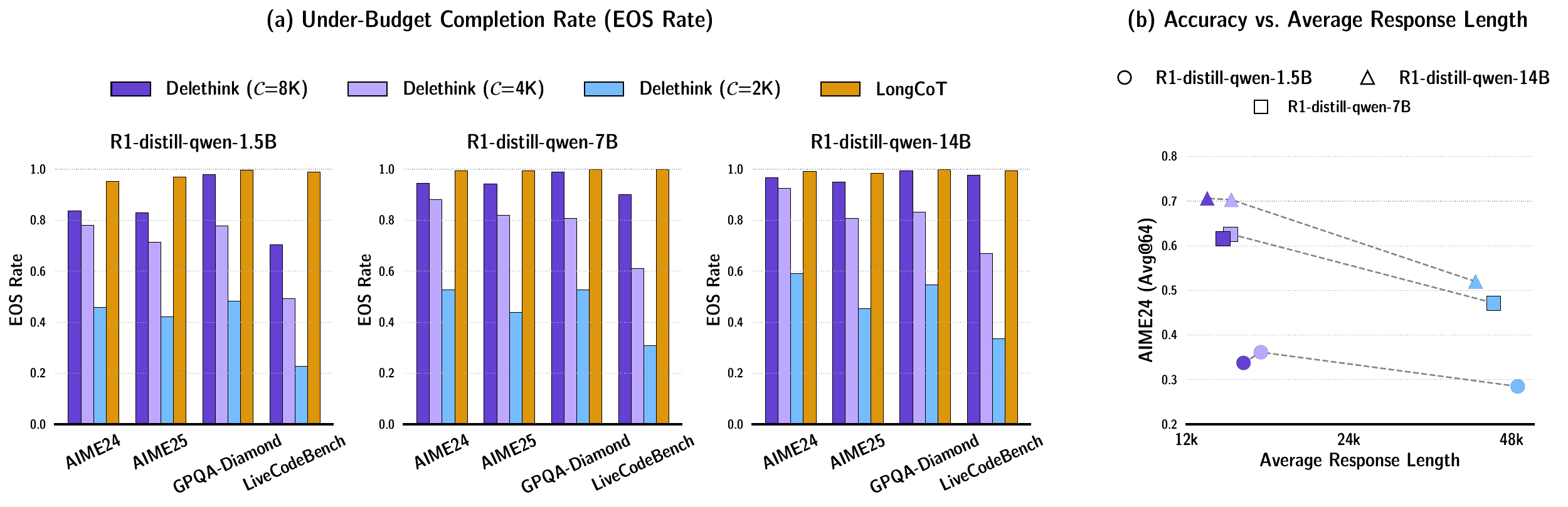}
    \caption{
    Limitation of extremely low thinking context size. 
    (a) EOS rate across tasks shows that under a 2k context, R1 models almost never terminate within budget, unlike 4k and 8k.  
    (b) On AIME’24, \(\mathcal{C}=2\text{k}\) produces much longer responses for all model sizes, indicating difficulty finalizing reasoning due to short local horizon and positional bias.  
    }
    \label{fig:context_2k_limit}
\end{figure}

The patterns become even clearer when examining completion behavior and response lengths (Figure~\ref{fig:context_2k_limit}).  
With $\mathcal{C}=2\text{k}$, models almost never terminate within budget, leading to dramatically lower EOS rates across tasks and scales.  
Correspondingly, the average response lengths at $\mathcal{C}=2\text{k}$ are much higher than at 4k or 8k, reflecting the model’s difficulty in finalizing its reasoning.  
This behavior is consistent with the limited horizon imposed by such a short per--chunk view: the model repeatedly resets positional embeddings, and the bias toward continuation overwhelms the signal to conclude.  
In practice, this prevents stable Markovian progression and explains why $\mathcal{C}=2\text{k}$ remains far below LongCoT except in a few cases (e.g., AIME at 1.5B scale).

Overall, these results provide further evidence that R1 models exhibit strong Markovian behavior even without explicit training.  
Smaller contexts can be unexpectedly effective: $\mathcal{C}=4\text{k}$ not only scales smoothly but, for the 1.5B model, even surpasses $\mathcal{C}=8\text{k}$ on several tasks.  
At the same time, the severe degradation of $\mathcal{C}=2\text{k}$ highlights the limits of compressing the local horizon too aggressively, as models struggle to converge their reasoning and rarely complete within budget.  
Together, the ablations show that test-time Markovian scaling is a robust property of off-the-shelf R1 models, while also clarifying the practical regimes in which context size supports or hinders long-horizon reasoning.

\section{Delethink Tracing on LongCoT checkpoints}
\label{sec:app:test_time_scale_longcot24k}
\begin{figure}[h]
    \centering
    \includegraphics[width=0.9\textwidth]{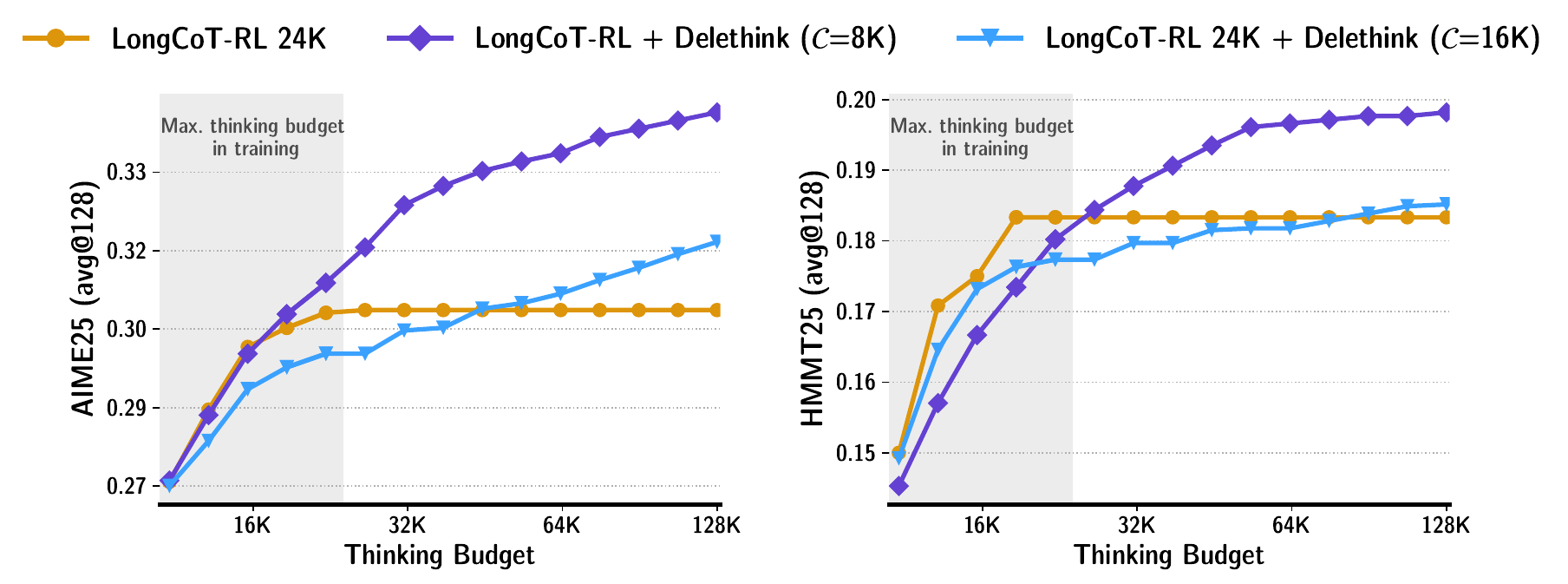}
    \caption{Delethink Tracing on a LongCoT–RL 24K checkpoint (AIME’25, avg@128). LongCoT plateaus near its trained budget, while Delethink (\(\mathcal{C}=8\text{k}\), \(\mathcal{C}=16\text{k}\)) keeps scaling; \(\mathcal{C}=8\text{k}\) yields the larger gain. The shaded region marks the trained budget.}
    \label{fig:delethink_on_longcot_checkpoint}
\end{figure}

As discussed in Section \ref{sec:why_works}, off-the-shelf LLMs show Markovian traces under Delethink, allowing us to generate traces even from models not explicitly trained with it. This suggests that applying Delethink Tracing could boost the performance of the LongCoT-$24\text{k}$ checkpoint, despite it not being trained with Delethink. We investigate this in \Cref{fig:delethink_on_longcot_checkpoint}. We find that Delethink Tracing enables the LongCoT-$24\text{k}$ checkpoint to test-time scale far beyond its training limits. Its performance increases by nearly $4\%$, almost as much as the gain it achieved from its entire RL training, at no additional training cost. This is a surprising phenomenon. Our hypothesis is that constraining queries to fewer than $8\textnormal{k}$ tokens creates the illusion, for the $24\textnormal{k}$ baseline, that it still has capacity left to continue. However, we find that when queries are kept under $16\textnormal{k}$, this effect vanishes, dissolving the \emph{infinite budget illusion}. While these results indicate that Delethink Tracing might enhance traditionally trained checkpoints, we note that such baselines require quadratically more compute to train than Delethink.

\section{Budget Force Ablation: Does S1 work on R1?}
\label{app:budget_force_ablation}

We evaluate \emph{Budget-Force} decoding scheme inspired by the S1 ``simple test-time scaling'' protocol~\citep{guo2025r1zeropaper}. We run decoding and, whenever the model \emph{finalizes early} before using the allotted token budget, we cut the response at the first finalization marker and append a short continuation cue before regenerating from the \emph{entire} prompt. Concretely, for each query we first sample with temperature $0.6$, top-$p=0.95$, $n=1$ under a 32k token budget. If the model emits any of \verb|</think>|, \verb|**Final Answer**|, or \verb|\boxed{| (or EOS) before the budget is consumed, we trim at the earliest occurrence, append the literal cue \verb|Wait\n|, and re-issue a generation from query + cut\_response. We repeat this micro-forcing until the round budget is exhausted or a continuation cap (20) is reached. To avoid distributional drift, we do not alter model decoding settings mid-run; the only intervention is the prompt-level cut-and-continue. At the end, if the model still has not emitted EOS after fully using the budget, we add a compact finalization hint (\verb|**Final Answer**\n|) and request a short natural continuation. All models identical sampling parameters across conditions.

\begin{table}[]
\centering
\caption{Accuracy (avg@128, \%) for LongCoT, Budget-Force, and Delethink across benchmarks, using identical sampling and each method’s native scaling protocol.}
\label{tab:main}
\resizebox{\textwidth}{!}{
\begin{tabular}{l l c c c c}
\toprule
\textbf{Model} & \textbf{Method} & \textbf{AIME24} & \textbf{AIME25} & \textbf{GPQA-Diamond} & \textbf{LiveCodeBench} \\
\midrule
\multirow{3}{*}{R1-distill-qwen-1.5B}
  & Long CoT 32k  & 29.3 & 23.6 & 33.9 & 19.2 \\
  & Budget Force  & 24.6 & 19.9 & 33.4 & 17.7 \\
  & Delethink     & \textbf{37.2} & \textbf{29.5} & \textbf{34.6} & \textbf{19.9} \\
\midrule\midrule
\multirow{3}{*}{R1-distill-qwen-7B}
  & Long CoT 32k  & 54.4 & 40.5 & \textbf{50.7} & 37.9 \\
  & Budget Force  & 50.1 & 35.1 & 47.8 & 36.6 \\
  & Delethink     & \textbf{64.1} & \textbf{47.1} & 50.2 & \textbf{39.7} \\
\midrule\midrule
\multirow{3}{*}{R1-distill-qwen-14B}
  & Long CoT 32k  & 68.9 & \textbf{52.8} & 55.6 & 51.5 \\
  & Budget Force  & 70.2 & 52.6 & 56.8 & 51.9 \\
  & Delethink     & \textbf{71.8} & 52.5 & \textbf{57.5} & \textbf{52.2} \\
\bottomrule
\end{tabular}
}
\end{table}

\begin{figure}[]
    \centering
    \includegraphics[width=0.8\textwidth]{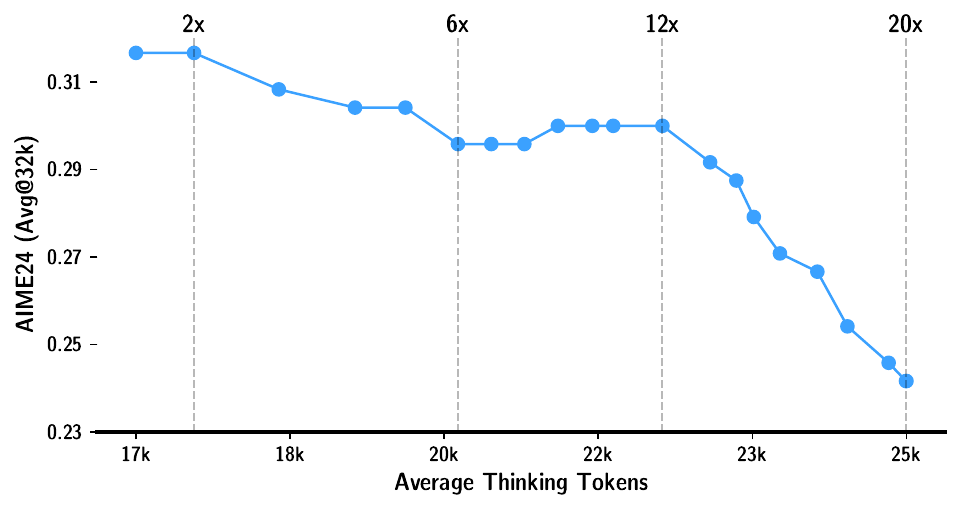}
    \caption{
    Budget forcing increases length but lowers accuracy. AIME’24 accuracy for R1–Distill–Qwen–1.5B as the number of forced cut\&continue rounds increases.  
    Average thinking tokens rise from about 17k to 25k, yet accuracy drops monotonically.  
    Vertical guides mark approximate $2\times$, $6\times$, $12\times$, and $20\times$ forcing.}
    \label{fig:budget_force}
\end{figure}

We compare three settings: (i) \textbf{Long-CoT 32k} (single pass, no forcing), (ii) \textbf{Budget-Force} as above, and (iii) \textbf{Delethink}. Unless noted. Scores reported in Table~\ref{tab:main} are accuracy on AIME'24/AIME'25, GPQA-Diamond, and LiveCodeBench under the same compute budgets. In almost all the cases, Budget Force peforms worse than the LongCoT baseline (normal sampling), demonstrating that such methods are not applicable to reasoning models like R1-distill family.

Figure~\ref{fig:budget_force} reports the outcome on AIME’24 with R1–Distill–Qwen–1.5B. As the number of forced restarts increases, the average thinking length rises from roughly \(17\text{k}\) to \(25\text{k}\) tokens, yet accuracy steadily declines. The vertical guides mark representative forcing levels that produce the indicated average lengths.

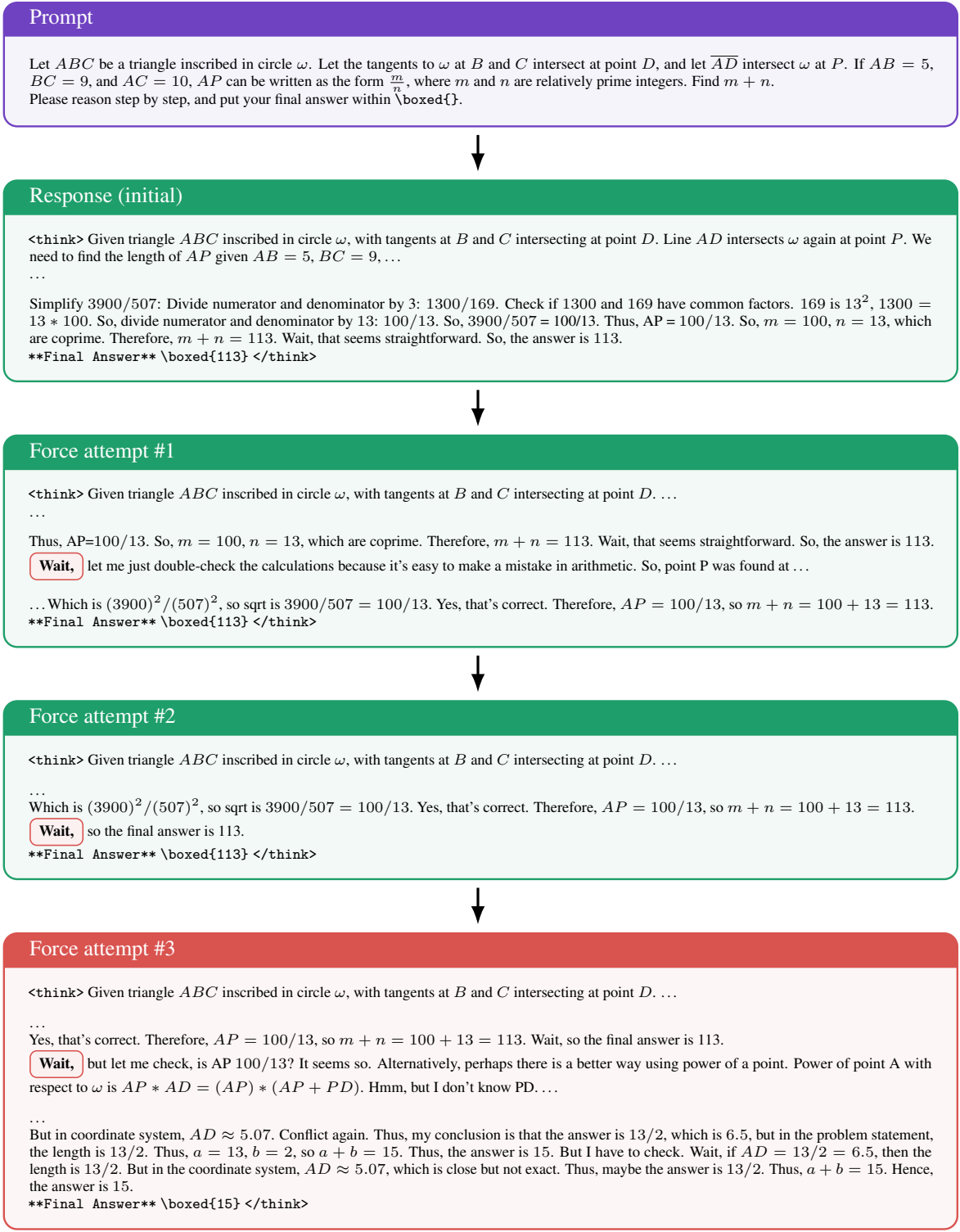
\begin{figure*}[]
\centering
\begin{tikzpicture}[node distance=6mm]

\node (p) {{
  \begin{tcolorbox}[bf-prompt, width=0.94\linewidth, title=Prompt]
    \scriptsize
    Let $ABC$ be a triangle inscribed in circle $\omega$. Let the tangents to $\omega$ at $B$ and $C$ intersect at point $D$, and let $\overline{AD}$ intersect $\omega$ at $P$. If $AB=5$, $BC=9$, and $AC=10$, $AP$ can be written as the form $\frac{m}{n}$, where $m$ and $n$ are relatively prime integers. Find $m + n$.

Please reason step by step, and put your final answer within \verb|\boxed{}|.
  \end{tcolorbox}
}};

\node[below=of p] (s1) {{
  \begin{tcolorbox}[bf-ok, width=0.94\linewidth, title=Response (initial)]
    \scriptsize
    \verb|<think>|
    Given triangle \(ABC\) inscribed in circle \(\omega\), with tangents at \(B\) and \(C\) intersecting at point \(D\). Line \(AD\) intersects \(\omega\) again at point \(P\). We need to find the length of \(AP\) given \(AB = 5\), \(BC = 9\),  \dots
    
    \dots
    \medskip
    
    Simplify $3900/507$: Divide numerator and denominator by 3: $1300/169$.
    Check if $1300$ and $169$ have common factors. $169$ is $13^2$, $1300=13*100$. So, divide numerator and denominator by $13$: $100/13$. So, $3900/507$ = 100/13. Thus, AP = $100/13$. So, $m = 100$, $n =13$, which are coprime. Therefore, $m + n = 113$. 
    Wait, that seems straightforward. So, the answer is $113$.

    \verb|**Final Answer**|
    \verb|\boxed{113}|
    \verb|</think>|
  \end{tcolorbox}
}};
\draw[very thick,-{Latex[length=3mm]}] (p.south) -- (s1.north);

\node[below=6mm of s1] (s2) {{
  \begin{tcolorbox}[bf-ok, width=0.94\linewidth, title=Force attempt \#1]
    \scriptsize
    \verb|<think>|
    Given triangle \(ABC\) inscribed in circle \(\omega\), with tangents at \(B\) and \(C\) intersecting at point \(D\). \dots

    \dots
    \medskip
    
    Thus, AP=$100/13$. So, $m =100$, $n=13$, which are coprime. Therefore, $m+n=113$. 
    Wait, that seems straightforward. So, the answer is $113$. 
    
    \WaitCue\ let me just double-check the calculations because it's easy to make a mistake in arithmetic. So, point P was found at
    \dots

    \medskip
    \dots
    Which is $(3900)^2 / (507)^2$, so sqrt is $3900/507 = 100/13$.\ Yes, that's correct.
    Therefore, $AP = 100/13$, so $m + n = 100 +13=113$.
    
    \verb|**Final Answer**|
    \verb|\boxed{113}|
    \verb|</think>|
    
  \end{tcolorbox}
}};
\draw[very thick,-{Latex[length=3mm]}] (s1.south) -- (s2.north);

\node[below=6mm of s2] (s3) {{
  \begin{tcolorbox}[bf-ok, width=0.94\linewidth, title=Force attempt \#2]
    \scriptsize
    \verb|<think>|
    Given triangle \(ABC\) inscribed in circle \(\omega\), with tangents at \(B\) and \(C\) intersecting at point \(D\). \dots

    \medskip
    \dots
    
    Which is $(3900)^2 / (507)^2$, so sqrt is $3900/507 = 100/13$.\ Yes, that's correct.
    Therefore, $AP = 100/13$, so $m + n = 100 +13=113$.

    \WaitCue\ so the final answer is 113.

    \verb|**Final Answer**|
    \verb|\boxed{113}|
    \verb|</think>|
    
  \end{tcolorbox}
}};
\draw[very thick,-{Latex[length=3mm]}] (s2.south) -- (s3.north);

\node[below=6mm of s3] (s4) {{
  \begin{tcolorbox}[bf-force, width=0.94\linewidth, title=Force attempt \#3]
    \scriptsize
    \verb|<think>|
    Given triangle \(ABC\) inscribed in circle \(\omega\), with tangents at \(B\) and \(C\) intersecting at point \(D\). \dots

    \medskip
    \dots
    
    Yes, that's correct. Therefore, $AP = 100/13$, so $m + n = 100 +13=113$. Wait, so the final answer is 113.

    \WaitCue\ but let me check, is AP $100/13$? It seems so. Alternatively, perhaps there is a better way using power of a point. Power of point A with respect to $\omega$ is $AP*AD=(AP)*(AP+PD)$. Hmm, but I don't know PD. \dots

    \medskip
    \dots

    But in coordinate system, $AD\approx5.07$. Conflict again. Thus, my conclusion is that the answer is $13/2$, which is $6.5$, but in the problem statement, the length is $13/2$. Thus, $a=13$, $b=2$, so $a+b=15$. Thus, the answer is $15$. But I have to check. Wait, if $AD=13/2=6.5$, then the length is $13/2$. But in the coordinate system, $AD\approx5.07$, which is close but not exact. Thus, maybe the answer is $13/2$. Thus, $a+b=15$. Hence, the answer is $15$.
    
    \verb|**Final Answer**|
    \verb|\boxed{15}|
    \verb|</think>|
    
  \end{tcolorbox}
}};
\draw[very thick,-{Latex[length=3mm]}] (s3.south) -- (s4.north);

\end{tikzpicture}

\caption{Failed example of S1-style budget forcing on an AIME’24 question with \textsc{R1-Distill-Qwen-1.5B}. The first pass is correct in 9{,}287 tokens (<32k budget). After two cut\&continue restarts, each appending the cue \WaitCue, pass \#2 remains correct (9{,}778 tokens), but pass \#3 drifts and ends with the wrong answer.}
\label{fig:budget-force-detailed}
\end{figure*}

The worked case in \ref{fig:budget-force-detailed} explains the mechanism. The first pass reaches the correct boxed answer well under budget. Two subsequent forced restarts, each prefixed with the continuation cue \texttt{Wait}, still land on the same correct solution. The next forced restart reopens already settled subproblems, triggers redundant re-derivations, and the chain drifts to an incorrect final answer. Regenerating from the full prompt with a generic continuation cue does not preserve a compact notion of progress. For reasoning models that already perform self-check and backtracking, the external nudge encourages second-guessing and exploration of alternative branches rather than continuation from a stable intermediate state. In short, budget forcing reliably buys more tokens, but those extra tokens often overwrite useful partial work instead of extending it, which explains the accuracy drops in Table~\ref{tab:main} even as length increases in Figure~\ref{fig:budget_force}.

\section{AIME'25 detailed Length vs. Accuracy}
\label{app:extended_thinking_aime_detailed}

\begin{figure}[!t]
    \centering
    \includegraphics[width=0.8
    \textwidth]{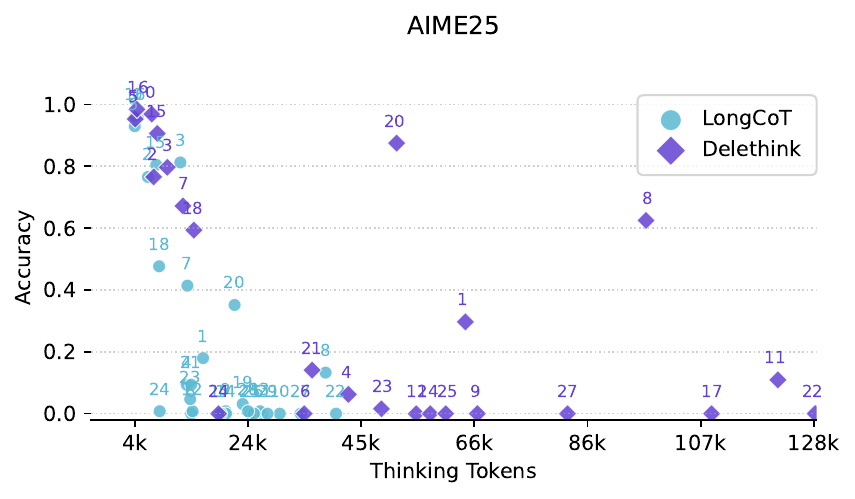}
    \caption{The average thinking length per question and its corresponding accuracy. Delethink Tracing truly test-time scales R1DistillQwen1.5B performance on AIME'25.}
    \label{fig:aime25_extended_thinking}
\end{figure}

We examine per–problem test–time scaling on AIME’25. For each of the 30 questions we sample 128 independent responses and record two quantities: the mean number of thinking tokens consumed by the model and the empirical Pass@1 on that question. Figure~\ref{fig:aime25_extended_thinking} plots these pairs as a scatter, with each marker annotated by the problem index. We report both LongCoT and Delethink under identical decoding settings. The $x$–axis is the average thinking length for that problem, measured under sequential Delethink tracing with a budget that can extend up to 128k tokens.

The plot shows genuine test–time scaling at the \emph{item} level. Several questions that are rarely solved at shorter traces become reliably solvable when Delethink is allowed to think longer. Notably, problems 1, 8, and 20 move to the upper right of the plot under Delethink, indicating that additional thinking tokens translate into higher per–item accuracy. By contrast, LongCoT points concentrate at shorter lengths and plateau at lower accuracy, aligning with the global curves in the main paper. The effect is not universal across all items, which is expected, but the presence of multiple questions whose accuracy improves only after very long traces illustrates how sequential thinking enables real test–time scaling rather than simple repetition.

\section{Variance from small k in avg@k}
\label{sec:variance_k}

Using small \(k\) yields high evaluation variance across benchmarks, with the effect most acute on small test sets.  
Since \(\mathrm{Var}[\widehat{\mathrm{avg@}k}] \approx \frac{p(1-p)}{Q\,k}\), variance scales inversely with both the number of problems \(Q\) and the number of sampled continuations \(k\); halving \(k\) roughly doubles the variance.  
Thus tasks with few questions (e.g., AIME’24/’25 with \(Q=30\)) are particularly sensitive, making \(\mathrm{avg@}16\) an unstable estimate of true performance and unreliable for model ranking or ablations.

\begin{figure}[]
    \centering
    \includegraphics[width=\textwidth]{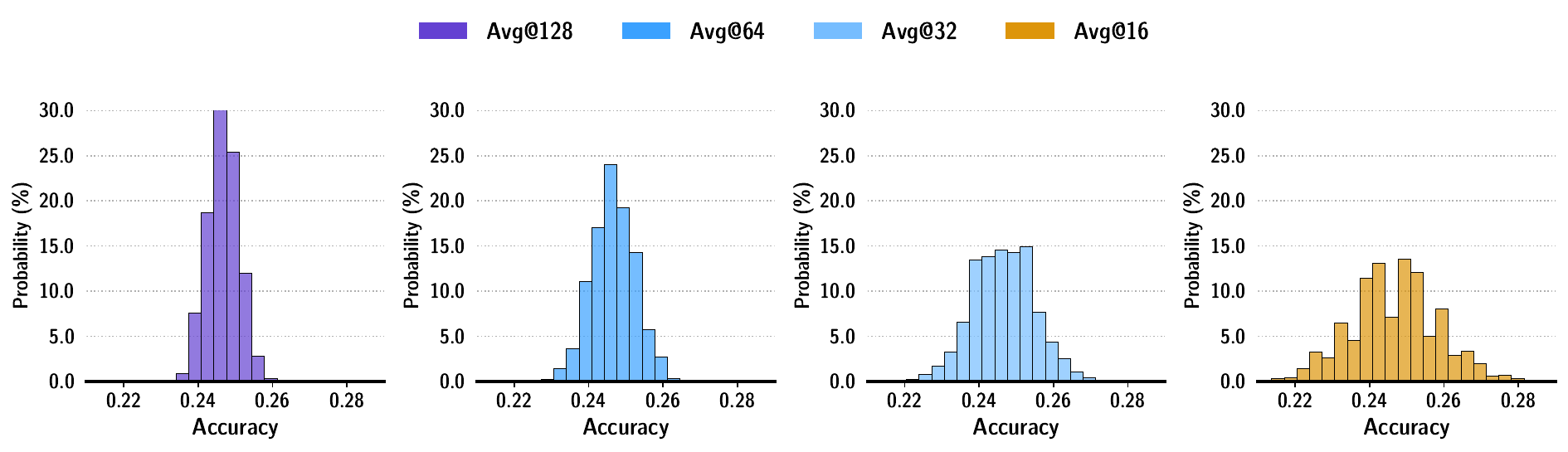}
    \caption{Example variance of \(\mathrm{avg@}k\) on AIME’25 for R1–Distill Qwen–1.5B (LongCoT, 32k) using \(N=512\) continuations and \(B=5000\) bootstrap replicates. Variance increases sharply as \(k\) decreases; small-\(Q\) tasks make \(\mathrm{avg@}16\) particularly unstable.}
    \label{fig:accuracy_histograms}
\end{figure}

To illustrate, we take R1–Distill Qwen–1.5B under LongCoT, 32k on AIME’25, generate \(N=512\) continuations per problem, and draw \(B=5000\) bootstrap replicates of \(\mathrm{avg@}k\) by resampling \(k\) continuations per problem with replacement and averaging across problems.  
Figure~\ref{fig:accuracy_histograms} shows the sampling distributions for \(k\in\{16,32,64,128\}\): the dispersion shrinks rapidly with \(k\); \(\mathrm{avg@}16\) spans several percentage points, \(\mathrm{avg@}32\) remains noisy, while \(\mathrm{avg@}128\) is sharply concentrated. In all main results we therefore report \(\mathrm{avg@}128\).  
When smaller \(k\) is unavoidable, we recommend at least \(k\ge 64\) and reporting bootstrap uncertainty bands to avoid overinterpreting chance differences, especially on small-\(Q\) evaluations.

\newpage

\section{The Difference Between Delethink and Window Attention}
\label{app:full-vs-stream-vs-delethink}

\paragraph{Delethink (a Markovian Thinking paradigm).}
Delethink is an RL formulation that instantiates the Markovian Thinking paradigm by \emph{redesigning the RL environment}. The policy is required to reason in a sequence of fixed-size chunks. At each chunk boundary, the environment stops generation, \emph{resets the context}, and reinitializes the prompt with the original query plus a short textual carryover from the previous chunk. The policy therefore is forces to learn to place a Markovian state near the end of every chunk so that, after reset, it can continue reasoning seamlessly in the next chunk. Delethink makes no architectural assumptions and, in principle, applies to any reasoning model.

\paragraph{Sliding-window and streaming attention (architecture-level methods).}
Sliding-window attention (SWA) restricts each token to attend only to the previous $w$ tokens, truncating attention edges to a width-$w$ band. In practice, pure SWA loses the query once the window slides past it, so implementations often interleave windowed attention with periodic full-attention blocks or use \emph{streaming attention}, which pins the prompt in context while applying a rolling window to the remainder. These mechanisms are primarily deployed to extend effective context length efficiently \citep{Agarwal2025:GPT-OSS}.
Delethink modifies the \emph{state definition and transition rule} of the RL environment, whereas SWA/streaming modify the \emph{model’s attention pattern}. Under Delethink, the only cross-chunk information available to the policy is the short textual carryover that the model explicitly writes—an intentional, bounded Markovian state. Under SWA/streaming, there is no reset: the model consumes a continuous token stream, and information outside the current window can persist \emph{implicitly} to the extent it has been encoded into tokens that remain within the window. This is a fundamental difference in how long-horizon information is preserved: Delethink enforces explicit state passing across chunk boundaries; SWA/streaming rely on implicit retention within a bounded receptive field (\Cref{fig:mask-vs-delethink-8tok}).

\paragraph{Orthogonality.}
These approaches are orthogonal and can be combined. Delethink can train policies that use full attention, SWA, or streaming attention inside chunks; conversely, applying windowed/streaming attention within Delethink can further reduce within-chunk cost. Delethink’s benefits do not depend on any specific attention variant.

\paragraph{A KV-state MDP (theoretical).}
\emph{One could, in principle, imagine} to  instantiate Markovian Thinking by defining the RL state as a \emph{KV cache enteries} and designing the environment to keep that state size fixed—thereby obtaining linear time and constant memory by construction, and recasting RL directly in terms of KV-state transitions. To the best of our knowledge, this has not been realized. A practical implementation would require careful and efficient GPU-kernels across inference and training stacks; current systems (e.g., SGLang, Hugging Face Transformers, FSDP) do not provide this at the time of writing. We leave an efficient KV-based variants of Markovian RL environment to future work.

To what level such KV-state MDP trains effectively compared to LongCoT-RL is beyond the scope of this paper. Delethink environment is deliberately designed to be simple: To demonstrate the possibility of Markovian Thinking and delivering linear compute and constant memory \emph{without} modifying the model architecture, enabling immediate implementation on existing infrastructure.

\begin{figure}[t]
\centering
\begin{tikzpicture}[x=0.48cm,y=0.48cm,font=\scriptsize]

\def\DX{13}      %
\def\DY{13}      %
\def\DXchunk{10}  %
\tikzset{panellabel/.style={font=\normalsize}}

\newcommand{\DrawGrid}[5]{%
  \begin{scope}[shift={(#1,#2)}]
    \draw[black] (0,0) rectangle (#3,#3);
    \foreach \i in {1,...,#3} {
      \draw[black!30] (0,\i) -- (#3,\i);
      \draw[black!30] (\i,0) -- (\i,#3);
    }
    \foreach \t [count=\i] in #4 { \node[anchor=south] at (\i-0.5,#3+0.30) {\t}; }
    \foreach \t [count=\i] in #5 { \node[anchor=east]  at (-0.35,\i-0.5) {\t}; }
  \end{scope}
}
\newcommand{\FillTri}[3]{%
  \begin{scope}[shift={(#1,#2)}]
    \foreach \r in {1,...,#3} { \foreach \c in {1,...,\r} {
      \fill[black!75] (\c-1,\r-1) rectangle (\c,\r); } }
  \end{scope}
}
\newcommand{\FillWindow}[4]{%
  \begin{scope}[shift={(#1,#2)}]
    \foreach \r in {1,...,#3} {
      \pgfmathtruncatemacro{\L}{max(1,\r-#4)}
      \foreach \c in {\L,...,\r} {
        \fill[black!75] (\c-1,\r-1) rectangle (\c,\r); } }
  \end{scope}
}
\newcommand{\FillCols}[4]{%
  \begin{scope}[shift={(#1,#2)}]
    \foreach \r in {1,...,#3} { \foreach \cc in #4 {
      \fill[black!75] (\cc-1,\r-1) rectangle (\cc,\r); } }
  \end{scope}
}

\def\TopEight   {A,B,C,D,E,F,G,H}
\def\LeftEight  {A,B,C,D,E,F,G,H}
\def\TopSix     {A,B,C,D,E,F}
\def\LeftSix    {A,B,C,D,E,F}
\def\TopSixAFGH {A,B,E,F,G,H}
\def\LeftSixAFGH{A,B,E,F,G,H}

\DrawGrid{0}{0}{8}{\TopEight}{\LeftEight}
\FillTri{0}{0}{8}
\node[panellabel,anchor=north west] at (-3.5,-1.5) {(a) Full attention (single pass; causal triangle)};

\DrawGrid{\DX}{0}{8}{\TopEight}{\LeftEight}
\FillWindow{\DX}{0}{8}{2}
\node[panellabel, anchor=north west] at (\DX,-1.5) {(b) Sliding/Streaming: window $w{=}2$ (single pass)};

\DrawGrid{0}{-\DY}{8}{\TopEight}{\LeftEight}
\FillWindow{0}{-\DY}{8}{2}
\FillCols{0}{-\DY}{8}{{1,2}} %
\node[panellabel, anchor=north west] at (-3.5,-\DY-1.5) {(c) Streaming w/ anchors: window $w{=}2$ + global A,B};

\pgfmathsetmacro{\Xd}{\DX}                %
\pgfmathsetmacro{\Yd}{-\DY}               %
\pgfmathsetmacro{\Xdr}{\DX+\DXchunk}      %

\DrawGrid{\Xd}{\Yd}{6}{\TopSix}{\LeftSix}
\FillTri{\Xd}{\Yd}{6}
\node[align=center, fill=white, inner sep=1pt] at (\Xd+3,-\DY+7.8) {chunk 1\\(full AR \emph{within} chunk)};

\DrawGrid{\Xdr}{\Yd}{6}{\TopSixAFGH}{\LeftSixAFGH}
\FillTri{\Xdr}{\Yd}{6}
\node[align=center, fill=white, inner sep=1pt] at (\Xdr+3,-\DY+7.8) {chunk 2\\(reset; only A,B,E,F carry)};

\draw[red,very thick]
  ({\Xd+7.2},{\Yd+0.2}) -- ({\Xd+7.9},{\Yd+2.0}) -- ({\Xd+7.2},{\Yd+3.8})
  -- ({\Xd+7.9},{\Yd+5.6}) -- ({\Xd+7.2},{\Yd+7.4});
\node[red,fill=white,inner sep=1pt,align=center] at ({\Xd+7.35},{\Yd+4.0}) {KV reset\\(new pass)};

\node[panellabel, anchor=north west] at (\Xd,-\DY-1.5) {(d) \textbf{Delethink}: two \emph{separate} passes (Markovian state is textual)};

\end{tikzpicture}

\caption{
(a) Full attention: single forward pass.
(b) Sliding/Streaming: local window $w$ in the same pass.
(c) Streaming+anchors: global A,B; recent K/V were computed when earlier tokens were present.
(d) \textbf{Delethink}: generate chunk~1 on $[A..F]$, then \emph{reset} and re-encode $[A,B,E,F,G,H]$ to continue. No cross-chunk K/V survives; any necessary info must be written into the carried \emph{textual} state before the reset.}
\label{fig:mask-vs-delethink-8tok}
\end{figure}

\end{document}